%% file: article.tex
\documentclass[10pt, twocolumn]{article}
\usepackage{framed,multirow,graphicx}
\usepackage[sort&compress,square,comma,numbers]{natbib}

\usepackage{amssymb}
\usepackage{latexsym}
\usepackage{url}
\makeatletter
\g@addto@macro{\UrlBreaks}{\UrlOrds}
\makeatother
\usepackage[dvipsnames]{xcolor}\usepackage{amsmath}
\usepackage{totcount}
\definecolor{newcolor}{rgb}{.8,.349,.1}
\usepackage[stable]{footmisc}

\usepackage[draft]{todonotes} 
\usepackage{lineno}
\usepackage{pgf}
\usepackage{transparent}
\usepackage[titletoc]{appendix}

\usepackage{xspace}

\makeatletter
\DeclareRobustCommand\onedot{\futurelet\@let@token\@onedot}
\def\@onedot{\ifx\@let@token.\else.\null\fi\xspace}

\def\eg{\emph{e.g}\onedot} 
\def\ie{\emph{i.e}\onedot} 
 
\def\etc{\emph{etc}\onedot} 
\def\wrt{w.r.t\onedot} 

\makeatother

\title{Recent Advances in Object Detection in the Age of Deep Convolutional Neural Networks}
\author{Shivang Agarwal$^{(*,1)}$, Jean Ogier du Terrail$^{(*,1,2)}$, Fr\'ed\'eric Jurie$^{(1)}$\\[1em]
$^{(*)}$ equal contribution\\
$^{(1)}$Normandie Univ, UNICAEN, ENSICAEN, CNRS\\
$^{(2)}$Safran Electronics and Defense}

\begin{document} 
\onecolumn 
\maketitle
\input{article_abstract.tex}
\pagebreak

\pagebreak

{\linespread{.95}\small\tableofcontents}

\twocolumn
\input{article_introduction}
\input{article_detector_design.tex}

\input{article_going_forward.tex}

\input{article_extensions.tex}

\input{article_conclusions}
\bibliographystyle{plainnat}
\bibliography{refs}
\pagebreak
\begin{appendices}
\input{article_datasets}

\end{appendices}
\end{document}

%% file: article_abstract.tex
\begin{abstract}
Object detection, the computer vision task dealing with detecting instances of objects of a certain class (\eg, 'car', 'plane', \etc) in images, attracted a lot of attention from the community during the last six years. This strong interest can be explained not only by the importance this task has for many applications but also by the phenomenal advances in this area since the arrival of deep convolutional neural networks (DCNNs). This article reviews the recent literature on object detection with deep CNN, in a comprehensive way. This study covers not only the design decisions made in modern deep (CNN) object detectors, but also provides an in-depth perspective on the set of challenges currently faced by the computer vision community, as well as some complementary and new directions on how to overcome them. In its last part it goes on to show how object detection can be extended to other modalities and conducted under different constraints. This survey also reviews in its appendix the public datasets and associated state-of-the-art algorithms.
\end{abstract}

%% file: article_introduction.tex
\section{Introduction} \label{sec:intro}
The task of automatically recognizing and locating objects in images and videos is important in order to make computers able to understand or interact with their surroundings. For humans, it is one of the primary tasks, in the paradigm of visual intelligence, in order to survive, work and communicate. If one wants machines to work for us or with us, they will need to make sense of their environment as good as humans or in some cases even better than humans. Solving the problem of object detection with all the challenges it presents has been identified as a major precursor to solving the problem of semantic understanding of the surrounding environment.

A large number of academics as well as industry researchers have already shown their interest in it by focusing on applications, such as autonomous driving, surveillance, relief and rescue operations, deploying robots in factories, pedestrian and face detection, brand recognition, visual effects in images, digitizing texts, understanding aerial images, \etc which have object detection as a major challenge at their core. 

The {\em Semantic Gap}, defined by \citeauthor{smeulders2000ContentBasedImageRetrieval} \cite{smeulders2000ContentBasedImageRetrieval} as the lack of coincidence between the information one can extract from some visual data and its interpretation by a user in a given situation, is one of the main challenges object detection must deal with. There is indeed a difference of nature between raw pixel intensities contained in images and semantic information depicting objects.

\begin{figure*}
\centering
    \includegraphics[width=\textwidth]{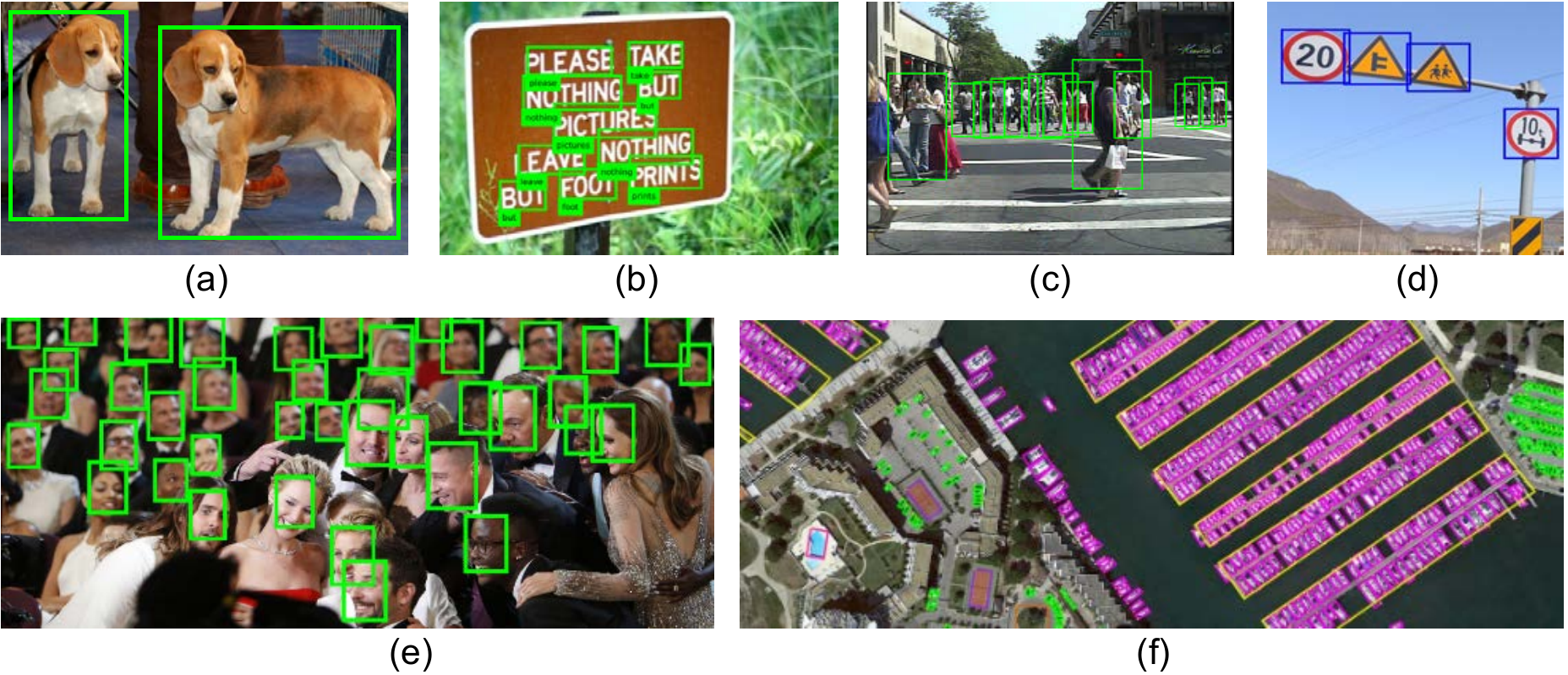}
    \caption{Visualization of sample examples form different kinds of dataset for the detection task. (a) generic object detection \cite{everingham2010pascal}, (b) text detection \cite{gupta2016SyntheticDataText}, (c) pedestrian detection \cite{walk2010new}, (d) traffic-sign detection \cite{zhu2016traffic}, (e) face detection \cite{yang2015FacialPartsResponses} and (f) objects in aerial images detection \cite{xia2017dota}. 
    \label{fig:datasets}}
\end{figure*}

Object detection is a natural extension of the classification problem. The added challenge is to correctly detect the presence and accurately locate the object instance(s) in the image (Figure \ref{fig:datasets}). It is (usually) a supervised learning problem in which, given a set of training images, one has to design an algorithm which can accurately locate and correctly classify as many object instances as possible in a rectangle box while avoiding false detections of background or multiple detections of the same instance. The images can have object instances from same classes, different classes or no instances at all. The object categories in training and testing set are then supposed to be statistically similar. The instance can occupy very few pixels, 0.01\% to 0.25\%, as well as the majority of the pixels, 80\% to 90\%, in an image. Apart from the variation in size the variation can be in lighting, rotation, appearance, background, \etc There may not be enough data to accurately cover all the variations well enough. Small objects, particularly, give low performance at being detected because the available information to detect them is present but compressed and hard to decode without some prior knowledge or context. Some object instances can also be occluded. 

An additional difficulty is that real world applications like video object detection demand this problem to be solved in real time. With the current state of the art detectors that is often not the case. Fastest detectors are usually worse than the best performing ones (\eg heavy ensembles). 

We present this review to connect the dots between various deep learning and data driven techniques proposed in recent years, as they have brought about huge improvements in the performance, even though the recently introduced object detection datasets are much more challenging. We intend to study what makes them work and what are their shortcomings. We discuss the seminal works in the field and the incremental works which are more application oriented. We also see their approach on trying to overcome each of the challenges. The earlier methods which were based on hand-crafted features are outside the scope of this review. The problems that are related to object detection such as semantic segmentation are also outside the scope of this review, except when used to bring contextual information to detectors. Salient object detection being related to semantic segmentation will also not be treated in this survey.

Several surveys related to object detection have been written in the past, addressing specific tasks such as pedestrian detection \citep{enzweiler2008monocular}, moving objects in surveillance systems \citep{joshi2012SurveyMovingObject}, object detection in remote sensing images \citep{cheng2016SurveyObjectDetection}, face detection \citep{hjelmas2001FaceDetectionSurvey,zhang2010SurveyRecentAdvances}, facial landmark detection \citep{wu2018FacialLandmarkDetection}, to cite only some illustrative examples. In contrast with this article, the aforementioned surveys do not cover the latest advances obtained with deep neural networks.
Recently four non peer reviewed surveys appeared on arXiv that also treat the subject of Object Detection using Deep Learning methods. 
 This article shares the same motivations as \citep{zhao2018ObjectDetectionDeep} and \citep{chahal2018survey}, but covers the topic more comprehensively and extensively as these two surveys which only cover the backbones and flagship articles associated with modern object detection. This work investigates more thoroughly papers that one would not necessarily call mainstream, like boosting methods or true cascades, and study related topics like weakly supervised learning and approaches that carry promises but that have yet to become widely used by the community (graph networks and generative methods). Concurrently to this article, the paper by \citep{liu2018deep} goes into many details about the modern object detectors. We wanted this survey to be more than just an inventory of existing methods but to provide the reader with a complete tool-set to be able to understand fully how the state of the art came to be and what are the potential leads to advance it further, by studying surrounding topics such as interpretability, lifelong detectors, few-shot learning or domain adaptation (in addition to delving into non mainstream methods already mentioned).

The following subsections  give an overview of the problem, some of the seminal works in the field (hand-crafted as well as data driven) and describe the task and evaluation methodology. Section~\ref{sec:detector_design}  goes into the detail of the design of the current state-of-the-art models. Section~\ref{sec:going_forward}  presents recent methodological advances as well as the main challenge modern detectors have to face. Section \ref{sec:extensions} shows how to extend the presented detectors to different detection tasks (video, 3D) or perform under different constraints (energy efficiency, training data, \etc). 
Finally, Section~\ref{sec:conclusion} concludes the review.
We also list a wide variety of datasets and the associated state of the art performances in the Appendix.

\subsection{What is object detection in images? How to evaluate detector performance?} \label{sec:intro_definition}

\subsubsection{Problem definition}
Object detection is one of the various tasks related to the inference of high-level information from images. Even if there is no universally accepted definition of it in the literature, it is usually defined as the task of locating all the instances of a given category (\eg 'car' instances in the case of car detection) while avoiding raising alarms when/where no instances are present. The localization can be provided as the center of the object on the image, as a bounding box containing the object, or even as the list of the pixels belonging to the object. In some rare cases, only the presence/absence of at least one instance of the category is sought, without any localization. 

Object detection is always defined with respect to a dataset containing images associated with a list containing some descriptions (position, scale, \etc) of the objects each image contains. Let's denote by $\cal I$ an image and $O(I)$ the set of $N_I^*$ object descriptions, with:
\begin{equation}
	O(I) = \{(Y_1^*,Z_1^*),\ldots,(Y_i^*,Z_i^*),\ldots,(Y_{N_i^*}^*,Z_{N_i^*}^*) \}\nonumber
\end{equation} 

where $Y_i^* \in \cal Y$ represent the category of the i-th object and $Z_{N_i^*}^* \in \cal Z$  a representation of its location/scale/geometry in the image. $\cal Y$ is the  set of possible categories, which can be hierarchical or not. $\cal Y$ is the space of possible locations/scales/geometries of objects in images. It can be the position of the center of the object $(x_c, y_c)\in R^2$,
 a bounding box  $(x_{min},y_{min},x_{max},y_{max})\in R^4$ encompassing the object, a mask, \etc.
 
Using these notations, object detection can be defined as the function associating an image with a set of detections 	\[D(I,\lambda) = \{(Y_1,Z_1),\ldots,(Y_i,Z_i),\ldots,(Y_{N_i(\lambda)},Z_{N_i(\lambda)}). \] The operating point $\lambda$ allows to fix a tradeoff between false alarms and missed detections.

Object detection is related but different from {\em object segmentation}, which aims to group pixels from the same object into a single region, or {\em semantic segmentation} which is similar to object segmentation except that the classes may also refer to varied backgrounds or 'stuff' (\eg 'sky', 'grass', 'water' \etc, categories). It is also different from {\em Object Recognition} which is usually defined as recognizing (\ie giving the name of the category) of an object contained in an image or a bounding box, assuming there is only one object in the image. For some authors {\em Object Recognition} involves detecting all the objects in an image. {\em Instance object detection} is more restricted than object detection as the detector is focused on a single object (\eg a particular car model) and not any object of a given category. In case of videos, object detection task is to detect the objects on each frame of the video.

\subsubsection{Performance evaluation}
Evaluating the function of detection for a given image $I$ is done by comparing the actual list of objects locations $O(I)$  (so-called the ground truth) of a given category with the detections $D(I,\lambda)$ provided by the detector. Such a comparison is possible only once the two following definitions are given:
\begin{enumerate}
    \item A geometric compatibility function \[G: (Z,Z^*)\in {\cal Z}^2 \rightarrow \{0,1\} \] defining the conditions that must be met for considering two locations as equivalent.
    \item An association matrix $A\in \{0,1\}^{N(I,\lambda)\times N^*(I)}$ defining a bipartite graph between the detected objects $\{Z_1, \cdots, Z_{N(I,\lambda)}\}$ and the $A(i,j)\leq 1$ ground truth objects $\{Z^*_1, \cdots, Z^*_{N^*(I)}\}$, with: \[ \sum_{j=1}^{N(I,\lambda)}A(i,j)\leq 1\] \[ \sum_{i=1}^{N^*(I)}A(i,j)\leq 1\] \[ G(Z^*_i,Z_j)=0 \implies A(i,j)=0 \] 
\end{enumerate}

\begin{figure}
\centering
    \includegraphics[width =8cm]{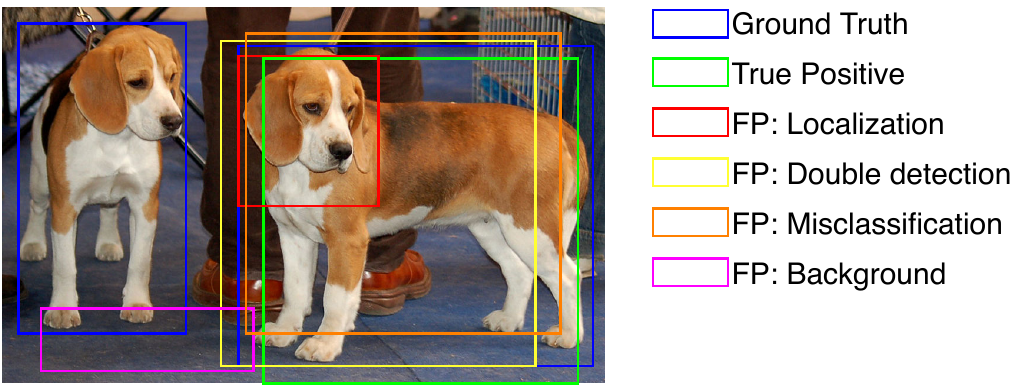}
    \caption{ An illustration of predicted boxes being marked as \textit{True Positive} (TP) or \textit{False Positive} (FP). The blue boxes are ground-truths for the class "dog". Predicted box is marked as TP if the predicted class is correct and the overlap with the ground truth is greater than a threshold. It is marked as FP if it has overlap less than that threshold or same object instance is detected again or it is misclassified or a background is predicted as an object instance. The left dog is marked as \textit{False Negative} (FN). Best viewed in color.
    \label{fig:evaluation}}
\end{figure}

With such definitions, the number of correct detections is given by \[TP(I,\lambda)=\sum_{i,j}A(i,j)\] If several association matrices $A$ satisfy the previous constraints, the one maximizing the number of correct detections is chosen. An illustration of TP and False Positives (FP) in an image is shown in Figure \ref{fig:evaluation}.

Such a definition can be viewed as the size of the maximal matching in a bipartite graph. Nodes are locations (ground truth, on the one hand, detections on the other hand). Edges are based on the acceptance criterion $G$ and the constraints stating that ground truth object and detected objets can be associated only once each.

It is possible to average the correct detections at the level of a test set $T$ through the two following ratios:

\[ Precision(\lambda) = \frac{\sum_I TP(I,\lambda)}{\sum_{I\in T} N(I,\lambda)}\]

\[ Recall(\lambda) = \frac{\sum_I TP(I,\lambda)}{\sum_{I\in T} N^*(I)}\]

Precision/Recall curve is obtained by varying the operational point $\lambda$. The Mean Average Precision can be computed by averaging the Precision for several Recall values (typically 11 equally spaced values). 

The definition of $G$ can vary from  data sets to data sets. However, only a few definitions  reflect most of the current research. One of the most common one comes the Pascal VOC challenge \citep{everingham2010pascal}. It assumes ground truths are defined by non-rotated rectangular bounding boxes containing object instances, associated with class labels. The diversity of the methods to be evaluated prevents the use of ROC (Receiver Operating Characteristic) or DET (Detection Error Trade-off), commonly used for face detection, as it would assume all the methods use the same window extraction scheme (such as the sliding window mechanism), which is not always the case. In the Pascal VOC challenge, object detection is evaluated by one separate AP score per category. For a given category, the Precision/Recall curve is computed from the ranked outputs (bounding boxes) of the method to be evaluated. Recall is the proportion of positive examples ranked above a given rank, while precision is the number of positive boxes above that rank. The AP summarizes the Precision/Recall curve and is defined as the mean (interpolated) precision of the set of eleven equally spaced recall levels. Output bounding boxes are judged as true positives (correct detections) if the overlap ratio (intersection over union or IOU) exceeds 0.50. Detection outputs are assigned to ground truth in the order given by decreasing confidence scores. Duplicated detections of the same object are considered as false detections. The performance over the whole dataset is computed by averaging the APs across all the categories.

The recent and popular MSCOCO challenge \citep{lin2014microsoft} relies on the same principles. The main difference is that the overall performance (mAP) is obtained by averaging the AP obtained with 10 different IOU thresholds between 0.50 and 0.95. The ImageNet Large Scale Visual Recognition Challenge (ILSVRC) also has a detection task in which algorithms have to produce triplets of class labels, bounding boxes and confidence scores. Each image has mostly one dominant object in it. Missing object detections are penalized in the same way as a duplicate detection and the winner of the detection challenge is the one who achieves first place AP on most of the object categories. The challenge also has the {\em Object Localization task}, with a slightly different definition. The motivation is not to  penalize algorithms if one of the detected objects is actually present while not included in the ground-truth annotations, which is not rare due to the size of the dataset and the number of categories (1000). Algorithms are expected to produce 5 class labels (in decreasing order of confidence) and 5 bounding boxes (one for each class label). The error of an algorithm on an image is 0 if one of the 5 bounding boxes is a true positive (correct class label and correct localization according to IOU), 1 otherwise. The error is averaged on all the images of the test set.

Some recent datasets, like DOTA \citep{xia2017dota}, proposed two tasks named as {\em detection on horizontal bounding boxes} and {\em detection on oriented bounding boxes}, corresponding to two different kinds of ground truths (with or without target orientations), no matter how those methods were trained. In some other datasets, the scale of the detection is not important and a detection is counted as a True Positive if its coordinates are close enough to the center of the object. This is the case for the VeDAI dataset \citep{razakarivony2016vehicle}. In the particular case of object detection in 3D point clouds, such as in the KITTI object detection benchmark \citep{geiger2012we}, the criteria is similar to Pascal VOC, except that the boxes are in 3D and the overlap is measured in terms of volume intersection. 

\paragraph{Object detection in videos}
Regarding the detection of objects in videos, the most common practice is to evaluate the performance by considering each frame of the video as being an independent image and averaging the performance over all the frames, as done in the ImageNet VID challenge \citep{russakovsky2015ImageNetLargeScale}. 

It is also possible to move away from the 2D case and to evaluate \textit{video-mAP} based on tubelets IoU, where a tubelet is detected if and only if the mean per frame IoU for every frame in the video is greater than a threshold, $\sigma$, and the tube label is correctly predicted. We take this definition directly from~\citep{gkioxari15tubes}, where they used it to compute mAP and ROC curves at a video-level.

\subsubsection{Other detection tasks}
This survey only covers the methodologies for performance evaluation found in the recent literature. But, beside these common evaluation measures, there are a lot of more specific ones, as object detection can be combined with other complex tasks, \eg, 3D orientation and layout inference in \citep{xiang2012EstimatingAspectLayout}. The reader can refer to the review by \citeauthor{mariano2002PerformanceEvaluationObject} \cite{mariano2002PerformanceEvaluationObject} to explore this topic. It is also worth mentioning the very recent work of \citeauthor{oksuz2018LocalizationRecallPrecision} \cite{oksuz2018LocalizationRecallPrecision} which proposes a novel metric providing richer and more discriminative information than AP, especially with respect to the localization error.

We have decided to orient this survey mainly on bounding boxes tasks even if there is a tendency to move away from this task considering the performances of the modern deep learning methods that already approach human accuracy on some datasets. The reason of this choice are numerous. First of all, historically speaking bounding boxes were one of the first object detection task and thus there is already a body of literature on this topic that is immense. Secondly, not all the datasets provide annotation down to the level of pixels. In aerial imagery for instance most of the datasets are only bounding boxes. It is also the case for some pedestrian detection datasets. Instance segmentation level annotations are still costly for the moment, even with the recent development of annotator friendly algorithms (\eg~\citep{kundu2017polygon, maninis2018deep}) that offer pixel level annotations at the expense of a few user clicks. Maybe in the future all datasets will contain annotations down to the level of pixels but it is not yet the case. Even when one has pixel-level annotations for tasks like instance segmentation, which is becoming the standard, bounding boxes are needed from the detector to distinguish between two instances of the same class, which explains that most modern instance segmentation pipelines like~\citep{he2017mask} have a bounding box branch. Therefore, metrics evaluating the bounding boxes from the models are still relevant in that case. One could also make the argument that bounding boxes are more robust annotations because they are less sensitive to the annotator noise but it is debatable. For all of these reasons the rest of this survey will tackle mainly bounding boxes and associated tasks.

\subsection{From Hand-crafted to Data Driven Detectors} \label{sec:intro_handcraftedtocnn}
While the first object detectors initially relied on mechanisms to align a 2D/3D model of the object on the image using simple features, such as edges \citep{lin2007hierarchical}, key-points \citep{lowe1999object} or templates \citep{pentland1994view}, the arrival of Machine Learning (ML) was the first revolution which had shaken up the area. One of the most popular ML algorithms used for object detection was boosting, \eg, \citep{schneiderman2004object}) or Support Vector Machines, \eg \citep{dalal2005histograms}. This first wave of ML-based detectors were all based on hand-crafted (engineered) visual features processed by classifiers or regressors. These hand-crafted features were as diverse as Haar Wavelets \citep{viola2005detecting}, edgelets \citep{wu2005detection}, shapelets \citep{sabzmeydani2007detecting}, histograms of oriented gradient \citep{dalal2005histograms}, bags-of-visual-words \citep{lampert2008beyond}, integral histograms \citep{porikli2005integral}, color histograms \citep{walk2010new}, covariance descriptors \citep{tuzel2008pedestrian}, linear binary patterns \citeauthor{wang2009lbp} \cite{wang2009lbp}, or their combinations \citep{enzweiler2011multilevel}. One of the most popular detectors before the DCNN revolution was the Deformable Part Model of \citeauthor{Felzenszwalb2010Object} \cite{Felzenszwalb2010Object} and its variants, \eg \citep{sadeghi201430HzObjectDetectiona}.

This very rich literature on visual descriptors has been wiped out in less than five years by Deep Convolutional Neural Networks, which is a class of deep, feed-forward artificial neural networks. DCNNs are inspired by the connectivity patterns between neurons of the human visual cortex and use no pre-processing as the network learns itself the filters previously hand-engineered by traditional algorithms, making them independent from prior knowledge and human effort. They are said to be {\em end-to-end trainable} and solely rely on the training data. This leads to their major disadvantage of requiring copious amounts of data. The first use of ConvNets for detection and localization goes back to the early 1990s for faces \citep{vaillant1994original}, hands \citep{nowlan1995convolutional} and multi-character strings \citep{matan1992reading}. Then in 2000s they were used in text \citep{delakis2008text}, face \citep{garcia2002neural,osadchy2007synergistic} and pedestrians \citep{sermanet2013pedestrian} detection. 

However, the merits of DCNN for object detection was generated in the community only after the seminal work of \citeauthor{krizhevsky2012imagenet} \cite{krizhevsky2012imagenet} and \citeauthor{sermanet2013overfeat} \cite{sermanet2013overfeat} on the challenging ImageNet dataset. \citeauthor{krizhevsky2012imagenet} \cite{krizhevsky2012imagenet} were the first to demonstrate localization through DCNN in the ILSVRC 2012 localization and detection tasks. Just one year later \citeauthor{sermanet2013overfeat} \cite{sermanet2013overfeat} were able to describe how the DCNN can be used to locate and detect objects instances. They won the ILSVRC 2013 localization and detection competition and also showed that combining the classification, localization and detection tasks can simultaneously boost the performance of all tasks.

The first DCNN-based object detectors applied a fine-tuned classifier on each possible location of the image in a sliding window manner \citep{oquab2015isobject}, or on some specific regions of interest \citep{girshick2014rich}, through a region proposal mechanism. \citeauthor{girshick2014rich} \cite{girshick2014rich} treated each region proposal as a separate classification and localization task. Therefore, given an arbitrary region proposal, they deformed it to a warped region of fixed dimensions. DCNN are used to extract a fixed-length feature vector from each proposal respectively and then category-specific linear SVMs were used to classify them. Since it was a region based CNN they called it R-CNN. Another important contribution was to show the usability of transfer learning in DCNN. Since data is scarce, supervised pre-training on an auxiliary task can lead to a significant boost to the performance of domain specific fine-tuning. \citeauthor{sermanet2013overfeat} \cite{sermanet2013overfeat}, \citeauthor{girshick2014rich} \cite{girshick2014rich} and \citeauthor{oquab2015isobject} \cite{oquab2015isobject} were among the first authors to show that DCNN can lead to dramatically higher object detection performance on ImageNet detection challenge \citep{deng2009imagenet} and PASCAL VOC \citep{everingham2010pascal} respectively as compared to previous state-of-the-art systems based on HOG \citep{dalal2005histograms} or SIFT \citep{lowe2004distinctive}.

Since most prevalent DCNN had to use a fixed size input, because of the fully connected layers at the end of the network, they had to either warp or crop the image to make it fit into that size. \citeauthor{he2015spatial} \cite{he2015spatial} came up with the idea of aggregating feature maps of the final convolutional layer. Thus, the fully connected layer at the end of the network gets a fixed size input even if the input images in the dataset are of varying sizes and aspect ratios. This helped reduce overfitting, increased robustness and improved the generalizability of the existing models. Compared to R-CNN which used one forward pass per proposal to generate the feature map, the methodology proposed by \citep{he2015spatial} allowed to share computation among all the proposals and do just one forward pass for the whole image and then select the region from the final feature map according to the regions proposed. This naturally increased the speed of the network by over one hundred times.

All the previous approaches train the network in multistage pipelines are complex, slow and inelegant. They include extracting features through CNNs, classifying through SVMs and finally fitting bounding box regressors. Since, each task is handled separately, convolutional layers cannot take advantage of end-to-end learning and bounding box regression. \citeauthor{girshick2015fast} \cite{girshick2015fast} helped alleviate this problem by streamlining all the tasks in a single model using a multitask loss. As we will explain later, this not only improved upon the accuracy but also made the network run faster at test time. 

\subsection{Overview of Recent Detectors}
\label{sec:intro_recent_methods}
The foundations of the DCNN based object detection, having been laid out, it allowed the field to mature and move further away from classical methods. The fully-convolutional paradigm glimpsed in \citep{sermanet2013overfeat} gained more traction every day in the community.

When \citeauthor{ren2015faster} \cite{ren2015faster} successfully replaced the only component of Fast R-CNN that still relied on non-learned heuristics by inventing RPN (Region Proposal Networks), it put the last nail in the coffin of traditional object detection and started the age of completely end-to-end architectures. Specifically, the anchor mechanism, developed for the RPN, was here to stay. This grid of fixed a-priori (or anchors), not necessarily corresponding to the receptive field of the feature map pixel they lied on, created a framework for fully-convolutional classification and regression and is used nowadays by most pipelines like \citep{liu2016ssd} or \citep{lin2017focal}, to cite a few. 

These conceptual changes make the detection pipelines far more elegant and efficient than their counterparts when dealing with big training sets. However, it comes at a cost. The resulting detectors become complete black boxes, and, because they are more prone to overfitting, they require more data than ever.

\citep{ren2015faster} and its other double stage variants are now the go-to methods for objects detection and will be thoroughly explored in Sec. \ref{sec:double_stages}. Although this line of work is now prominent, other choices were explored all based on fully-convolutional architectures. 

Single-stage algorithms that were completely abandoned since \citeauthor{viola2005detecting} \cite{viola2005detecting} have now become reasonable alternatives thanks to the discriminative power of the CNN features. \citeauthor{redmon2016you} \cite{redmon2016you} first showed that the simplest architectural design could bring unfathomable speed with acceptable performances. \citeauthor{liu2016ssd} \cite{liu2016ssd} sophisticated the pipeline by using anchors at different layers while making it faster and more accurate than \citeauthor{redmon2016you} \cite{redmon2016you}. These two seminal works gave birth to a considerable amount of literature on single stage methods that we will cover in Sec. \ref{sec:single_stage}. Boosting and Deformable part-based models, that were once the norm, have yet to make their comebacks into the mainstream. However, some recent popular works used close ideas like \citeauthor{dai2017deformable} \cite{dai2017deformable} and thus these approaches will also be discussed in the survey sections \ref{sec:boost} and \ref{sec:dpm}.

The fully-convolutional nature of these new dominant architectures allows all kinds of implementation tricks during training and at inference time that will be discussed at the end of the next section. However, it makes the subtle design choices of the different architectures something of a dark art to the newcomers. 

The goal of the rest of the survey is to provide a complete view of this new landscape while giving the keys to understand the underlying principles that guide interesting new architectural ideas. Before diving into the subject, the survey starts by reminding the readers about the object detection task and the metrics associated with it.

\bigskip
\noindent After introducing the topic and touching upon some general information, next section will get right into the heart of object detection by presenting the designs of recent deep learning based object detectors. 

%% file: article_detector_design.tex
\section{On the Design of Modern Deep Detectors \label{sec:detector_design}}
\begin{figure*}[!htp]
\centering
    \includegraphics[width=\textwidth]{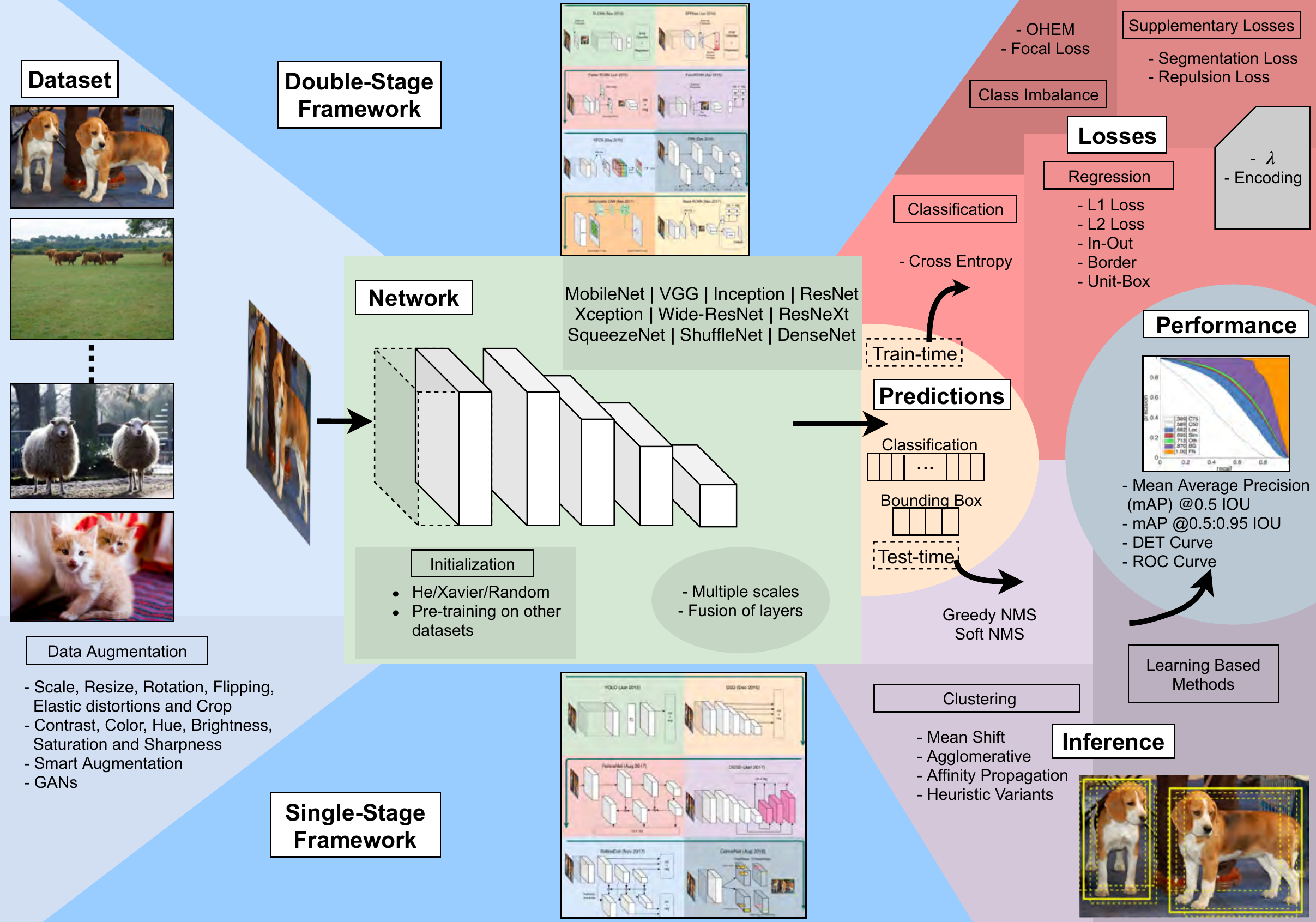}
    \caption{A \textit{map of object detection} pipeline along with various options available in different parts. Images are taken from a \textit{Dataset} and then fed to a \textit{Network}. They use one of the \textit{Single-Stage} (see Figure \ref{fig:single_Stage}) or \textit{Double-Stage Framework} (see Figure \ref{fig:double_Stage}) to make \textit{Predictions} of the probabilities of each class and their bounding boxes at each spatial location of the feature map. During training these predictions are fed to \textit{Losses} and during testing these are fed to \textit{Inference} for pruning. Based on the final predictions, a \textit{Performance} is evaluated against the ground-truths. All the ideas are referenced in the text. Best viewed in color. 
    \label{fig:map}}
\end{figure*}
Here we analyze, investigate and dissect the current state-of-the-art models and the intuition behind their approaches. We can divide the whole detection pipeline into three major parts. The first part focuses on the arrangement of convolutional layers to get proposals (if required) and box predictions. The second part is about setting various training hyper-parameters, deciding upon the losses,  \etc to make the model converge faster. The third part's center of attention will be to know various approaches to refine the predictions from the converged model(s) at test time and therefore get better detection performances. The first part has been able to get the attention of most of the researchers and second and third part not so much. To give a clear overview of all the major components and popular options available in them, we present a map of object detection pipeline in Figure \ref{fig:map}.

Most of the ideas from the following sub-sections have achieved top accuracy on the challenging MS COCO \citep{lin2014microsoft} object detection challenge and PASCAL VOC \citep{everingham2010pascal} detection challenge or on some other very challenging datasets.

\subsection{Architecture of the Networks}
The architecture of the DCNN object detectors follows a Lego-like construction pattern based on chaining different building blocks.
The first part of this Section will focus on what researchers call the {\em backbone} of the DCNN, meaning the feature extractor from which the detector draws its discriminative power. We will then tackle diverse arrangements of increasing complexity found in DCNN detectors: from single stage to multiple stages methods. Finally, we will talk about the Deformable Part Models and their place in the deep learning landscape.

\subsubsection{Backbone Networks}
\label{sec:backbone}
A lot of deep neural networks originally designed for classification tasks have been adopted for the detection task as well. And a lot of modifications have been done on them to adapt for the additional difficulties encountered. The following discussion is about these networks and the modifications in question.

\paragraph{Backbones:}
Backbone networks play a major role in object detection models. \citeauthor{huang2017speed} \cite{huang2017speed} partially confirmed the common observation that, as the classification performance of the backbone increases on ImageNet classification task \citep{russakovsky2015ImageNetLargeScale}, so does the performance of object detectors based on those backbones. It is the case at least for popular double-stage detectors like Faster-RCNN \citep{ren2015faster} and R-FCN \citep{dai2016r} although for SSD \citep{liu2016ssd} the object detection performance remains around the same (see the following Sections for details about these 3 architectures). 

However, as the size of the network increases, the inference and the training become slower and require more data. The most popular architectures in increasing order of inference time are MobileNet \citep{howard2017mobilenets}, VGG \citep{simonyan2014very}, Inception \citep{szegedy2015going,Ioffe_2017_nips,szegedy2015RethinkingInceptionArchitecture}, ResNet \citep{he2016deep}, Inception-ResNet \citep{szegedy2017inception}, \etc All of the above architectures were first borrowed from the classification problem with little or no modification. 

\begin{figure*}[]
\centering
    \includegraphics[height=7.5cm,width=\textwidth]{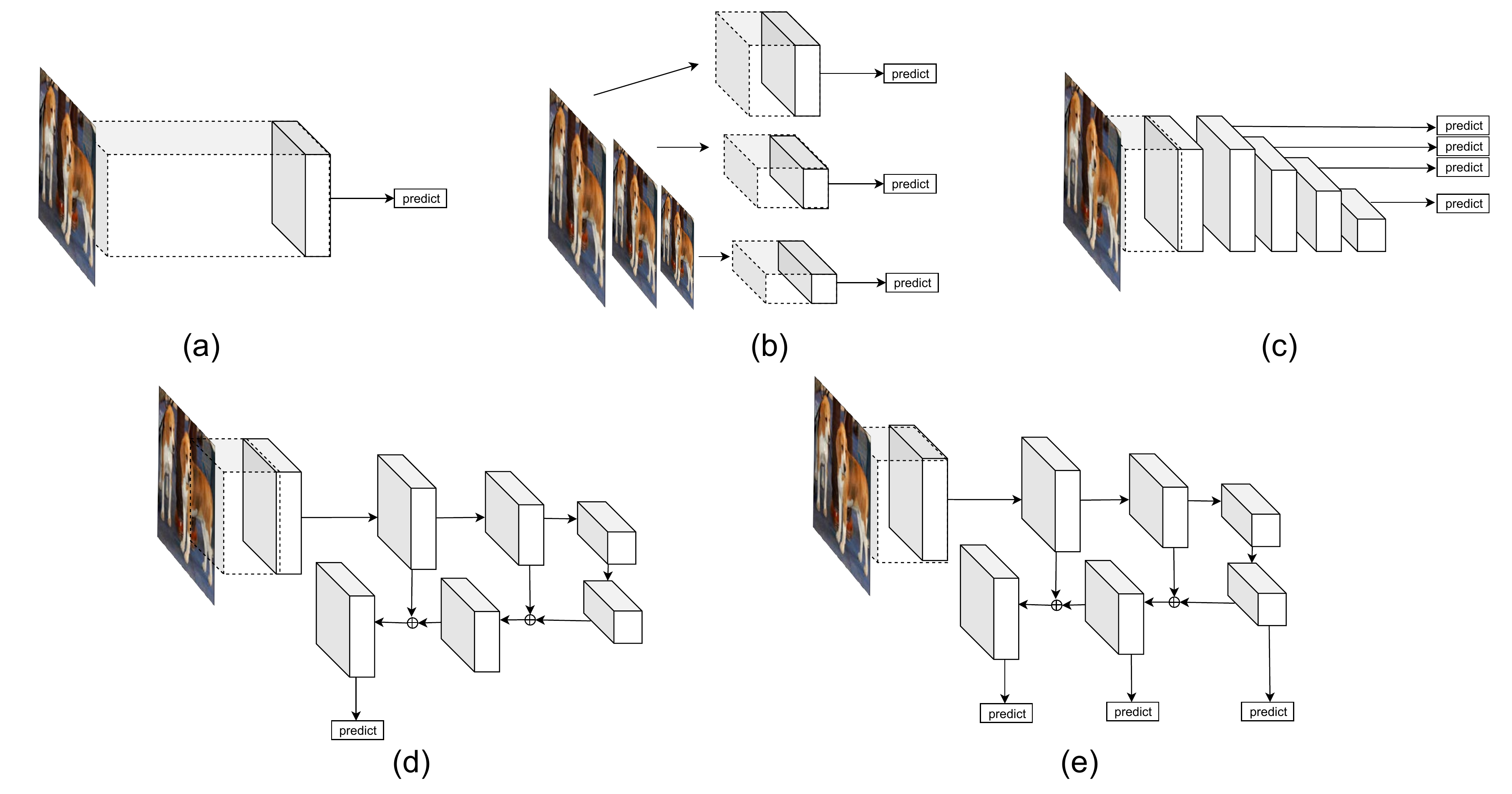}
    \caption{An illustration of how the backbones can be modified to give predictions at multiple scales and through fusion of features. (a) An unmodified backbone. (b) Predictions obtained from different scales of image. (c) Feature maps added to the backbone to get predictions at different scales. (d) A top down network added in parallel to backbone. (e) Top down network along with predictions at different scales.
    \label{fig:scale_and_fusion}}
\end{figure*}

Some other backbones used in object detectors which were not included in the analysis of \citep{huang2017speed} but have given state-of-the-art performances on ImageNet \citep{deng2009imagenet}, or COCO~\citep{lin2014microsoft} detection tasks are Xception \citep{chollet2016xception}, DarkNet \citep{redmon2016YOLO9000BetterFaster}, Hourglass \citep{newell2016stacked}, Wide-Residual Net \citep{zagoruyko2016wide,lee2017wide}, ResNeXt \citep{xie2017aggregated}, DenseNet \citep{huang2017densely}, Dual Path Networks \citep{chen2017dual} and Squeeze-and-Excitation Net \citep{hu2017squeeze}. The recent DetNet \citep{li2018DetNetBackboneNetwork}, proposed a backbone network, is designed specifically for high performance detection. It avoided large down-sampling factors present in classification networks. Dilated Residual Networks \citep{yu2017DilatedResidualNetworks} also worked with similar motivations to extract features with fewer strides. SqueezeNet \citep{iandola2016squeezenet} and ShuffleNet \citep{DBLP:journals/corr/ZhangZLS17} choose instead to focus on speed. More information for networks focusing on speed can be found in Section~\ref{subsec:fastandlowpower}.

Adapting the mentioned backbones to the inherent multi-scale nature of object detection is a challenge, we will give in the following paragraph examples of commonly used strategies.

\paragraph{Multi-scale detections:}
Papers \citep{cai2016unified,li2017scale,yang2016exploit} made independent predictions on multiple feature maps to take into account objects of different scales. The lower layers with finer resolution have generally been found better for detecting small objects than the coarser top layers. Similarly, coarser layers are better for the bigger objects. \citeauthor{liu2016ssd} \cite{liu2016ssd} were the first to use multiple feature maps for detecting objects. Their method has been widely adopted by the community. Since final feature maps of the networks may not be coarse enough to detect sizable objects in large images, additional layers are also usually added. These layers have a wider receptive field.

\paragraph{Fusion of layers:}
In object detection, it is also helpful to make use of the context pixels of the object \citep{zeng2017crafting,zagoruyko2016multipath,gidaris2015object}. One interesting argument in favor of fusing different layers is it integrates information from different feature maps with different receptive fields, thus it can take help of surrounding local context to disambiguate some of the object instances. 

Some papers \citep{chen2017weaving,fu2017dssd,jeong2017enhancement,lee2017residual,zheng2018extend} have experimented with fusing different feature layers of these backbones so that the finer layers can make use of the context learned in the coarser layers. \citet{lin2017focal,lin2017feature, shrivastava2016beyond,woo2017stairnet} took one step ahead and proposed a whole additional top-down network in addition to standard bottom-up network connected through lateral connections. The bottom-top network used can be any one of the above mentioned. While \citeauthor{shrivastava2016beyond} \cite{shrivastava2016beyond} used only the finest layer of top-down architecture for detection, Feature Pyramid Network (FPN) \citep{lin2017focal} and RetinaNet \citep{lin2017feature} used all the layers of top-down architecture for detection. FPN used the feature maps thus generated in a two-stage detector fashion while RetinaNet used them in a single-stage detector fashion (See Section~\ref{sec:double_stages} and Section~\ref{sec:single_stage} for more details). FPN \citep{lin2017feature} has been a part of the top entries in MS COCO 2017 challenge. An illustration of multiple scales and fusion of layers is shown in Figure \ref{fig:scale_and_fusion}.

Now that we have seen how to best use the feature maps of the object detectors backbones we can explore the architectural details of the different major players in DCNN object detection, starting with the most immediate methods: single-stage detectors.

\subsubsection{Single Stage Detectors}
\label{sec:single_stage}
The two most popular approaches in single stage detection category are YOLO \citep{redmon2016you} and SSD \citep{liu2016ssd}. In this Section we will go through their basic functioning, some upsides and downsides of using these two approaches and further improvements proposed on them.

\paragraph{YOLO:}
\citeauthor{redmon2016you} \cite{redmon2016you} presented for the first time a single stage method for object detection where raw image pixels were converted to bounding box coordinates and class probabilities and can be optimized end-to-end directly. This allowed to directly predict boxes in a single feed-forward pass without reusing any component of the neural network or generating proposals of any kind, thus speeding up the detector.

They started by dividing the image into a $S \times S$ grid and assuming B bounding boxes per grid. Each cell containing the center of an object instance is responsible for the detection of that object. Each bounding box predicts 4 coordinates, objectness and class probabilities. This reframed the object detection as a regression problem. To have a receptive field cover that covers the whole image they included a fully connected layer in their design towards the end of the network.

\begin{figure}[htp]
\centering
    \includegraphics[width=0.4\textwidth, height=0.3\textwidth]{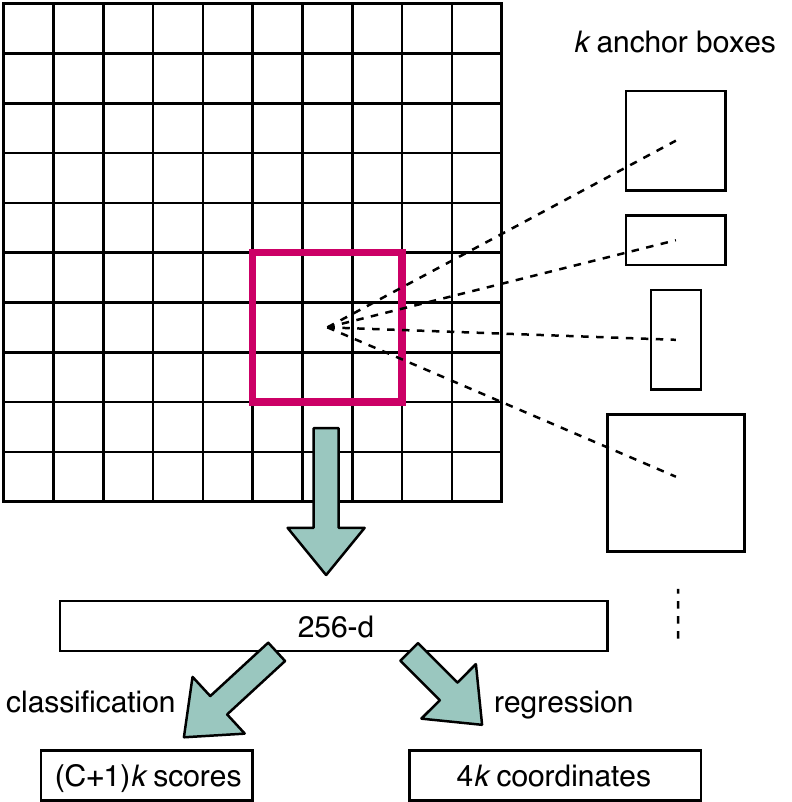}
    \caption{The workings of anchors. $k$ anchors are declared at each spatial location of the final feature map(s). Classification score for each class (including background) is predicted for each anchor. Regression coordinates are predicted only for anchors having an overlap greater than a pre-decided threshold with the ground-truth. For the special case of predicting objectness, $C$ is set to one. This idea was introduced in \cite{ren2015faster}.
    \label{fig:anchor}}
\end{figure}

\paragraph{SSD:}
\citeauthor{liu2016ssd} \cite{liu2016ssd}, inspired by the Faster-RCNN architecture, used reference boxes of various sizes and aspect ratios to predict object instances (Figure \ref{fig:anchor}) but they completely got rid of the region proposal stage (discussed in the following Section). They were able to do this by making the whole network work as a regressor as well as a classifier. During training, thousands of default boxes corresponding to different anchors on different feature maps learned to discriminate between objects and background. They also learned to directly localize and predict class probabilities for the object instances. This was achieved with the help of a multitask loss. Since, during inference time a lot of boxes try to localize the objects, generally a post-processing step like Greedy NMS is required to suppress duplicate detections.

In order to accommodate objects of all the sizes they added additional convolutional layers to the backbone and used them, instead of a single feature map, to improve the performance. This method was later applied to approaches related to two-stage detectors too \citep{lin2017feature}.

\paragraph{Pros and Cons:}
Oftentimes single stage detectors do not give as good performance as the double-stage ones, but they are a lot faster \citep{huang2017speed} although some double-stages detectors can be faster than single-stages due to architectural tricks and modern single-stage detectors outperform the older multi-stages pipelines. 

The various advantages of YOLO strategy are that it is extremely fast, with 45 to 150 frames per second. It sees the entire image as opposed to region proposal based strategies which is helpful for encoding contextual information and it learns generalizable representations of objects. But it also has some obvious disadvantages. Since each grid cell has only two bounding boxes, it can only predict at most two objects in a grid cell. This is particularly inefficient strategy for small objects. It struggles to precisely localize some objects as compared to two stages. Another drawback of YOLO is that it uses coarse feature map at a single scale only. 

To address these issues, SSD used a dense set of boxes and considered predictions from various feature maps instead of one. It improved upon the performance of YOLO. But since it has to sample from these dense set of detections at test time it gives lower performance on MS COCO dataset as compared to two-stage detectors. The two-stage object detectors get a sparse set of proposals on which they have to perform predictions.

\begin{figure*}
\centering
    \includegraphics[width=\textwidth]{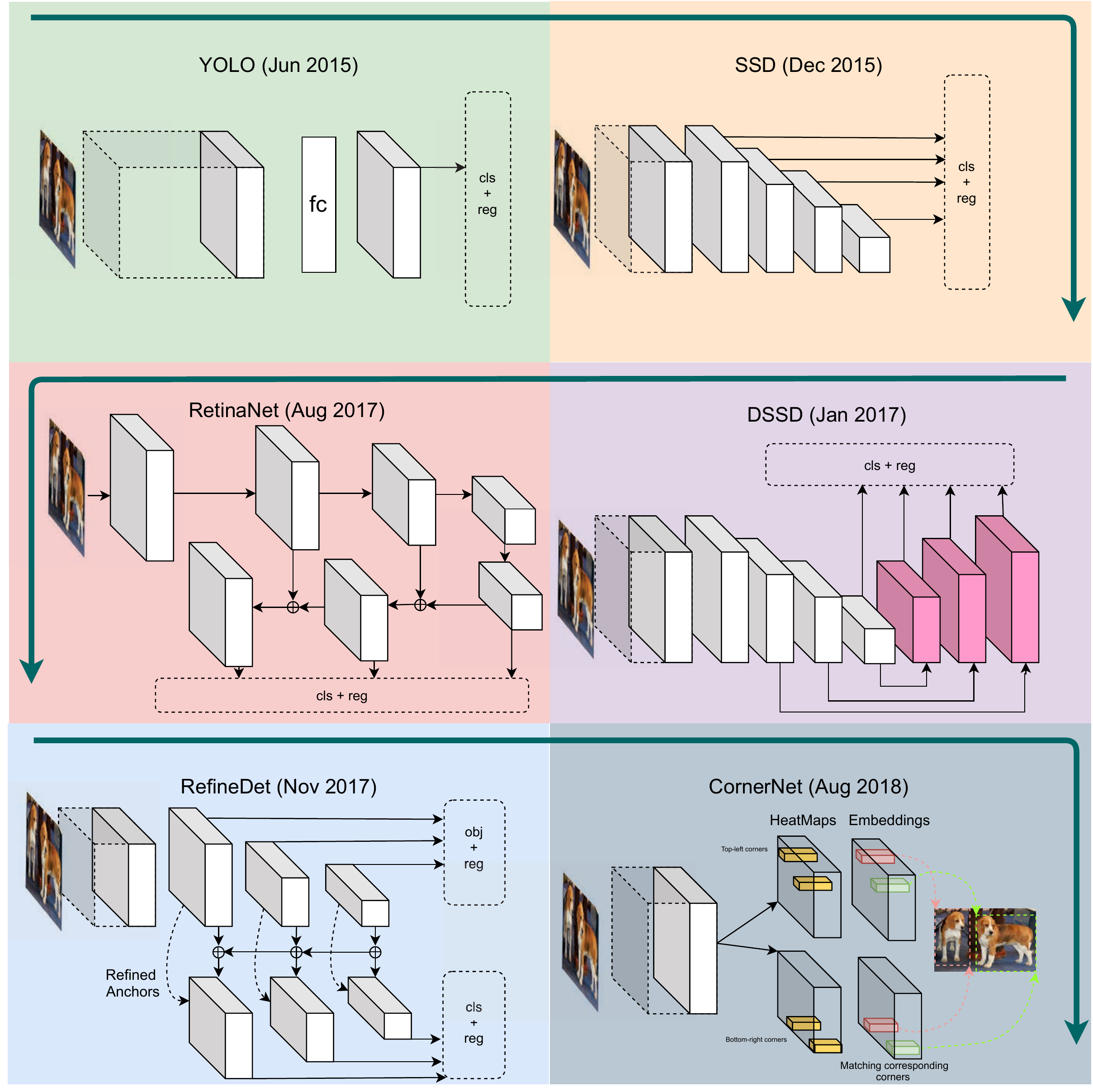}
    \caption{Evolution of single stage detectors over the years. Major advancements in chronological order are YOLO \cite{redmon2016you}, SSD \cite{liu2016ssd}, DSSD \cite{fu2017dssd}, RetinaNet \cite{lin2017focal}, RefineDet \cite{zhang2018single} and CornerNet \cite{law2018cornernet}. Cuboid boxes, solid rectangular box, dashed rectangular boxes and arrows represent convolutional layer, fully connected layer, predictions and flow of features respectively. \textit{obj, cls} and \textit{reg} stand for objectness, classification and regression losses. Best viewed in color.
    \label{fig:single_Stage}}
\end{figure*}

\paragraph{Further improvements:}
\citeauthor{redmon2016YOLO9000BetterFaster} \cite{redmon2016YOLO9000BetterFaster} and \citeauthor{redmon2018YOLOv3IncrementalImprovement} \cite{redmon2018YOLOv3IncrementalImprovement} suggested a lot of small changes in versions 2 and 3 of the YOLO method. The changes like applying batch normalization, using higher resolution input images, removing the fully connected layer and making it fully convolutional, clustering box dimensions, location prediction and multi-scale training helped to improve performance while a custom network (DarkNet) helped to improve speed.

Many further developments by many researchers have been proposed on Single Shot MultiBox Detector. The major advancements over the years have been illustrated in Figure \ref{fig:single_Stage}. Deconvolutional Single Shot Detector (DSSD) \citep{fu2017dssd}, instead of the element-wise sum, used a deconvolutional module to increase the resolution of top layers and added each layer, through element-wise products to previous layer. Rainbow SSD \citep{jeong2017enhancement} proposed to concatenate features of shallow layers to top layers by max-pooling as well as features of top layers to shallow layers through deconvolution operation. The final fused information increased from few hundreds to 2,816 channels per feature map. RUN \citep{lee2017residual} proposed a 3-way residual block to combine adjacent layers before final prediction. \citeauthor{cao2017FeatureFusedSSDFast} \cite{cao2017FeatureFusedSSDFast} used concatenation modules and element-sum modules to add contextual information in a slightly different manner. \citeauthor{zheng2018extend} \cite{zheng2018extend} slightly tweak DSSD by fusing lesser number of layers and adding extra ConvNets to improve speed as well as performance.

They all improved upon the performance of conventional SSD and they lie within a small range among themselves on Pascal VOC 2012 test set \citep{everingham2010pascal}, but they added considerable amount of computational costs, thus making it little slower. WeaveNet \citep{chen2017weaving} aimed at reducing computational costs by gradually sharing the information from adjacent scales in an iterative manner. They hypothesized that by weaving the information iteratively, sufficient multi-scale context information can be transferred and integrated to current scale.

Recently three strong candidates have emerged for replacing the undying YOLO and SSD variants:
\begin{itemize}
\item RetinaNet \citep{lin2017focal} borrowed the FPN structure but in a single stage setting. It is similar in spirit to SSD but it deserves its own paragraph given its growing popularity based on its speed and performance. The main new advance of this pipeline is the focal loss, which we will discuss in Section~\ref{subsec:losses}. 
\item RefineDet~\citep{zhang2018single} tried to combine the advantages of double-staged methods and single-stage methods by incorporating two new modules in the single stage classic architecture. The first one, the ARM (Anchor Refinement modules), is used in multiple staged detectors' fashion to reduce the search space and also to iteratively refine the localization of the detections. The ODM (Object Detection Module) took the output of the ARM to output fine-grained classification and further improve the localization.
\item CornerNet \citep{law2018cornernet} offered a new approach for object detection by predicting bounding boxes as paired top-left and bottom right keypoints. They also demonstrated that one can get rid of the prominent anchors step while gaining accuracy and precision. They used fully convolutional networks to produce independent score heat maps for both corners for each class in addition to learning an embedding for each corner. The embedding similarities were then used to group them into multiple bounding boxes. It beat its two (less original) competing rivals on COCO.
\end{itemize}

\noindent However, most methods used in competitions until now are predominantly double-staged methods because their structure is better suited for fine-grained classification. It is what we are going to see in the next Section.

\subsubsection{Double Stage Detectors}
\label{sec:double_stages}
The process of detecting objects can be split into two parts: proposing regions \& classifying and regressing bounding boxes. The purpose of the proposal generator is to present the classifier with class-agnostic rectangular boxes which try to locate the ground-truth instances. The classifier, then, tries to assign a class to each of the proposals and further fine-tune the coordinates of the boxes.

\paragraph{Region proposal:} \citeauthor{hosang2014HowGoodAre} \cite{hosang2014HowGoodAre} presented an in-depth review of ten "non-data driven" object proposal methods including Objectness \citep{alexe2010object,alexe2012measuring}, CPMC \citep{carreira2010constrained,carreira2011cpmc}, \citet{endres2010category,endres2014category}, Selective Search \citep{van2011segmentation,uijlings2013selective}, \citeauthor{rahtu2011learning} \cite{rahtu2011learning}, Randomized Prim \citep{manen2013prime}, Bing \citep{cheng2014bing}, MCG \citep{pont2017multiscale}, \citeauthor{rantalankila2014generating} \cite{rantalankila2014generating}, \citeauthor{humayun2014rigor} \cite{humayun2014rigor} and EdgeBoxes \citep{zitnick2014EdgeBoxesLocating} and evaluated their effect on the detector's performance. Also, \citeauthor{xiao2015ComplexityadaptiveDistanceMetric} \cite{xiao2015ComplexityadaptiveDistanceMetric} developed a novel distance metric for grouping two super-pixels in high-complexity scenarios. Out of all these approaches Selective Search and EdgeBoxes gave the best recall and speed. The former is an order of magnitude slower than Fast R-CNN while the latter, which is not as efficient, took as much time as a detector. The bottleneck lied in the region proposal part of the pipeline.

Deep learning based approaches \citep{erhan2014ScalableObjectDetection,szegedy2015ScalableHighQualityObject} had also been used to propose regions but they were not end-to-end trainable for detection and required input images to be of fixed size. In order to address strong localization bias \citep{chen2015ImprovingObjectProposals} proposed a box-refinement method based on the super-pixel tightness distribution. DeepMask \citep{pinheiro2015LearningSegmentObject} and SharpMask \citep{pinheiro2016learning} proposed segmentation based object proposals with very deep networks. \citep{kang2015data} estimated the objectness of image patches by comparing them with exemplar regions from prior data and finding the ones that are most similar to it. 

The next obvious question became apparent. How can deep learning methods be streamlined into existing approaches to give an elegant, simple, end-to-end trainable and fully convolutional model? In the discussion that follows we will discuss two widely adopted approaches in two-stage detectors, pros and cons of using such approaches and further improvements made on them.

\paragraph{R-CNN and Fast R-CNN:} The first modern ubiquitous double-staged deep learning detection method is certainly~\citep{girshick2014rich}. Although it is has now been abandoned due to faster alternatives it is worth mentioning to better understand the next paragraphs. Closer to the traditional non deep-learning methods the first stage of the method is the detection of objects in pictures to reduce the number of false positive of the subsequent stage. It is done using a hierarchical pixel grouping algorithm widely popular in the 2000s called selective search~\citep{van2011segmentation}. Once the search space has been properly narrowed, all regions above a certain score are warped to a fixed size so that a classifier can be applied on top of it. Further fine-tuning on the last layers of the classifier is necessary on the classes of the dataset used (they replace the last layer so that it has the right number of classes) and an SVM is used on top of the fixed fine-tuned features to further refine the localization of the detection. This method was the precursor of all the modern deep learning double-staged methods, in spite of the fact that this first iteration of the method was far from the elegant paradigm used nowadays. Fast R-CNN~\citeauthor{girshick2015fast} \cite{girshick2015fast} from the same author is built on top of this previous work. The author started to refine R-CNN by being one of the first researcher with \citeauthor{he2015spatial} \cite{he2015spatial} to come-up with his own deep-learning detection building block. This differentiable mechanism called RoI-pooling (Region of Interest Pooling) was used for resizing fixed regions (also extracted with selective-search) coming not from the image directly but from the feature computed on the full image, which kept the spatial layout of the original image. Not only did that bring speed-up to the slow R-CNN method (x200 in inference) but it also came with a net gain in performances (around 6 points in mAP).

\begin{figure*}
\centering
    \includegraphics[width=0.55\textwidth]{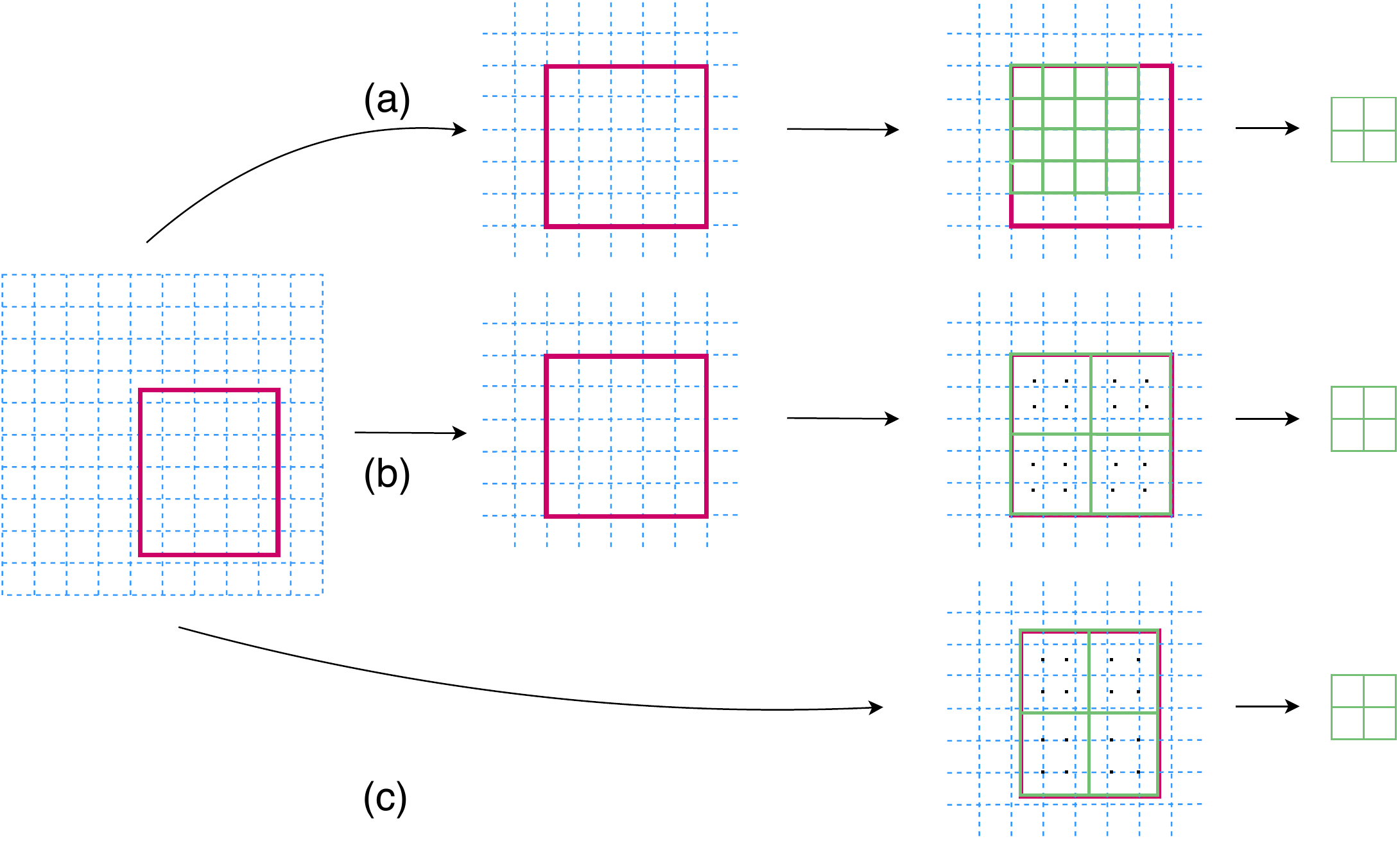}
    \caption{Graphical explanation of RoIPooling, RoIWarping and RoIAlign (actual dimensions of pooled feature map differ). The red box is the predicted output of Region Proposal Network (RPN) and the dashed blue grid is the feature map from which proposals are extracted. (a) RoI Pooling first aligns the proposal to the feature map (first quantization) and then max-pools or average-pools the features (second quantization). Note that some information is lost because the quantized proposal is not an integral multiple of the final map's dimensions. (b) RoI Warping retains the first quantization but deals with second one through bilinear interpolation, calculated from four nearest features, through sampling $N$ points (black dots) for each cell of final map. (c) RoI Align removed both the quantizations by directly sampling $N$ points on the original proposal. $N$ is set to four generally. Best viewed in color.
    \label{fig:roi}}
\end{figure*}

\paragraph{Faster-RCNN:}
The seminal Faster-RCNN paper \citep{ren2015faster} showed that the same backbone architecture used in Fast R-CNN for classification can be used to generate proposals as well. They proposed an efficient fully convolutional data driven based approach for proposing regions called Region Proposal Network (RPN). RPN learned the "objectness" of all instances and accumulated the proposals to be used by the detector part of the backbone. The detector further classified and refined bounding boxes around those proposals. RPN and detector can be trained separately as well as in a combined manner. When sharing convolutional layers with the detector they result in very little extra cost for region proposals. Since it has two parts for generating proposals and detection, it comes under the category of two-stage detectors. 

Faster-RCNN used thousands of reference boxes, commonly known as anchors. Anchors formed a grid of boxes that act as starting points for regressing bounding boxes. These anchors were then trained end-to-end to regress to the ground truth and an objectness score was calculated per anchor. The density, size and aspect ratio of anchors are decided according to the general range of size of object instances expected in the dataset and the receptive field of the associated neuron in the feature map.

RoI Pooling, introduced in \citep{girshick2015fast}, warped the proposals generated by the RPN to fixed size vectors for feeding to the detection sub-network as its inputs. The quantization and rounding operation defining the pooling cells introduced misalignments and actually hurt localization.

\paragraph{R-FCN:}
To avoid running the costly RoI-wise subnetwork in Faster-RCNN hundreds of times, \ie once per proposal, \citeauthor{dai2016r} \cite{dai2016r} got rid of it and shared the convolutional network end to end. To achieve this they proposed the idea of position sensitive feature maps. In this approach each feature map was responsible for outputting score for a specific part, like top-left, center, bottom right, \etc, of the target class. The parts were identified with RoI-Pooling cells which were distributed alongside each part-specific feature map. Final scores were obtained by average voting every part of the RoI from the respective filter. This implementation trick introduced some more translational variance to structures that were essentially translation-invariant by construction. Translational variance in object detection can be beneficial for learning localization representations. Although this pipeline seems to be more precise, it is not always better performance-wise than its Faster R-CNN counterpart. 

From an engineering point of view, this method of Position sensitive RoI-Pooling (PS Pooling) also prevented the loss of information at RoI Pooling stage in Faster-RCNN. It improved the overall inference time speed of two-stage detectors but performed slightly worse.

\paragraph{Pros and Cons:}
RPNs are generally configured to generate nearly 300 proposals to get state-of-the-art performances. Since each of the proposal passed through a head of convolutional layers and fully connected layers to classify the objects and fine tune the bounding boxes, it decreased the overall speed. Although they are slow and not suited to real-time applications, the ideas based on these approaches give one the best performances in the challenging COCO detection task. Another drawback is that \citeauthor{ren2015faster} \cite{ren2015faster} and \citeauthor{dai2016r} \cite{dai2016r} used coarse feature maps at a single scale only. This is not sufficient when objects of diverse sizes are present in the dataset.

\begin{figure*}
\centering
    \includegraphics[width=0.8\textwidth]{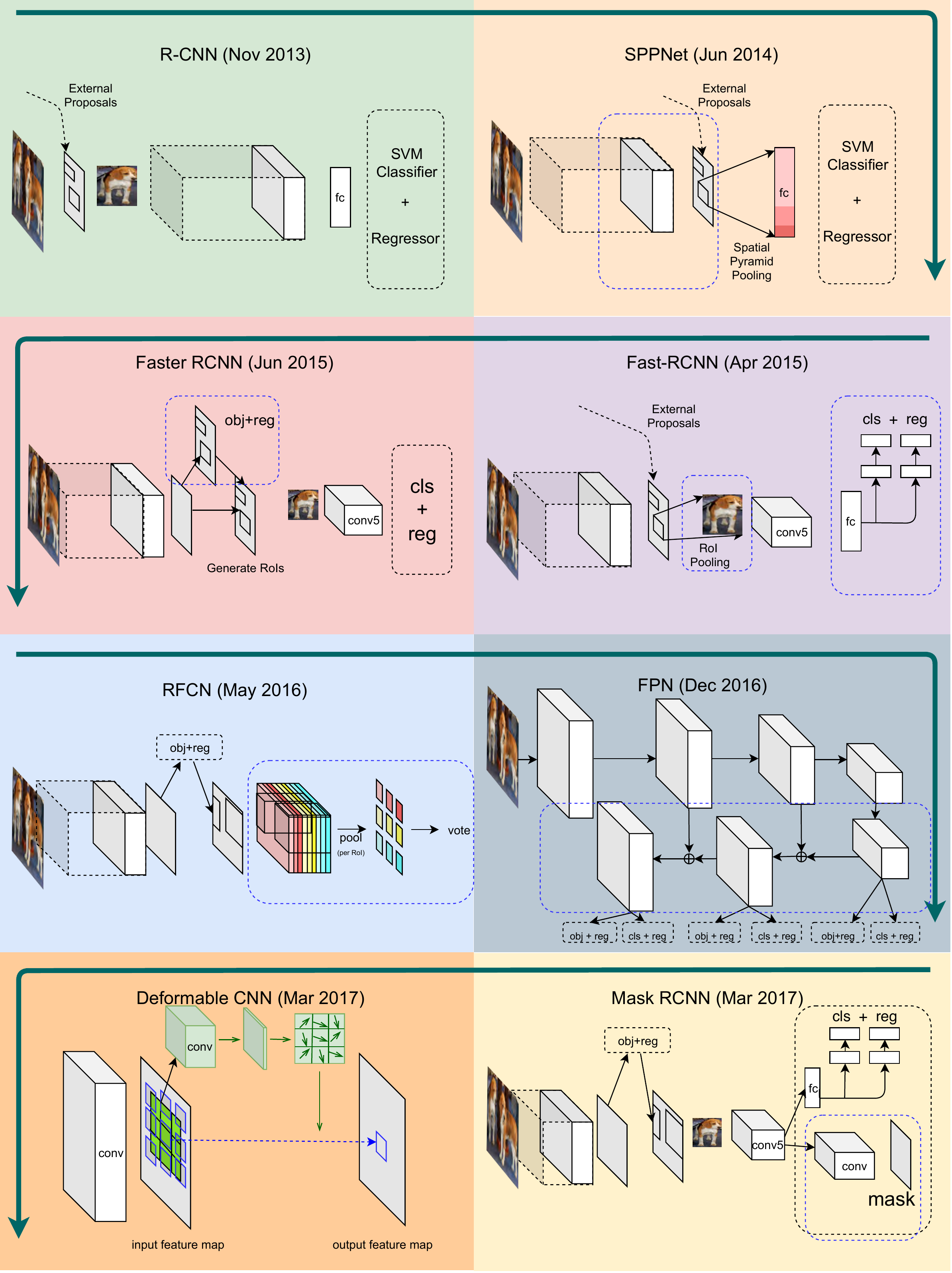}
    \caption{Evolution of double stage detectors over the years. Major advancements in chronological order are R-CNN \cite{girshick2014rich}, SPPNet \cite{he2015spatial}, Fast-RCNN \cite{girshick2015fast}, Faster RCNN \cite{ren2015faster}, RFCN \cite{dai2016r}, FPN \cite{lin2017feature}, Mask RCNN \cite{he2017mask}, Deformable CNN \cite{dai2017deformable} (only the modification is shown and not the entire network). The main idea is marked in dashed blue rectangle wherever possible. Other information is same as in Figure \ref{fig:single_Stage}.
    \label{fig:double_Stage}}
\end{figure*}

\paragraph{Further improvements:}
Many improvements have been suggested on the above methodologies concerning speed, performance and computational efficiency.

DeepBox \citep{kuo2015deepbox} proposed a light weight generic objectness system by capturing semantic properties. It helped in reducing the burden of localization on the detector as the number of classes increased. Light-head R-CNN \citep{li2017LightHeadRCNNDefense} proposed a smaller detection head and thin feature maps to speed up two-stage detectors. \citeauthor{singh2017RFCN300030fpsDecoupling} \cite{singh2017RFCN300030fpsDecoupling} brought R-FCN to 30 fps by sharing position sensitive feature maps across classes. Using slight architectural changes, they were also able to bring the number of classes predicted by R-FCN to 3000 without losing too much speed.

Several improvements have been made to RoI-Pooling. The spatial transformer of \citep{jaderberg2015SpatialTransformerNetworks} used a differentiable re-sampling grid using bilinear interpolation and can be used in any detection pipeline. \citeauthor{chen2016SupervisedTransformerNetwork} \cite{chen2016SupervisedTransformerNetwork} used this for Face detection, where faces were warped to fit canonical poses. \citeauthor{dai2016instance} \cite{dai2016instance} proposed another type of pooling called RoI Warping based on bilinear interpolation. \citeauthor{ma2018ArbitraryOrientedSceneText} \cite{ma2018ArbitraryOrientedSceneText} were the first to introduce a rotated RoI-Pooling working with oriented regions (More on oriented RoI-Pooling can be found in Section~\ref{sec:rotation}). Mask R-CNN \citep{he2017mask} proposed RoI Align to address the problem of misalignment in RoI Pooling which used bilinear interpolation to calculate the value of four regularly sampled locations on each cell. It allowed to fix to some extents the alignment between the computed features and the regions they were extracted from. It brought consistent improvements to all Faster R-CNN baselines on COCO. A comparison is shown in Figure \ref{fig:roi}. Recently, \citeauthor{jiang2018acquisition} \cite{jiang2018acquisition} introduced a Precise RoI Pooling based on interpolating not just 4 spatial locations but a dense region, which allowed full differentiability with no misalignments.

\citet{li2016ObjectDetectionEndtoEnd,yu2016RoleContextSelection} also used contextual information and aspect ratios while StuffNet \citep{brahmbhatt2017StuffNetUsingStuff} trained for segmenting amorphous categories such as ground and water for the same purpose. \citeauthor{chen2017SpatialMemoryContext} \cite{chen2017SpatialMemoryContext} made use of memory to take advantage of context in detecting objects. \citeauthor{li2018RFCNAccurateRegionBased} \cite{li2018RFCNAccurateRegionBased} incorporated Global Context Module (GCM) to utilize contextual information and Row-Column Max Pooling (RCM Pooling) to better extract scores from the final feature map as compared to the R-FCN method.

Deformable R-FCN \citep{dai2017deformable} brought flexibility to the fixed geometric transformations at the Position sensitive RoI-Pooling stage of R-FCN by learning additional offsets for each spatial sampling location using a different network branch in addition to other tricks discussed in Section~\ref{sec:dpm}. \citeauthor{lin2017feature} \cite{lin2017feature} proposed to use a network with multiple final feature maps with different coarseness to adapt to objects of various sizes. \citeauthor{zagoruyko2016multipath} \cite{zagoruyko2016multipath} used skip connections with the same motivation. Mask-RCNN \citep{he2017mask} in addition to RoI-align added a branch in parallel to the classification and bounding box regression for optimizing the segmentation loss. Additional training for segmentation led to an improvement in the performance of object detection task as well. The advancements of two stage detectors over the years is illustrated in Figure \ref{fig:double_Stage}.

\bigskip\noindent The double-staged methods have now by far attained supremacy over best performing object detection DCNNs. However, for certain applications two-stage methods are not enough to get rid of all the false positives.

\subsubsection{Cascades}
\label{sec:boost}
Traditional one-class object detection pipelines resorted to boosting like approaches for improving the performance where uncorrelated weak classifiers (better than random chance but not too correlated with the true predictions) are combined to form a strong classifier. With modern CNNs, as the classifiers are quite strong, the attractiveness of those methods has plummeted. However, for some specific problems where there are still too many false positives, researchers still find it useful. Furthermore, if the weak CNNs used are very shallow it can also sometimes increase the overall speed of the method. 

One of the first ideas that were developed was to cascade multiple CNNs. \citeauthor{li2015convolutional} \cite{li2015convolutional} and \citeauthor{yang2016MultiScaleCascadeFully} \cite{yang2016MultiScaleCascadeFully} both used a three-staged approach by chaining three CNNs for face detection. The former approach scanned the image using a $12 \times 12$ patch CNN to reject 90\% of the non-face regions in a coarse manner. The remaining detections were offset by a second CNN and given as input to a $24 \times 24$ CNN that continued rejecting false positives and refining regressions. The final candidates were then passed on to a $48 \times 48$ classification network which output the final score. The latter approach created separate score maps for different resolutions using the same FCN on different scales of the test image (image pyramid). These score maps were then up-sampled to the same resolution and added to create a final score map, which was then used to select proposals. Proposals were then passed to the second stage where two different verification CNNs, trained on hard examples, eradicated the remaining false positives. The first one being a four-layer FCN trained from scratch and the second one an AlexNet \citep{krizhevsky2012imagenet} pre-trained on ImageNet.
 
All the approaches mentioned in the last paragraph are ad hoc: the CNNs are independent of each other, there is no overall design, therefore, they could benefit from integrating the elegant zooming module that is the RoI-Pooling. The RoI-Pooling can act like a glue to pass the detections from one network to the other, while doing the down-sampling operation locally. 
\citeauthor{dai2016instance} \cite{dai2016instance} used a Mask R-CNN like structure that first proposed bounding boxes, then predicted a mask and used a third stage to perform fine grained discrimination on masked regions that are RoI-Pooled a second time.

\citet{ouyang2017LearningChainedDeep,wang2017EvolvingBoxesFast} optimized in an end-to-end manner a Faster R-CNN with multiple stages of RoI-Pooling. Each stage accepted only the highest scored proposals from the previous stage and added more context and/or localized the detection better. Then additional information about context was used to do fine grained discrimination between hard negatives and true positives in \citep{ouyang2017LearningChainedDeep}, for example.
On the contrary, \citeauthor{zhang2016FasterRCNNDoing} \cite{zhang2016FasterRCNNDoing} showed that for pedestrian detection RoI-Pooling, too coarse a feature map actually hurts the result. This problem has been alleviated by the use of feature pyramid networks with higher resolution feature maps. Therefore, they used the RPN proposals of a Faster R-CNN in a boosting pipeline involving a forest (\citeauthor{tang2017VehicleDetectionAerial} \cite{tang2017VehicleDetectionAerial} acted similarly for small vehicle detection). 

\citeauthor{yang2016exploit} \cite{yang2016exploit}, aware of the problem raised by \citeauthor{zhang2016FasterRCNNDoing} \cite{zhang2016FasterRCNNDoing}, used RoI-Pooling on multiple scaled feature maps of all the layers of the network. The classification function on each layer was learned using the weak classifiers of AdaBoost and then approximated using a fully connected neural network.
While all the mentioned pipelines are hard cascades where the different classifiers are independent, it is sometimes possible to use a soft cascade where the final score is a linear weighted combination of the scores given by the different weak classifiers like in \citeauthor{angelova2015real} \cite{angelova2015real}. They used 200 stages (instead of 2000 stages in their baseline with AdaBoost \citep{benenson2012pedestrian}) to keep recall high enough while improving precision. To save computations that would be otherwise unmanageable, they terminated the computations of the weighted sum whenever the score for a certain number of classifiers fell under a specified threshold (there are, therefore, as many thresholds to learn as there are classifiers). These thresholds are then really important because they control the trade-off between speed, recall and precision.

All the previous works in this Section involved a small fixed number of localization refinement steps, which might cause proposals to be not perfectly aligned with the ground truth, which in turn might impact the accuracy. That is why lots of work proposed iterative bounding box regression (while loop on localization refinement until condition is reached).
\citeauthor{najibi2016GCNNIterativeGrid} \cite{najibi2016GCNNIterativeGrid}, \citeauthor{rajaram2016RefineNetIterativeRefinement} \cite{rajaram2016RefineNetIterativeRefinement} started with a regularly spaced grid of sparse pyramid boxes (only 200 non-overlapping in \citeauthor{najibi2016GCNNIterativeGrid} \cite{najibi2016GCNNIterativeGrid} whereas, \citeauthor{rajaram2016RefineNetIterativeRefinement} \cite{rajaram2016RefineNetIterativeRefinement} used all Faster R-CNN anchors on the grid) that were iteratively pushed towards the ground truth according to the feature representation obtained from RoI-Pooling the current region. An interesting finding was that even if the goal was to use as many refinement steps as necessary if the seed boxes or anchors span the space appropriately, regressing the boxes only twice can in fact be sufficient \citep{najibi2016GCNNIterativeGrid}. 
Approaches proposed by \citeauthor{gidaris2016AttendRefineRepeat} \cite{gidaris2016AttendRefineRepeat} and \citeauthor{li2017ZoomOutandInNetwork} \cite{li2017ZoomOutandInNetwork} can also be viewed, internally, as iterative regression based methods proposing regions for detectors, such as Fast R-CNN.  

Recently, \citeauthor{cai2018cascade} \citep{cai2018cascade} noticed that when increasing the IoU threshold for a window to be considered positive in the training (to get better quality hypothesis for the next stages), one loses a lot of positive windows. Thus one has to keep using the low 0.5 threshold to prevent overfitting and thus one gets bad quality hypothesis in the next stages. This is true for all the works mentioned in this section that are based on Faster R-CNN (\eg \cite{ouyang2017LearningChainedDeep,wang2017EvolvingBoxesFast}). To combat this effect, they slowly increase the IoU threshold over the stages to get different sets of detectors using the latest stage proposals as input distribution for the next one. With only 3 to 4 stages they consistently improve the quality of a wide range of detectors with an average of 3 points gained \wrt the non-cascaded version. This algorithmic advance is used in most of the winning entries of the 2018 COCO challenge (used at least by the first three teams).

Orthogonal to this approach~\citeauthor{jiang2018acquisition} \cite{jiang2018acquisition} frames the regression of the multi-stage cascade as an optimization problem thus introducing a proxy for a smooth measure of confidence of the bounding box localization. This article among others will be discussed in more details in the Section~\ref{subsec:losses}.

\bigskip\noindent Boosting and multistage ($>2$) methods, we have seen previously, exhibit very different possible combinations of DCNNs. But we thought it would be interesting to still have a Section for a special kind of method that was hinted at in the previous Sections, namely the part-based models, if not for their performances at least for their historical importance.

\subsubsection{Parts-Based Models}
\label{sec:dpm}
Before the reign of CNN methods, the algorithms based on Deformable Parts-based Model (DPM) and HoG features used to win all the object detection competitions. In this algorithm latent (not supervised) object parts were discovered for each class and optimized by minimizing the deformations of the full objects (connections were modeled by springs forces). The whole thing was built on a HoG image pyramid.

When Region based DCNNs started to beat the former champion, researchers began to wonder if it was only a matter of using better features. If this was the case then the region based approach would not necessarily be a more powerful algorithm. 
The DPM was flexible enough to integrate the newer more discriminative CNN features. Therefore, some research works focused in this research direction.

In 2014, \citeauthor{savalle2014DeformablePartModels} \cite{savalle2014DeformablePartModels} tried to get the best of both worlds: they replaced the HoG feature pyramids used in the DPM with the CNN layers. Surprisingly, the performance they obtained, even if far superior to the DPM+HoG baseline, was considerably worse than the R-CNN method. The authors suspected the reason for it was the fixed size aspect ratios used in the DPM together with the training strategy. \citeauthor{girshick2015DeformablePartModels} \cite{girshick2015DeformablePartModels} put more thought on how to mix CNN and DPM by coming up with the distance transform pooling thus bringing the new DPM (DeepPyramidDPM) to the level of R-CNN (even slightly better). \citeauthor{ranjan2015DeepPyramidDeformable} \cite{ranjan2015DeepPyramidDeformable} built on it and introduced a normalization layer that forced each scale-specific feature map to have the same activation intensities. They also implemented a new procedure of sampling optimal targets by using the closest root filter in the pyramid in terms of dimensions. This allowed them to further mimic the HOG-DPM strengths. Simultaneously, \citeauthor{wan2015EndtoendIntegrationConvolutional} \cite{wan2015EndtoendIntegrationConvolutional} also improved the DeepPyramidDPM but failed short compared to the newest version of R-CNN, fine-tuned (R-CNN FT). Therefore, in 2015 it seemed that the DPM based approaches have hit a dead end and that the community should focus on R-CNN type methods.

However, the flexibility of the RoI-Pooling of Fast R-CNN was going to help making the two approaches come together.
\citeauthor{ouyang2015DeepIDNetDeformableDeep} \cite{ouyang2015DeepIDNetDeformableDeep} combined Fast R-CNN to get rid of most backgrounds and a DeepID-Net, which introduced a max-pooling penalized by the deformation of the parts called def-pooling. The combination improved over the state-of-the-art. As we mentioned in Section~\ref{sec:double_stages}, \citeauthor{dai2017deformable} \cite{dai2017deformable} built on R-FCN and added deformations in the Position Sensitive RoI-Pooling: an offset is learned from the classical Position Sensitive pooled tensor with a fully connected network for each cell of the RoI-Pooling thus creating "parts" like features. This trick of moving RoI cells around is also present in \citep{mordan2017DeformablePartbasedFully}, although slightly different because it is closer to the original DPM. \citeauthor{dai2017deformable} \cite{dai2017deformable} even added offsets to convolutional filters cells on Conv-5, which became doable thanks to bilinear interpolation. It, thus, became a truly deformable fully convolutional network. However, \citeauthor{mordan2017DeformablePartbasedFully} \cite{mordan2017DeformablePartbasedFully} got better performances on VOC without it. Several works used deformable R-FCN like \citep{xu2017DeformableConvNetAspect} for aerial imagery that used a different training strategy. However, even if it is still present in famous competitions like COCO, it is less used than its counterparts with fixed RoI-Pooling. It might come back though thanks to recent best performing models like \citep{singh2018analysis} that used \citep{dai2017deformable} as their baseline and selectively back-propagated gradients according to the object size.

\subsection{Model Training}
The next important aspect of the detection model's design is the losses being used to converge the huge number of weights and the hyper-parameters that must be conducive to this convergence. Optimizing for a wrongfully crafted loss may actually lead the model to diverge instead. Choosing incorrect hyper-parameters, on the one hand, can stagnate the model, trap it in a local optima or, on the other hand, over-fit the training data (causing poor generalizations). Since DCNNs are mostly trained with mini-batch SGD (see for instance \citep{lecun2012efficient}), we focus the following discussion on losses and on the optimization tricks necessary to attain convergence. We also review the contribution of pre-training on some other dataset and data augmentation techniques which bring about an excellent initialization point and good generalizations respectively.

\subsubsection{Losses}
\label{subsec:losses}
Multi-variate cross entropy loss, or log loss, is generally used throughout the literature to classify images or regions in the context of detectors. However, detecting objects in large images comes with its own set of specific challenges: regress bounding boxes to get precise localization, which is a hard problem that is not present at all in classification and an imbalance between target object regions and background regions.

A binary cross entropy loss is formulated as shown in Eq.~\ref{eq:cross_entropy}. It is used for learning the combined objectness. All instances, $y$, are marked as positive labels with a value one. This equation constraints the network to output the predicted confidence score, $p$, to be $1$ if it thinks there is an object and $0$ otherwise.

\begin{equation}
\label{eq:cross_entropy}
CE(p,y)=
\begin{cases}
-log(p) \qquad \quad \text{if } y=1 \\
-log(1-p) \quad \text{otherwise} \\
\end{cases}
\end{equation}

A multi-variate version of the log loss is used for classification (Eq. \ref{eq:multivariate_cross_entropy}). $p_{o,c}$ predicts the probability of observation $o$ being class $c$ where $c \in \{1,.., C\}$. $y_{o,c}$ is $1$ if observation $o$ belongs to class $c$ and $0$ otherwise. $c=0$ is accounted for the special case of background class.
\begin{equation}
\label{eq:multivariate_cross_entropy}
MCE(p,y)= -\sum_{c=0}^{C}y_{o,c}\text{log}(p_{o,c})
\end{equation}

Fast-RCNN \citep{girshick2015fast} used a multitask loss (Eq. \ref{eq:frcnn_mutitask}) which is the de-facto equation used for classifying as well as regressing. The losses are summed over all the regions proposals or default reference boxes, $i$. The ground-truth label, $p_i^*$, is $1$ if the proposal box is positive, otherwise $0$. Regularization is learned only for positive proposal boxes.
\begin{equation}
\label{eq:frcnn_mutitask}
\begin{aligned}[b]
L(\{p_i\},\{t_i\}) = \frac{1}{N_{cls}}\sum_{i}L_{cls}(p_i, p^*_i) +\\
\lambda\frac{1}{N_{reg}}\sum_{i}p^*_iL_{reg}(t_i, t^*_i) 
\end{aligned}
\end{equation}
where $t_i$ is a vector representing the 4 coordinates of the predicted bounding box and similarly $t_i^*$ represents the 4 coordinates of the ground truth. Eq. \ref{eq:frcnn_bbox} presents the equation for exact parameterized coordinates. $\{x_a,y_a,w_a,h_a\}$ are the center x and y coordinates, width and height of the default anchor box respectively. Similarly $\{x_a^*,y_a^*,w_a^*,h_a^*\}$ are ground truths and $\{x,y,w,h\}$ are the coordinates to be predicted. The two terms are normalized by mini-batch size, $N_{cls}$, and number of proposals/default reference boxes, $N_{reg}$, and weighted by a balancing parameter $\lambda$.
\begin{equation}
\label{eq:frcnn_bbox}
\begin{aligned}
t_x = \frac{x-x_a}{w_a}, \quad t_y = \frac{y-y_a}{h_a} \\
t_w = log\frac{w}{w_a}, \quad t_h = log\frac{h}{h_a} \\
t^*_x = \frac{x^*-x_a}{w_a}, \quad t^*_y = \frac{y^*-y_a}{h_a} \\
t^*_w = log\frac{w^*}{w_a}, \quad t^*_h = log\frac{h^*}{h_a} 
\end{aligned}
\end{equation}
$L_{reg}$ is a smooth $L_1$ loss defined by Eq. \ref{eq:Lreg}. In its place some papers also use $L_2$ losses.
\begin{equation}
\label{eq:Lreg}
\begin{aligned}
l_{reg}(t, t^*)=
\begin{cases}
0.5(t-t^*)^2 \quad \quad \text{if } |t-t^*| < 1 \\
|t-t^*| - 0.5 \qquad \text{otherwise}
\end{cases}
\end{aligned}
\end{equation}

\paragraph{Losses for regressing bounding boxes:} Since accurate localization is a major issue, papers have suggested a more sophisticated localization loss. \citep{gidaris2016LocNetImprovingLocalization} came up with a binary logistic type regression loss. After dividing the image patch into $M$ columns and $M$ rows, they computed the probability of each row and column being inside or outside the predicted observation box (in-out loss) (Eq. \ref{eq:locnet_in_out}).

\begin{equation}
\label{eq:locnet_in_out}
\begin{aligned}
&L_{in-out} = \sum_{a \in \{x,y\}} \sum_{i=1}^{M}T_{a,i}\text{log}(p_{a,i}) +\\ &(1-T_{a,i})\text{log}(1-p_{a,i}) \\
&\forall i \in \{1,...,M\}, \quad T_{x,i} = 
\begin{cases}
1, \qquad \text{if } B_l^{gt} \leq i \leq B_r^{gt}\\
0, \quad \quad \text{otherwise} \\
\end{cases} \\
&\forall i \in \{1,...,M\}, \quad T_{y,i} = 
\begin{cases}
1, \qquad \text{if } B_t^{gt} \leq i \leq B_b^{gt}\\
0, \quad \quad \text{otherwise} \\
\end{cases}
\end{aligned}
\end{equation}
where $\{B_l^{gt}, B_r^{gt}, B_t^{gt}, B_b^{gt}\}$ are the left, right, top and bottom edges of the bounding box respectively. $T_x$ and $T_y$ are the binary positive or negative values for rows and columns respectively. $p$ is the probability associated with it respectively. 

In addition, they also compute the confidence for each column and row being the exact boundary of the predicted observation or not (Eq. \ref{eq:locnet_border}).
\begin{equation}
\label{eq:locnet_border}
\begin{aligned}
L_{border} &= \sum_{s \in \{l,r,t,b\}} \sum_{i=1}^{M}\lambda^{+}  T_{s,i}\text{log}(p_{s,i}) + &\\ & \lambda^{-}(1-T_{s,i})\text{log}(1-p_{s,i}) \\
&\forall i \in \{1,...,M\}, \quad T_{s,i} = 
\begin{cases}
1, \qquad \text{if } i = B_s^{gt} \\
0, \quad \quad \text{otherwise} \\
\end{cases}&
\end{aligned}
\end{equation}
where $\quad \lambda^{-} = 0.5 \frac{M}{M-1}, \quad \lambda^{+} = (M-1)\lambda^{-}$. The notations can be inferred from Eq. \ref{eq:locnet_in_out}. In the second paper \citep{gidaris2016AttendRefineRepeat}, related to the same topic, applied the regression losses iteratively at the region proposal stage in a class agnostic manner. They used final convolutional features and predictions from last iteration to further refine the proposals.

It was also found out to be beneficial to optimize the loss directly over Intersection over Union (IoU) which is the standard practice to evaluate a bounding box or segmentation algorithm. \citeauthor{yu2016UnitBoxAdvancedObject} \cite{yu2016UnitBoxAdvancedObject} presented Eq. \ref{eq:unit_box} for regression loss.
\begin{equation}
\label{eq:unit_box}
\begin{aligned}
L_{unit-box} = -\text{ln}(IoU(gt,pred)) \\
\end{aligned}
\end{equation}
The terms are self-explanatory. \citeauthor{jiang2018acquisition} \cite{jiang2018acquisition} also learned to predict IoU between predicted box and ground truth. They made a case to use localization confidence instead of classification confidence to suppress boxes at NMS stage. It gave higher recall on MS COCO dataset. This loss is however very unstable and has a number of regions where the IoU has zero-gradient and thus it is undefined. \citeauthor{tychsen2018fitness} \citep{tychsen2018fitness} adapt this loss to make it more stable by adding hard bounds, which prevent the function from diverging. They also factorize the score function by adding a fitness term representing the IoU of the box \wrt the ground truth.
    

\paragraph{Losses for class imbalance:} 
Since in recent detectors there are a lot of anchors which most of the time cover background, there is a class imbalance between positive and negative anchors. An alternative is Online Hard Example Mining (OHEM). \citeauthor{shrivastava2016training} \cite{shrivastava2016training} performed to select only worst performing examples (so-called hard examples) for calculating gradients. Even if by fixing the ratio between positive and negative instances, generally 1:3, one can partly solve this imbalance. \citeauthor{lin2017focal} \cite{lin2017focal} proposed a tweak to the cross entropy loss, called focal loss, which took into account all the anchors but penalized easy examples less and hard examples more. Focal loss (Eq. \ref{eq:focal_loss}) was found to increase the performance by 3.2 mAP points on MS COCO, in comparison to OHEM on a ResNet-50-FPN backbone and 600 pixel image scale.
\begin{equation}
\label{eq:focal_loss}
FL(p,y)=
\begin{cases}
-\alpha_t(1-p)^{\gamma}log(p) \quad \text{if } y=1 \\
-\alpha_{t }p^{\gamma}log(1-p) \quad \quad \text{otherwise} \\
\end{cases}
\end{equation}
One can also adopt simpler strategies like rebalancing the cross-entropy by putting more weights on the minority class \citep{ogier2017icip}.

\paragraph{Supplementary losses:} 
In addition to classification and regression losses, some papers also optimized extra losses in parallel. 
\citeauthor{dai2016instance} \cite{dai2016instance} proposed a three-stage cascade for differentiating instances, estimating masks and categorizing objects. Because of this they achieved competitive performance on object detection task too. They further experimented with a five-stage cascade also. UberNet \citep{kokkinos2016UberNetTrainingUniversal} trained on as many as six other tasks in parallel with object detection. \citeauthor{he2017mask} \cite{he2017mask} have shown that using an additional segmentation loss by adding an extra branch to the Faster R-CNN detection sub-network can also improve detection performance.
\citet{DBLP:conf/cvpr/LiQDJW17} introduced position-sensitive inside/outside score maps to train for detection and segmentation simultaneously. \citeauthor{wang2018RepulsionLossDetecting} \cite{wang2018RepulsionLossDetecting} proposed an additional repulsion loss between predicted bounding boxes in order to have one final prediction per ground truth. Generally, it can be observed, \emph{instance segmentation} in particular, aids the object detection task.

\subsubsection{Hyper-Parameters}
The detection problem is a highly non-convex problem in hundreds of thousands of dimensions. Even for classification where the structure is much simpler, no general strategy has emerged yet on how to use mini-batch gradient descent correctly. Different popular versions of mini-batch Stochastic Gradient Descent(SGD) \citep{rumelhart1985learning} have been proposed based on a combination of momentum, to accelerate convergence, and using the history of the past gradients, to dampen the oscillations when reaching a minimum: AdaDelta \citep{zeiler2012adadelta}, RMSProp \citep{tieleman2012lecture} and the unavoidable ADAM \citep{kingma2014adam,reddi2018convergence} are only the most well-known. However, in object detection literature authors, use either plain SGD or ADAM, without putting too much thought into it. The most important hyper-parameters remain the learning rate and the batch size.

\paragraph{Learning rate:} 
There is no concrete way to decide the learning rate policy over the period of the training. It depends on a myriad of factors like optimizer, number of training examples, model, batch size, \etc We cannot quantify the effect of each factor; Therefore, the current way to determine the policy is by hit-and-trial. What works for one setting may or may not work for other settings. If the policy is incorrect then the model might fail to converge at all. Nevertheless, some papers have studied it and have established general guidelines that have been found to work better than others. A large learning rate might never converge while a small learning rate gives sub-optimal results. Since, in the initial stage of training the change in weights is dramatic, \citeauthor{goyal2017accurate} \cite{goyal2017accurate} have proposed a \textit{Linear Gradual Warmup} strategy in which learning rate is increased every iteration during this period. Then starting from a point (for $e.g.$ $10^{-3}$) the policy was to decrease learning rate over many epochs. \citeauthor{krizhevsky2014one} \cite{krizhevsky2014one} and \citeauthor{goyal2017accurate} \cite{goyal2017accurate} also used a \textit{Linear Scaling Rule} which linearly scaled the learning rate according to the mini-batch size.

\paragraph{Batch size:}
The object detection literature doesn't generally focus on the effects of using a bigger or smaller batch size during training. Training modern detectors requires working on full images and therefore on large tensors which can be troublesome to store on the GPU RAM. It has forced the community to use small batches, of 1 to 16 images, for training (16 in RetinaNet \citep{lin2017focal} and Mask R-CNN \citep{he2017mask} with the latest GPUs). 

One obvious advantage of increasing the batch size is that it reduces the training time but since the memory constraint restricts the number of images, more GPUs have to be employed. However, using extra large batches have been shown to potentially lead to big improvements in performances or speed. For instance, batch normalization \citep{Ioffe_2015_ICML} needs many images to provide meaningful statistics. Originally batch size effects were studied by \citep{goyal2017accurate} on ImageNet dataset. They were able to show that by increasing the batch size from 256 to 8192, train time can be reduced from 29 hours to just 1 hour while maintaining the same accuracy. Further, \citeauthor{you2017ImageNetTrainingMinutes} \cite{you2017ImageNetTrainingMinutes} and \citeauthor{akiba2017ExtremelyLargeMinibatch} \cite{akiba2017ExtremelyLargeMinibatch} brought down the training time to below 15 minutes by increasing the batch size to 32k.

Very recently, MegDet \citep{peng2017megdet}, inspired from \citep{goyal2017accurate}, have shown that by averaging gradients on many GPUs to get an equivalent batch size of 256 and adjusting the learning rates could lead to some performance gains. It is hard to say now which strategy will eventually win in the long term but they have shown that it is worth exploring.

\subsubsection{Pre-Training}
Transfer learning was first shown to be useful in a supervised learning approach by \citeauthor{girshick2014rich} \cite{girshick2014rich}. The idea is to fine-tune from a model already trained on a dataset that is similar to the target dataset. This is usually a better starting point for training instead of randomly initializing weights. For $e.g.$ model pre-trained on ImageNet being used for training on MS COCO. And since, COCO dataset's classes is a superset of PASCAL VOC's classes most of the state-of-the-art approaches pre-train on COCO before training it on PASCAL VOC. If the dataset at hand is completely unrelated to dataset used for pre-training, it might not be useful. For \eg model pre-trained on ImageNet being used for detecting cars in aerial images. 

\citeauthor{singh2018analysis} \cite{singh2018analysis} made a compelling case for the minimum difference in scales of object instances between classification dataset used for pre-training and detection dataset to minimize domain shift while fine-tuning. They asked should we pre-train CNNs on low resolution classification dataset or restrict the scale of object instances in a range by training on an image pyramid? By minimizing scale variance they were able to get better results than the methods that employed scale invariant detector. The problem with the second approach is some instances are so small that in order to bring them in the scale range, the images have to be upscaled so much that they might not fit in the memory or they will not be used for training at all. Using a pyramid of images and using each for inference is also slower than methods that use input image exactly once.

Section~\ref{subsec:domain_adaptation} covers pre-training and other aspects of it like fine-tuning and beyond in great detail. There are only, to the best of our knowledge, two articles that tried to match the performances of ImageNet pre-training by training detectors from scratch. The first one being \citep{shen2017dsod} that used deep supervision (dense access to gradients for all layers) and very recently \citep{shen2017LearningObjectDetectors} that adaptively recalibrated supervision intensities based on input object sizes. 

\subsubsection{Data Augmentation}

\begin{figure}
\centering
    \includegraphics[]{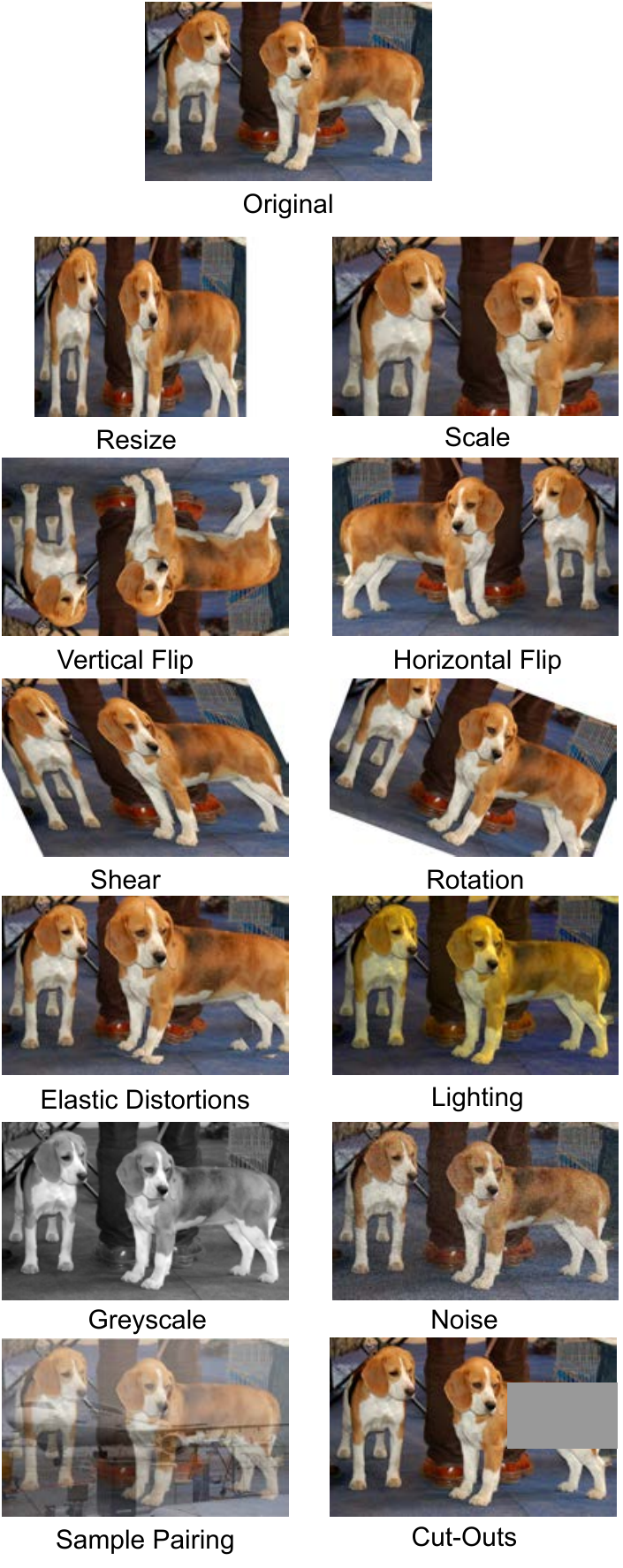}
    \caption{Different kinds of data augmentation techniques used to improve the generalization of the network. The modification done for each image is mentioned below the figure. Best viewed in color. 
    \label{fig:augmentation}}
\end{figure}

The aim of augmenting the train set images is to create diversity, avoid overfitting, increase amount of data, improve generalizability and overcome different kinds of variances. This can easily be achieved without any extra annotations efforts by manually designing many augmentation strategies. The general practices that are followed include and are not limited to scale, resize, translation, rotation, vertical and horizontal flipping, elastic distortions, random cropping and contrast, color, hue, brightness, saturation and sharpness adjustments \etc The two recent and promising but not widely adapted techniques are Cutout \citep{devries2017improved} and Sample Pairing \citep{inoue2018data}. \citeauthor{taylor2017ImprovingDeepLearning} \cite{taylor2017ImprovingDeepLearning} benchmarked various popular data augmentation schemes to know the ones that are most appropriate, and found out that {\em cropping} was the most influential in their case. 

Although there are many techniques available and each one of them is easy to implement, it is difficult to know in advance, without expert knowledge, which techniques assist the performance for a target dataset. For example, vertical flipping in case of traffic signs dataset is not helpful because one is not likely to encounter inverted signs in the test set. It is not trivial to select the approaches for each target dataset and test all of them before deploying a model. Therefore, \citeauthor{cubuk2018autoaugment} \cite{cubuk2018autoaugment} proposed a search algorithm based on reinforcement learning to find the best augmentation policy. Their approach tried to find the best suitable augmentation operations along with their magnitude and probability of happening. Smart Augmentation \citep{lemley2017smart} worked by creating a network that learned how to automatically generate augmented data during the training process of a target network in a way that reduced the loss. \citeauthor{tran2017bayesian} \cite{tran2017bayesian} proposed a Bayesian approach, where new annotated training points are treated as missing variables and generated based on the distribution learned from the training set. \citeauthor{devries2017dataset} \cite{devries2017dataset} applied simple transformations such as adding noise, interpolating, or extrapolating between data points. They performed the transformation, not in input space, but in a learned feature space. All the above approaches are implemented in the domain of classification only but they might be beneficial for the detection task as well and it would be interesting to test them.

Generative adversarial networks (GANs) have also been used to generate the augmented data directly for classification without searching for the best policies explicitly \citep{perez2017effectiveness, mun2017generative, antoniou2018data,sixt2018rendergan}. \citeauthor{ratner2017learning} \cite{ratner2017learning} used GANs to describe data augmentation strategies. GAN approaches may not be as effective for detection scenarios yet because generating an image with many object instances placed in a relevant background is much more difficult than generating an image with just one dominant object. This is also an interesting problem which might be addressed in the near future and is explored in Section~\ref{sec:gan}. 

\begin{figure*}[htb]
\centering
    \includegraphics[width=\textwidth]{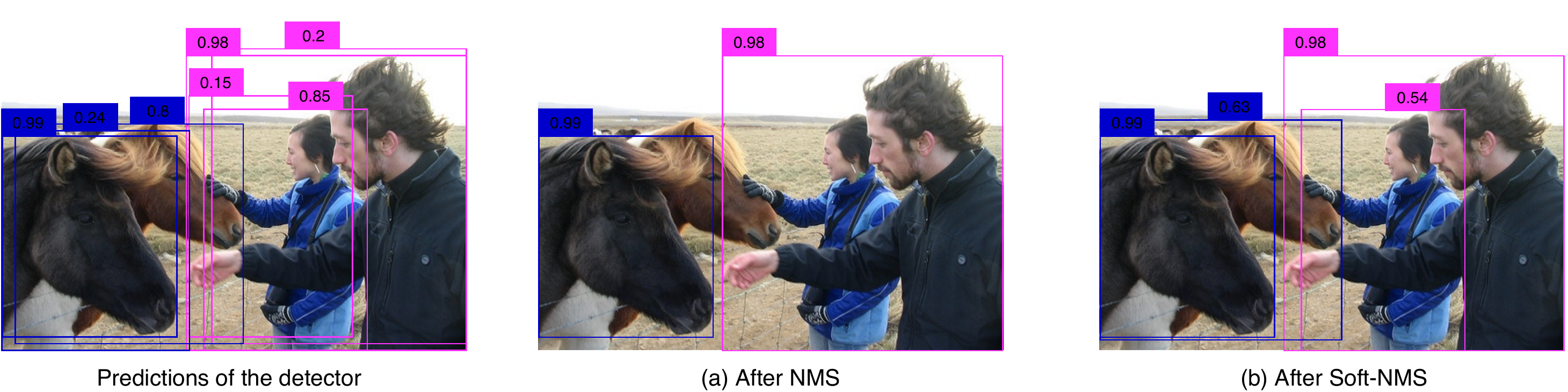}
    \caption{An illustration of the inference stage. In this example, bounding boxes around horses (blue) and persons (pink) are obtained from the detector (along with the confidence scores mentioned on top of each box). (a) NMS chooses the most confident box and suppresses all other boxes having an IoU greater than a threshold. Note, it sometimes leads to suppression of boxes around other occluded objects. (b) Soft-NMS deals with this situation by reducing the confidence scores of boxes instead of completely suppressing them. 
    \label{fig:inference}}
\end{figure*}
\subsection{Inference}
\label{sec:inference}
The behavior of the modern detectors is to pick up pixels of target objects, propose as many windows as possible surrounding those pixels and estimate confidence scores for each of the window. It does not aim to suggest one box exactly per object. Since all the reference boxes act independently during test time and similar input pixels are picked up by many neighboring anchors, each positive prediction in the prediction set highly overlaps with other boxes. If the best ones out of these are not selected, it will lead to many double detections and thus false positives. The ideal result would be to predict exactly one prediction box per ground-truth object that has high overlap with it. To reach near this ideal state, some sort of post-processing needs to be done.

Greedy Non-maximum suppression (NMS) \citep{dalal2005histograms} is the most frequent technique used for inference modules to suppress double detections through hard thresholding. In this approach, the prediction box with the highest confidence was chosen and all the boxes having an Intersection over Union (IoU) higher than a threshold, $N_t$, were suppressed or rescored to zero. This step was made iteratively till all the boxes were covered. Because of its nature there is no single threshold that works best for all datasets. Datasets with just one object per image will trivially apply NMS by choosing only the highest-ranking box. Generally, datasets with sparse and fewer number of objects per image (2 to 3 objects) require a lower threshold. While datasets with cramped and higher numbers of objects per image (7 and above) give better results with a higher threshold. The problems that arose with this naive and hand-crafted approach was that it may completely suppress nearby or occluded true positive detections, choose top scoring box which might not be the best localized one and its inability to suppress false positives with insufficient overlap.

To improve upon it, many approaches have been proposed but most of them work for a very special case such as pedestrians in highly occluded scenarios. We discuss the various directions they take and the approaches that work better than the greedy NMS in the general scenario. Most of the following discussion is based on \citep{hosang2017learning} and \citep{bodla2017soft}, who, in their papers, provided us with an in-depth view of the alternate strategies being used. 

Many clustering approaches for predicted boxes have been proposed. A few of them are mean shift clustering \citep{dalal2005histograms,wojek2008sliding}, agglomerative clustering \citep{bourdev2010detecting}, affinity propagation clustering \citep{mrowca2015spatial}, heuristic variants \citep{sermanet2013overfeat}, \etc. \citeauthor{rothe2014non} \cite{rothe2014non} presented a learning based method which "passes messages between windows" or clustered the final detections to finally select exemplars from each cluster. \citeauthor{mrowca2015spatial} \cite{mrowca2015spatial} deployed a multi-class version of this paper. Clustering formulations with globally optimal solutions have been proposed in \citep{tang2015subgraph}. All of them worked for special cases but are less consistent than Greedy NMS, generally.

Some papers learn NMS in a convolutional network. \citeauthor{henderson2016end} \cite{henderson2016end} and \citeauthor{wan2015EndtoendIntegrationConvolutional} \cite{wan2015EndtoendIntegrationConvolutional} tried to incorporate NMS procedure at training time. \citeauthor{stewart2016end} \cite{stewart2016end} generated a sparse set of detections by training an LSTM. \citeauthor{hosang2016convnet} \cite{hosang2016convnet} trained the network to find different optimal cutoff thresholds ($N_t$) locally. \citeauthor{hosang2017learning} \cite{hosang2017learning} took one step further and got rid of the NMS step completely by taking into account double detections in the loss and jointly processed neighboring detections. The former inclined the network to predict one detection per object and the latter provided with the information if an object was detected multiple times. Their approach worked better than greedy NMS and they obtained a performance gain of 0.8 mAP on COCO dataset.

Most recently, greedy NMS was improved upon by \citeauthor{bodla2017soft} \cite{bodla2017soft}. Instead of setting the score of neighboring detections as zero they decreased the detection confidence as an increasing function of overlap. It improved the performance by 1.1 mAP for COCO dataset. There was no extra training required and since it is hand-crafted it can be easily integrated in object detection pipeline. It is used in current top entries for MS COCO object detection challenge.

\citeauthor{jiang2018acquisition} \cite{jiang2018acquisition} performed NMS based on separately predicted localization confidence instead of usually accepted classification confidence. Other papers rescored detections locally \citep{chen2013detection,tu2010auto} or globally \citep{vezhnevets2015object}. Some others detected objects in pairs in order to handle occlusions \citep{ouyang2013single,sadeghi2011recognition,tang2014detection}. \citeauthor{rodriguez2011density} \cite{rodriguez2011density} made use of the crowd density. Quadratic unconstrained binary optimization (QUBO) \citep{rujikietgumjorn2013optimized} used detection scores as a unary potential and overlap between detections as a pairwise potential to obtain the optimal subset of detection boxes. \citeauthor{niepert2016learning} \cite{niepert2016learning} saw overlapping windows as edges in a graph.

As a bonus, in the end, we also throw some light on the inference "tricks" that are generally known to the experts participating in the competitions. The tricks that are used to further improve the evaluation metrics are: Doing multi-scale inference on an image pyramid (see Section~\ref{subsec:scalevariance} for training); Doing inference on the original image and on its horizontal flip (or on different rotated versions of the image if the application domain does not have a fixed direction) and aggregating results with NMS; Doing bounding box voting as in \citep{gidaris2015object} using the score of each box as its weight; Using heavy backbones, as observed in the backbone section; Finally, averaging the predictions of different models in ensembles. For the last trick often it is better to not necessarily use the top-N best performing models but to prefer instead  uncorrelated models so that they can correct each other's weaknesses. Ensembles of models are outperforming single models by often a large margin and one can average as many as a dozen models to outrank its competitors. Furthermore, with DCNNs generally one does not need to put too much thought on normalizing the models as each one gives bounded probabilities (because of the softmax operator in the last layer). 

\begin{figure*}
\centering
    \includegraphics[width=\textwidth]{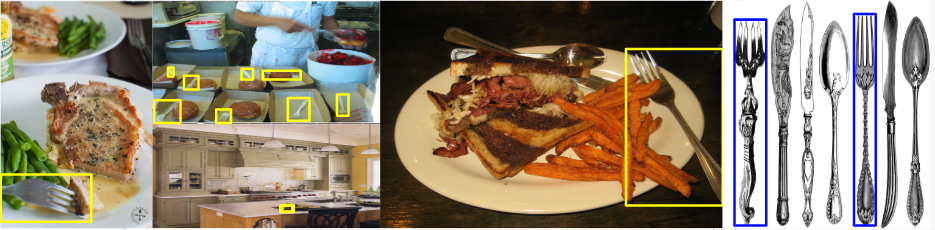}
    \caption{An illustration of challenges in object detection. To detect all instances of the class "fork" (yellow bounding boxes) from the COCO dataset~\citep{lin2014microsoft}, a detector should be able to handle \textbf{small objects} (lower middle picture) as well as big objects (third column photograph). It needs to be \textbf{scale invariant} as well as \textbf{being rotation invariant} (all forks have different orientation in the pictures). It should also manage \textbf{occlusions} as in the left-hand side photograph. After being trained on the pictures in the first three columns, detection algorithms are expected to generalize to the "cartoon" image on the right (\textbf{domain adaptation}).
    \label{fig:coco_challenge}}
\end{figure*}

\subsection{Concluding Remarks}
This concludes a general overview of the landscape of the mainstream object detection halfway through 2018. Although the methods presented are all different, it has been shown that in fact most papers have converged towards the same crucial design choices. All pipelines are now fully convolutional, which brings structure (regularization), simplicity, speed and elegance to the detectors. The anchors mechanism of \citeauthor{ren2015faster} \cite{ren2015faster} has now also been widely adopted and has not really been challenged yet, although iteratively regressing a set of seed boxes show some promise \citep{najibi2016GCNNIterativeGrid,gidaris2016AttendRefineRepeat}. The need to use multi-scale information from different layers of the CNN is now apparent \citep{kong2016HyperNetAccurateRegion,lin2017feature,lin2017focal}.
The RoI-Pooling module and its cousins can also be cited as one of the main architectural advances of recent years but might not ultimately be used by future works. 

With that said, most of the research being done now in the mainstream object recognition consists of inventing new ways of passing the information through the different layers or coming up with different kinds of losses or parametrization \citep{yu2016UnitBoxAdvancedObject,gidaris2016LocNetImprovingLocalization}.
There is a small paradox now in the fact that even if man-made features are now absent of most modern detectors, more and more research is being done on how to better hand-craft the CNN architectures and modules.

%% file: article_going_forward.tex

\section{Going Forward in Object Detection}
\label{sec:going_forward}

\begin{figure}
    \centering
    \includegraphics[height=4.5cm, width=7cm]{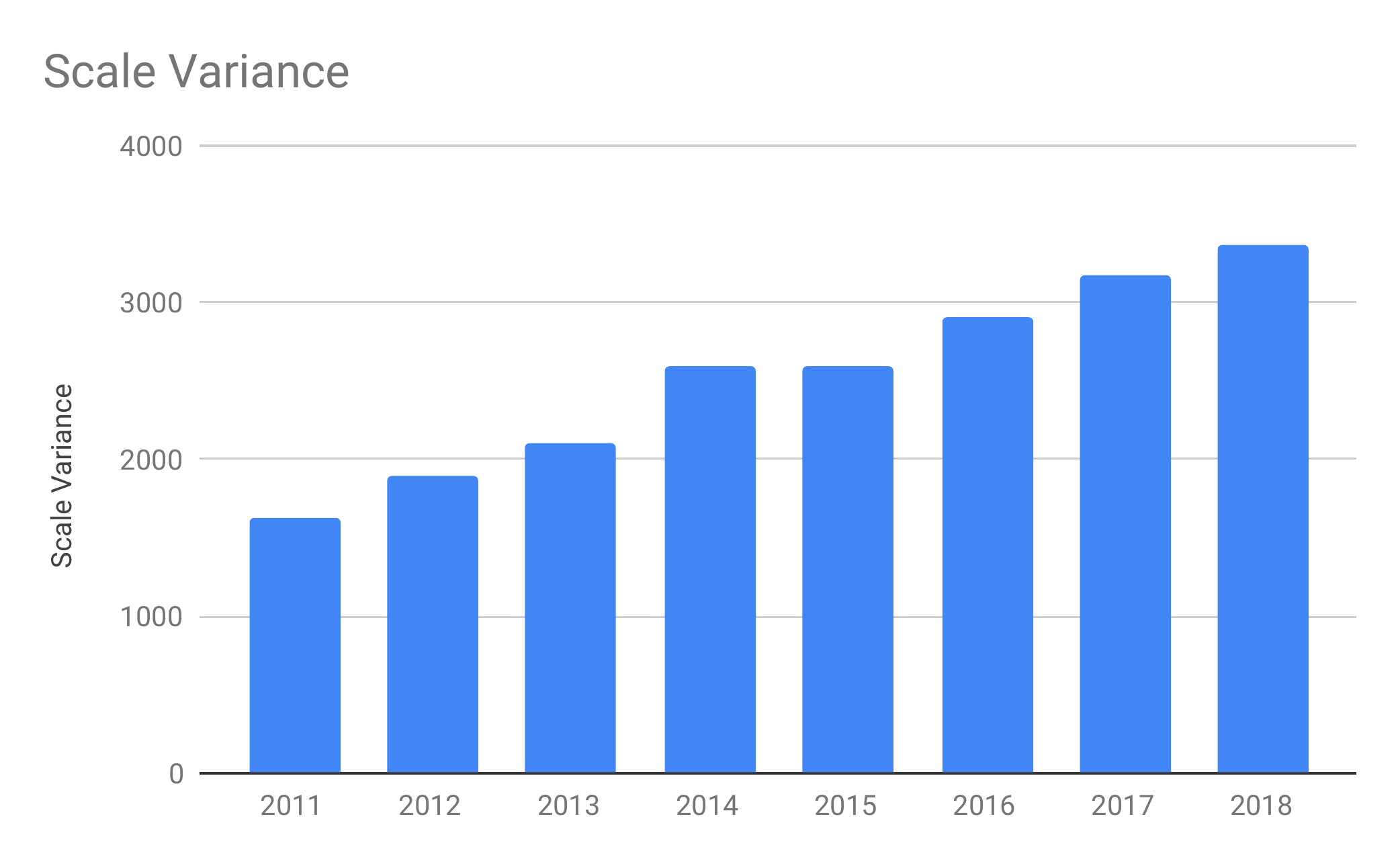}
    \caption{Number of papers published each year for challenge of scale variance. Query used in Google Scholar: ("scale variance" OR "scale invariance" OR "scale invariant") AND "object detection".}
    \label{fig:scale_variance}
\end{figure}

\begin{figure}
    \centering
    \includegraphics[height=4.5cm,width=7cm]{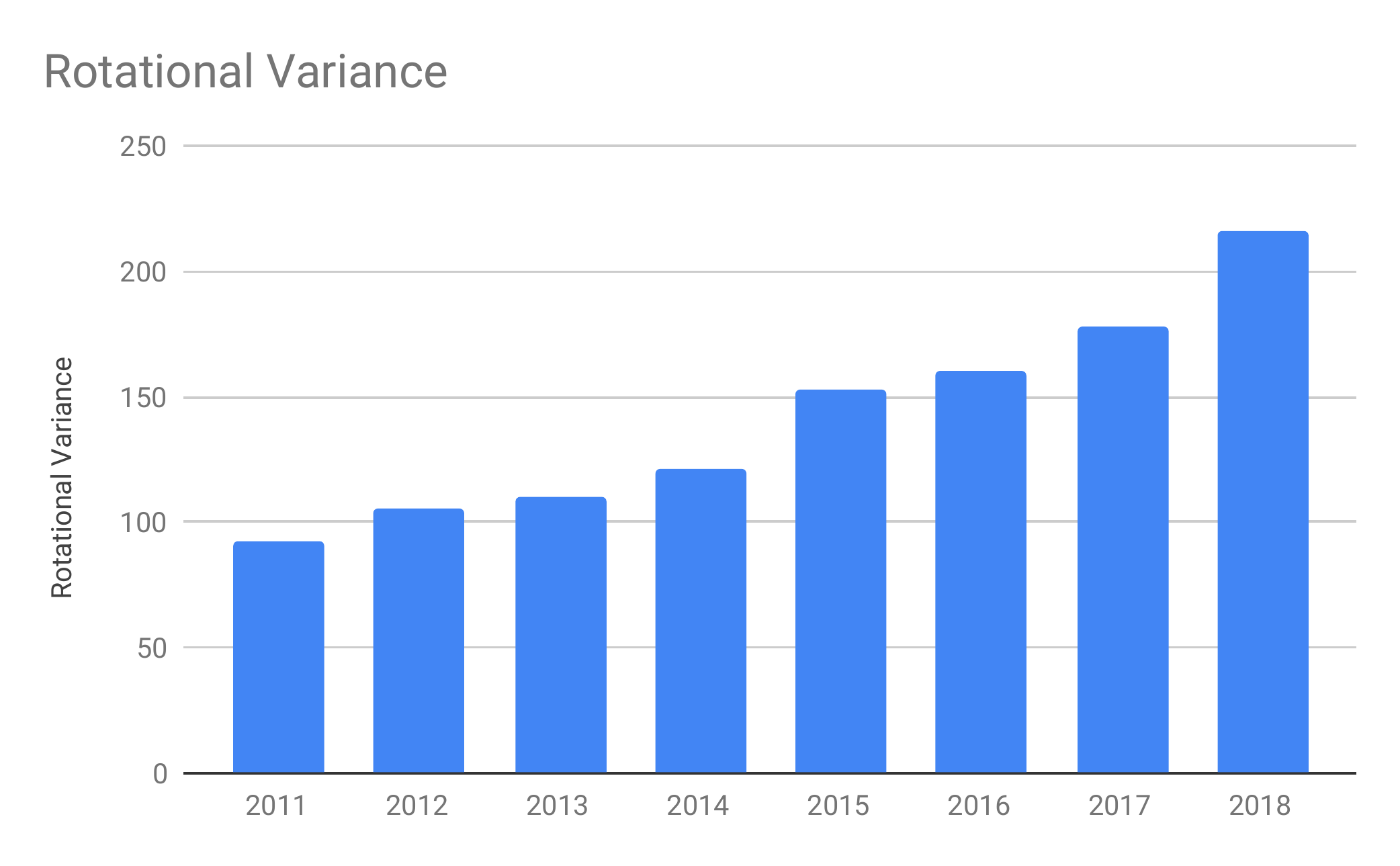}
    \caption{Number of papers published each year for challenge of rotational variance. Query used in Google Scholar: ("rotational variance" OR "rotational invariance" OR "rotational invariant") AND "object detection".}
    \label{fig:rotational_varaiance}
\end{figure}

\begin{figure}
    \centering
    \includegraphics[height=4.5cm,width=7cm]{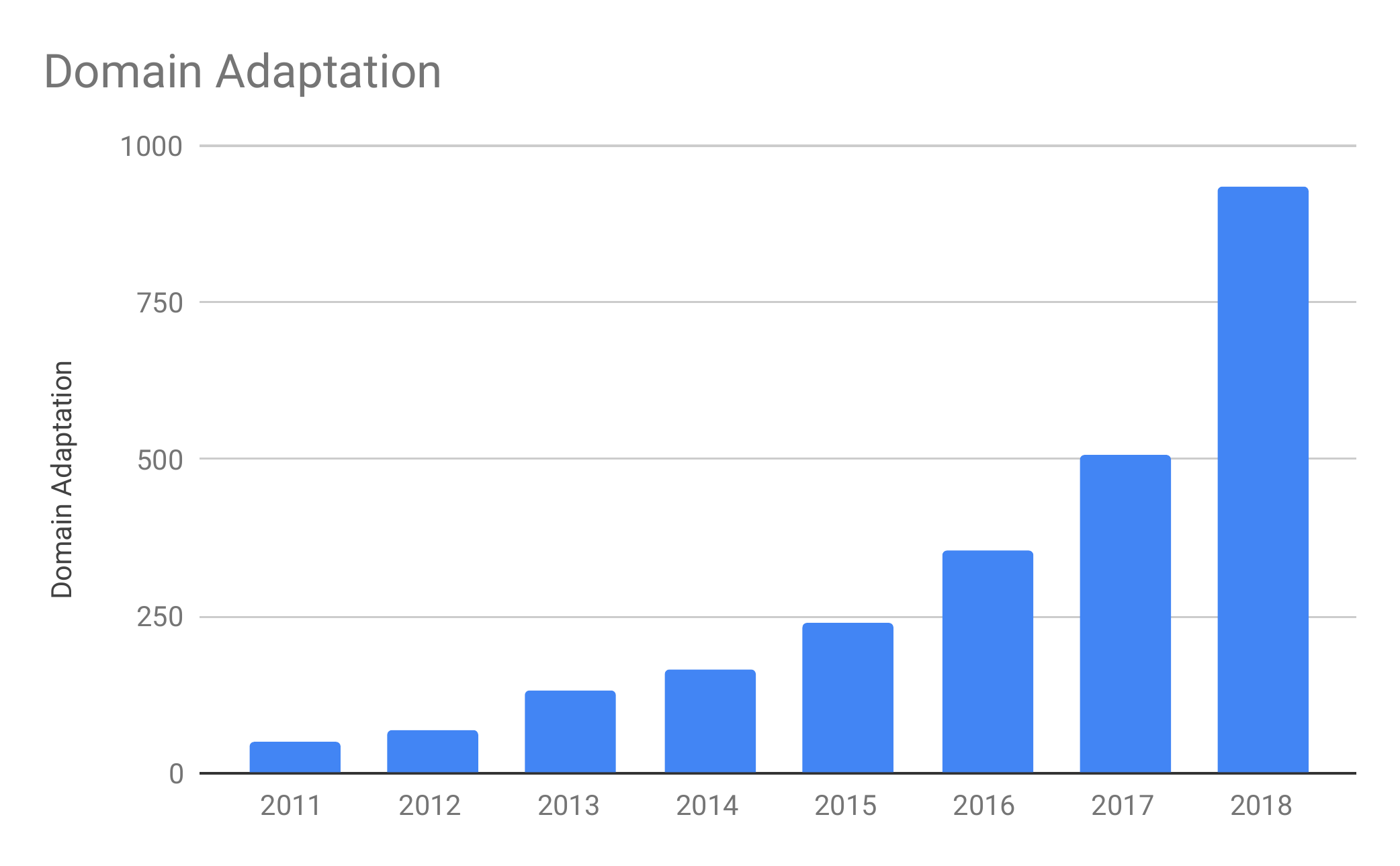}
    \caption{Number of papers published each year for challenge of domain adaptation. Query used in Google Scholar: ("domain adaptation" OR "adapting domains") AND "object detection".}
    \label{fig:domain_adaptation}
\end{figure}

\begin{figure}
    \centering
    \includegraphics[height=4.5cm,width=7cm]{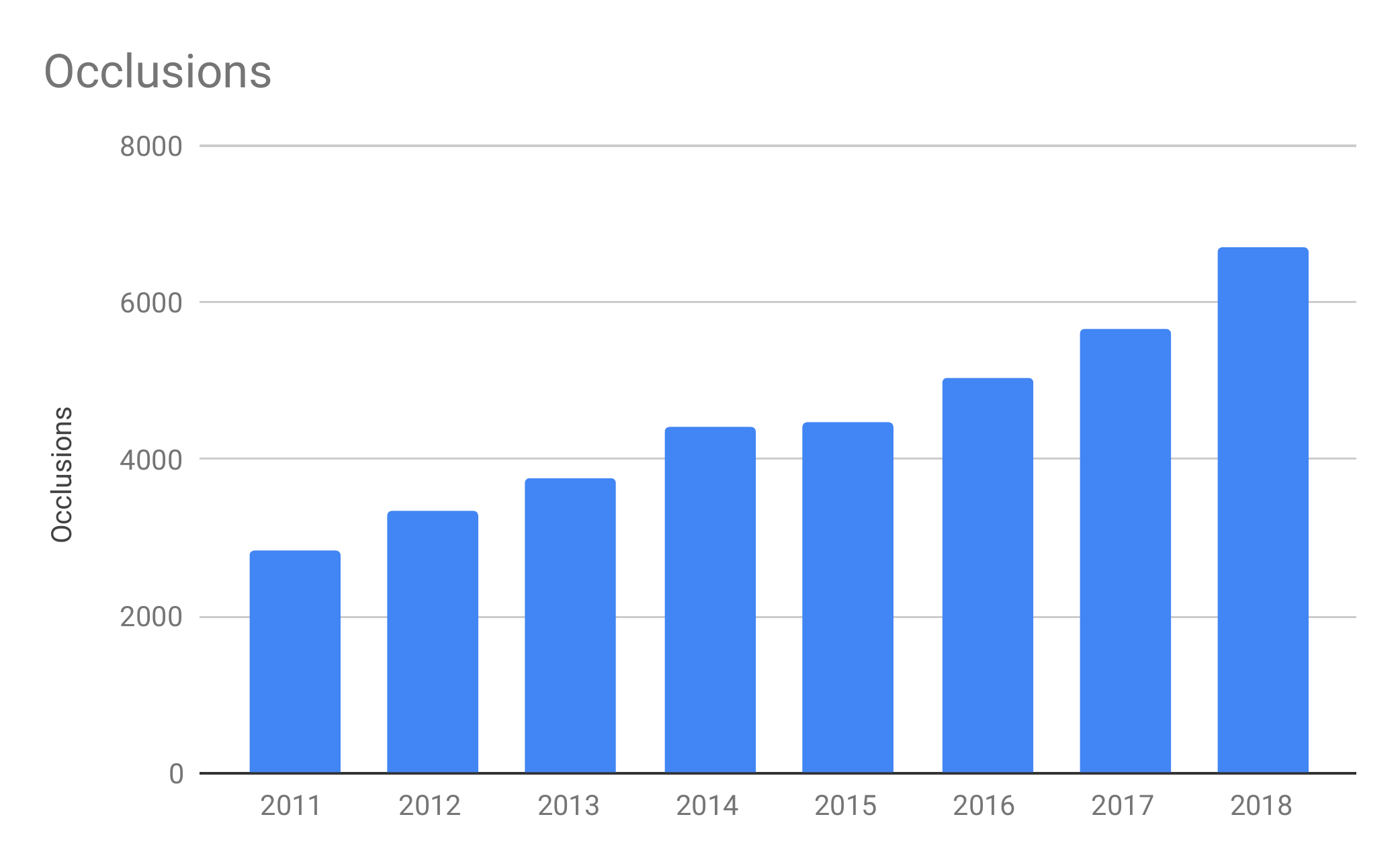}
    \caption{Number of papers published each year for challenge of occlusion. Query used in Google Scholar: (occlusions OR occlusion OR occluded) AND "object detection".}
    \label{fig:occlusions}
\end{figure}

\begin{figure}
    \centering
    \includegraphics[height=4.5cm,width=7cm]{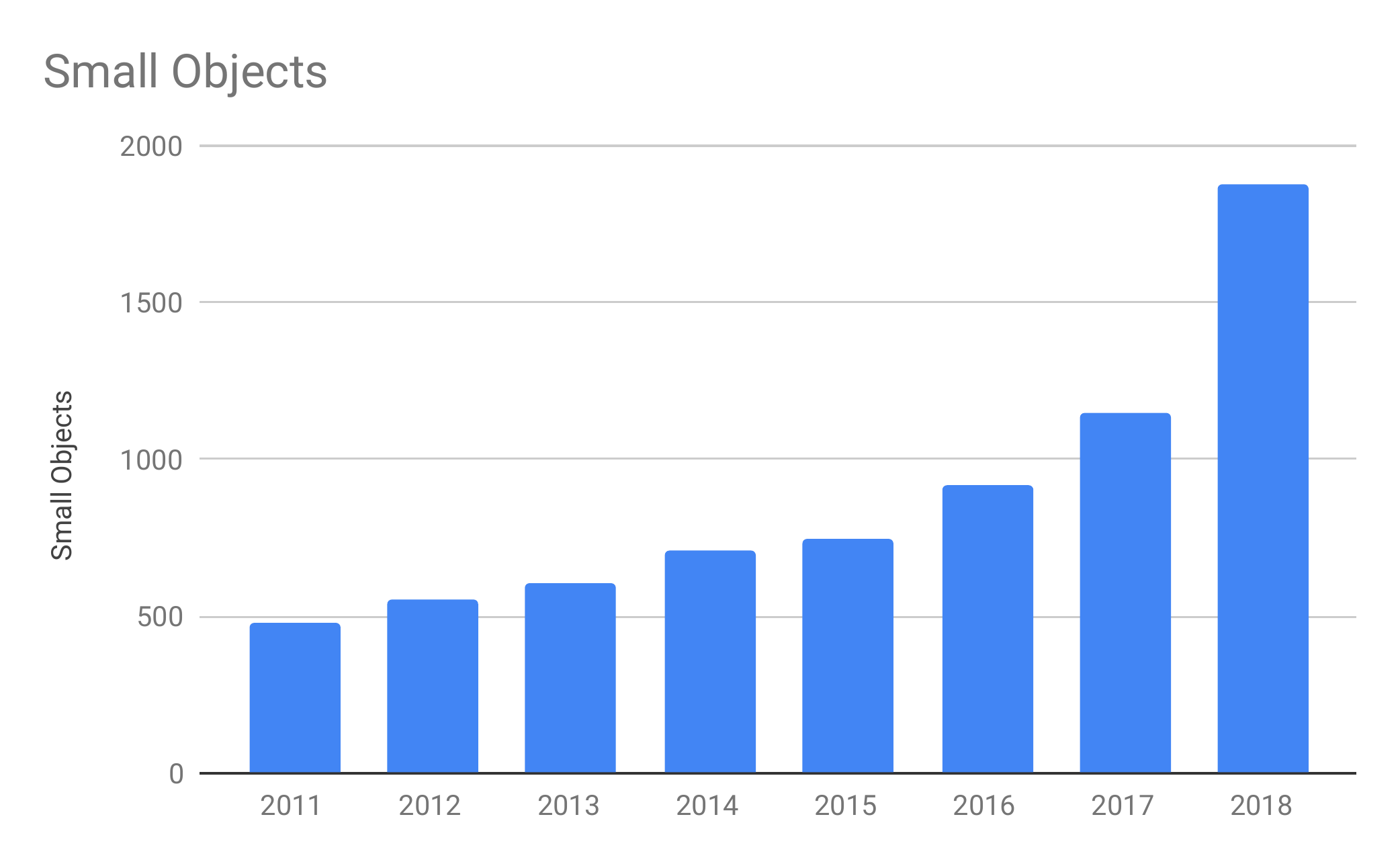}
    \caption{Number of papers published each year for challenge of small objects. Query used in Google Scholar: ("small objects" OR "small object") AND "object detection".}
    \label{fig:small_objects}
\end{figure}

While we demonstrated that object detection has already been turned upside-down by CNN architectures and that nowadays most methods revolve around the same architectural ideas, the field has not yet reached a status quo, far from it. Completely new ideas and paradigms are being developed and explored as we write this survey, shaping the future of object detection. This section lists the major challenges that remain mostly unsolved and the attempts to get around them using such ideas. To have an idea of number of papers being published targeting each challenge, we ran a corresponding query on advanced search of Google Scholar. The exact query is mentioned below each figure respectively (see Figure \ref{fig:scale_variance}, \ref{fig:rotational_varaiance}, \ref{fig:domain_adaptation}, \ref{fig:occlusions} and \ref{fig:small_objects}). We report these numbers from year $2011$ to $2018$. We note that this method doesn't give the exact number of papers targeting each challenge but still gives us a rough idea of the interest of the community in each challenge. We couldn't use this for the localization challenge because almost all object detection papers mention localization even if they are not targeting to solve it.

\subsection{Major Challenges}

There are some walls that the current models cannot overcome without heavy structural changes, we list these challenges in Figure~\ref{fig:coco_challenge}. 

Often, when we hear that object recognition is solved, we argue that the existence of these walls are solid proof that it is not. Although we have advanced the field, we cannot rely indefinitely on the current DCNNs. This section shows how the recent literature addressed these topics.
\subsubsection{Scale Variance}
\label{subsec:scalevariance} 

In the past three years a lot of approaches have been proposed to deal with the challenge of scale variance. On the one hand, object instances in the image may fill only 0.01\% to 0.25\% of the pixels, and, on the other hand, the instance may fill 80\% to 90\% of the whole image. It is tough to make a single feature map predict all the objects, with this huge variance, because of the limited receptive field that it's neurons have. Particularly small objects (discussed in Section~\ref{sec:small_objects}) are difficult to classify and localize. In this section we will discuss three main approaches that are used to tackle the challenge of scale variance.

First, is to make image pyramids \citep{he2015spatial,girshick2015fast,Felzenszwalb2010Object,sermanet2013overfeat}. This helps enlarge small objects and shrink the large objects. Although the variance is reduced to an extent but each image has to be pass forwarded multiple times thus, making it computationally expensive and slower than the approaches discussed in the following discussion. This approach is different from data augmentation techniques \citep{cubuk2018autoaugment} where an image is randomly cropped, zoomed in or out, rotated etc. and used exactly once for inference. \citet{DBLP:journals/pami/RenHGZS17} extracted feature maps from a frozen network at different image scales and merged them using maxout \citep{DBLP:conf/icml/GoodfellowWMCB13}. \citeauthor{singh2018analysis} \cite{singh2018analysis} selectively back-propagated the gradients of object instances if they fall in a predetermined size range. This way, small objects must be scaled up to be considered for training. They named their technique Scale Normalization for Image Pyramids (SNIP). \citeauthor{singh2018sniper} \cite{singh2018sniper} optimized this approach by processing only context regions around ground-truth instances, referred to as chips.

Second, a set of default reference boxes, with varied size and aspect ratios that cover the whole image uniformly, were used. \citeauthor{ren2015faster} \cite{ren2015faster} proposed a set of reference boxes at each sliding window location which are trained to regress and classify. If an anchor box has a significant overlap with the ground truth it is treated as positive otherwise, it is ignored or treated as negative. Due to the huge density of anchors most of them are negative. This leads to an imbalance in the positive and negative examples. To overcome it OHEM \citep{shrivastava2016training} or Focal Loss \citep{lin2017focal} are generally applied at training time. One more downside of anchors is that their design has to be adapted according to the object sizes in the dataset. If large anchors are used with too many small objects then, and vice versa, then they won't be able to train as efficiently. Default reference boxes are an important design feature in double stage \citep{dai2016r} as well as single-stage methods \citep{redmon2016YOLO9000BetterFaster,liu2016ssd}. Most of the top winning entries \citep{he2017mask,lin2017feature,lin2017focal,dai2017deformable} use them in their models. \citeauthor{bodla2017soft} \cite{bodla2017soft} helped by improving the suppression technique of double detections, generated from the dense set of reference boxes, at inference time.

Third, multiple convolutional layers were used for bounding box predictions. Since a single feature map was not enough to predict objects of varied sizes, SSD \citep{liu2016ssd} added more feature maps to the original classification backbones. \citeauthor{cai2016unified} \cite{cai2016unified} proposed regions as well as performed detections on multiple scales in a two-stage detector. \citeauthor{najibi2017SSHSingleStage} \cite{najibi2017SSHSingleStage} used this method to achieve state-of-the-art on a face dataset \citep{yang2016wider} and \citeauthor{li2017scale} \cite{li2017scale} on pedestrian dataset \citep{ess2007depth}. \citeauthor{yang2016exploit} \cite{yang2016exploit} used all the layers to reject easy negatives and then performed scale-dependent pooling on the remaining proposals. Shallower or finer layers are deemed to be better for detecting small objects while top or coarser layers are better at detecting bigger objects. In the original design, all the layers predict the boxes independently and no information from other layers is combined or merged. Many papers, then, tried to fuse different layers \citep{chen2017weaving,lee2017residual} or added additional top-down network \citep{shrivastava2016beyond,woo2017stairnet}. They have already been discussed in Section~\ref{sec:backbone}.

Fourth, Dilated Convolutions (a.k.a. atrous convolutions) \citep{DBLP:journals/corr/YuK15} were deployed to increase the filter's stride. This helped increase the receptive field size and, thus, incorporate larger context without additional computations. Obviously smaller receptive fields are also needed if the objects are small and thus only a clever combination of larger receptive field with atrous convolutions and smaller ones like in ASPP~\citep{DBLP:journals/pami/ChenPKMY18} (Atrous Spatial Pyramid Pooling) can lead to a successful scale invariance in detection.  It has been successfully applied in the context of object detection \citep{dai2016r} and semantic segmentation \citep{DBLP:journals/pami/ChenPKMY18}. \citeauthor{dai2017deformable} \cite{dai2017deformable} presented a generalized version of it by learning the deformation offsets additionally.

\subsubsection{Rotational Variance}
\label{sec:rotation}
In the real world object instances are not necessarily present in an upright manner but can be found at an angle or even inverted. While it is hard to define rotation for flexible objects like a cat, a pose definition would be more appropriate, it is much easier to define it for texts or objects in aerial images which have an expected rigid shape. It is well known that CNNs as they are now do not have the ability to deal with the rotational variance of the data. More often than not, this problem is circumvented by using data augmentation: showing the network slightly rotated versions of each patch. When training on full images with multiple annotations it becomes less practical. Furthermore, like for occlusions, this might work but it is disappointing as one could imagine incorporating rotational invariance into the structure of the network. 

Building rotational invariance can be simply done by using oriented bounding boxes in the region proposal step of modern detectors.
\citeauthor{jiang2017R2CNNRotationalRegion} \cite{jiang2017R2CNNRotationalRegion} used Faster R-CNN features to predict oriented bounding boxes, their straightened versions were then passed on to the classifier. 
More elegantly, few works like \citep{ma2018ArbitraryOrientedSceneText,he2018SingleShotTextSpotter,michalbusta2017DeepTextSpotterEndToEnd} proposed to construct different kinds of RoI-pooling module for oriented bounding boxes.
\citeauthor{ma2018ArbitraryOrientedSceneText} \cite{ma2018ArbitraryOrientedSceneText} transformed the RoI-Pooling layer of Faster R-CNN by rotating the region inside the detector to make it fit the usual horizontal grid, which brought an astonishing increase of performances from the 38.7\% of regular Faster R-CNN to 71.8\% with additional tricks on MSRA. 
Similarly, \citeauthor{he2018SingleShotTextSpotter} \cite{he2018SingleShotTextSpotter} used a rotated version of the recently introduced RoI-Align to pool oriented proposals to get more discriminative features (better aligned with the text direction) that will be used in the text recognition parts.
\citeauthor{michalbusta2017DeepTextSpotterEndToEnd} \cite{michalbusta2017DeepTextSpotterEndToEnd} also used rotated pooling by bilinear interpolation to extract oriented features to recognize text after having rendered YOLO to be able to predict rotated bounding boxes.
\citeauthor{shi2017detecting} \cite{shi2017detecting} detected in the same way, oriented bounding boxes (called segments) with a similar architecture but differ from \citep{ma2018ArbitraryOrientedSceneText,he2018SingleShotTextSpotter,michalbusta2017DeepTextSpotterEndToEnd} because it also learned to merge the oriented segments appropriately, if they cover the same word or sentence, which allowed greater flexibility.

\citeauthor{liu2017DeepMatchingPrior} \cite{liu2017DeepMatchingPrior} needed slightly more complicated anchors: quadrangles anchors, and regressed compact text zones in a single-stage architecture similar to Faster R-CNN's RPN. This system being more flexible than the previous ones, necessitated more parameters. They used Monte-Carlo simulations to compute overlaps between quadrangles. 
\citeauthor{liao2018RotationSensitiveRegressionOriented} \cite{liao2018RotationSensitiveRegressionOriented} directly rotated convolution filters inside the SSD framework, which effectively rendered the network rotation-invariant for a finite set of rotations (which is generalized in the recent \citep{Weiler_2018_CVPR} for segmentation).
However, in the case of text detection even oriented bounding boxes can be insufficient to cover text with a layout with too much curvature and one often sees the same failure cases in different articles (circle-shaped texts for instance).

A different kind of approach for translation invariance was taken by the two following works of \citeauthor{cheng2016RIFDCNNRotationInvariantFisher} \cite{cheng2016RIFDCNNRotationInvariantFisher} and \citeauthor{laptev2016TIPOOLINGTransformationinvariantPooling} \cite{laptev2016TIPOOLINGTransformationinvariantPooling} that made use of metric-learning.
Former proposed an original approach of using metric learning to force features of an image and its rotated versions to be close to each other hence, somehow invariant to rotations. In a somewhat related approach the latter found a canonical pose for different rotated versions of an image and used a differentiable transformation to make every example canonical and to pool the same features.

The difficulty of predicting oriented bounding boxes is alleviated if one resorts to semantic segmentation like in \citep{zhang2016MultiorientedTextDetectionentedTextDetectionentedTextDetection}. They learned to output semantic segmentation then oriented bounding boxes were found based on the output score map. However, it shares the same downsizes as other approaches \citep{ma2018ArbitraryOrientedSceneText,he2018SingleShotTextSpotter,michalbusta2017DeepTextSpotterEndToEnd,liao2018RotationSensitiveRegressionOriented,liu2017DeepMatchingPrior} for text detection because in the end one still has to fit oriented rectangles to evaluate the performances. 

Other applications than text detection also require rotation invariance.
In the domain of aerial imagery, the recently released DOTA \citep{xia2017dota} is one of the first datasets of its kind expecting oriented bounding boxes for predictions. One can anticipate an avalanche of papers trying to use text detection techniques like \citep{tianyutang2017ArbitraryOrientedVehicleDetection}, where the SSD framework is used to regress bounding box angles or the former metric learning technique from \citeauthor{cheng2016RIFDCNNRotationInvariantFisher} \cite{cheng2016RIFDCNNRotationInvariantFisher} and \citeauthor{cheng2016learning} \cite{cheng2016learning}. 
For face detection, paper like \citep{Shi_2018_CVPR} relied on oriented proposals too. The diversity of the methods show that no real standard has emerged yet. Even the most sophisticated detection pipelines are only rotation invariant to a certain extent. 

The detectors presented in this section do not yet have the same popularity as the vertical ones because all the main datasets like COCO do not present rotated images. One could define a rotated-COCO or rotated-VOC to evaluate the benefit these pipelines could bring over their vertical versions but it is obviously difficult and would not be accepted as is by the community without a strong, well-thought-evaluation protocol.

\subsubsection{Domain Adaptation}
\label{subsec:domain_adaptation}
It is often needed to repurpose a detector trained on domain A to function on domain B. In most cases this is because the dataset in domain A has lots of training examples and the categories in it are generic whereas the dataset in domain B has less training examples and objects that are very specific or distinct from A. There are surprisingly very few recent articles that tackled explicit domain adaptation \emph{in the context of object detection} --\citep{tang2012ShiftingWeightsAdapting,sun2014virtual,xu2014domain} did it for HOG based features -- even though the literature for domain adaptation for classification is dense, as shown by the recent survey of \citeauthor{csurka2017DomainAdaptationVisual} \cite{csurka2017DomainAdaptationVisual}. 
For instance when one trains a Faster R-CNN on COCO and want to test it off-the-shelf on the car images of KITTI \citeauthor{geiger2012we} \cite{geiger2012we} ('car' is one of the 80 classes of COCO) one gets only 56.1\% AP \wrt 83.7\% using more similar images because of the differences between the domains (see \citep{Tremblay_2018_CVPR_Workshops}) .

Most works adapt the features learned in another domain (mostly classification) by simply fine-tuning the weights on the task at hand. Since \citep{girshick2014rich}, literally every state-of-the-art detectors are pre-trained on ImageNet or on an even bigger dataset. This is the case even for relatively large object detection datasets like COCO. There is no fundamental reason for it to be a requirement. The objects of the target domains have to be similar and of the same scales as the objects on which the network was pre-trained as pointed out by \citeauthor{singh2018analysis} \cite{singh2018analysis}, that detected small cars in aerial imagery by first pre-training on ImageNet. The seminal work of \citeauthor{hoffman2014lsda} \cite{hoffman2014lsda}, already evoked in the weakly supervised Section~\ref{subsec:weakly_supervised}, showed how to transfer a good classifier trained on large scale image datasets to a good detector trained on few images by fine-tuning the first layers of a convnet trained on classification and adapting the final layer using nearest neighbor classes. \citeauthor{hinterstoisser2017PreTrainedImageFeatures} \cite{hinterstoisser2017PreTrainedImageFeatures} demonstrated another example of transfer learning where they froze the first layers of detectors trained on synthetic data and fine-tuned only the last layers on the target task.

We discuss below all the articles we found that go farther than simple transfer learning for domain adaptation for object detection. \citeauthor{raj2015SubspaceAlignmentBased} \cite{raj2015SubspaceAlignmentBased} aligned features subspace from different domains for each class using Principal Component Analysis (PCA). \citeauthor{chen2018DomainAdaptiveFaster} \cite{chen2018DomainAdaptiveFaster} used H-divergence theory and adversarial training to bridge the distribution mismatches. All the mentioned articles worked on adapting the features. Thanks to GANs some of them are trying to adapt directly to the image \citep{inoue2018cross}, which used CycleGAN from \citep{zhu2017UnpairedImagetoImageTranslation} to convert images directly from one domain to the other. The object detection community needs to evolve if we want to move beyond transfer-learning.

One of the end goals of domain adaptation would be to be able to learn a model on synthetic data, which is available (almost) for free and to have it performing well on real images. \citeauthor{pepik2015holding} \cite{pepik2015holding} was, to the best of our knowledge, the first to point out that, even though CNNs are texture sensitive, wire-framed and CAD models used in addition to real data can improve the performances of detectors. \citeauthor{peng2015LearningDeepObject} \cite{peng2015LearningDeepObject} augmented PASCAL-VOC data with 3D CAD models of the objects found in PASCAL-VOC (planes, horses, potted plants, etc.) and then rendered them in backgrounds where they are likely to be found and improved overall detection performances. Following this line, several authors introduced synthetic data for various tasks such as i) persons: \citeauthor{varol2017LearningSyntheticHumans} \cite{varol2017LearningSyntheticHumans} ii) furniture: \citeauthor{massa2016DeepExemplar2D3D} \cite{massa2016DeepExemplar2D3D} created rendered CAD furnitures on real backgrounds by using grayscale images to avoid color artifacts and improved the detection performances on the IKEA dataset. iii) text: \citeauthor{gupta2016SyntheticDataText} \cite{gupta2016SyntheticDataText} created an oriented text detection benchmark by superimposing synthetic text to existing scenes while respecting geometric and uniformity constraints and showed better results on ICDAR iv) logos: \citeauthor{su2017DeepLearningLogo} \cite{su2017DeepLearningLogo} did the same without any constraints by superimposing transparent logos to existing images. 

\citeauthor{georgakis2017SynthesizingTrainingData} \cite{georgakis2017SynthesizingTrainingData} synthesized new instances of 3D CAD models by copy pasting rendered objects on surface normals, very close to \citep{rajpura2017ObjectDetectionUsing}, which used Blender to put instances of objects inside a refrigerator. Later \citeauthor{dwibedi2017CutPasteLearn} \cite{dwibedi2017CutPasteLearn} with the same approach but without respecting any global consistency shown promise. For them only local consistency is important for modern object detectors. Similar to \citep{georgakis2017SynthesizingTrainingData}, they used different kinds of blending to make the detector robust to the pasting artifacts (more details can be found in \citep{dwibedi2017SynthesizingScenesInstance}). More recently, \citeauthor{dvornik2018ModelingVisualContext} \cite{dvornik2018ModelingVisualContext} extended \citep{dwibedi2017CutPasteLearn} by first finding locations in images with high likelihood of object presence before pasting objects. 
Another recent approach \citep{Tremblay_2018_CVPR_Workshops} found that domain randomization when creating synthetic data is vital to train detectors: training on Virtual KITTI \citet{Gaidon:Virtual:CVPR2016}, a dataset that was built to be close to KITTI (in terms of aspects, textures, vehicles and bounding boxes statistics), is not sufficient to be state-of-the-art on KITTI. One can gain almost one point of AP when building his own version of Virtual KITTI by introducing more randomness than was present in the original in the form of random textures and backgrounds, random camera angles and random flying distractor objects. Randomness was apparently absent from KITTI but is beneficial for the detector to gain generalization capabilities.

Several authors have shown interest in proposing tools for generating artificial images at a large scale. \citeauthor{qiu2016UnrealCVConnectingComputer} \cite{qiu2016UnrealCVConnectingComputer} created the open-source plug-in {\em UnrealCV} for a popular game engine Unreal Engine 4 and showed applications to deep network algorithms. \citeauthor{tian2017TrainingTestingObject} \cite{tian2017TrainingTestingObject} used the graphical model CityEngine to generate a synthetic city according to the layout of existing cities and added cars, trucks and buses to it using a game engine (Unity3D). The detectors trained on KITTI and this dataset are again better than just with KITTI. \citeauthor{alhaija2017AugmentedRealityMeets} \cite{alhaija2017AugmentedRealityMeets} pushed Blender to its limits to generate almost real-looking 3D CAD cars with environment maps and pasted them inside different 2D/3D environments including KITTI, VirtualKITTI (and even Flickr). It is worth noting that some datasets included real images to better simulate the scene viewed by a robot in active vision settings, as in~\citep{ammirato2017DatasetDevelopingBenchmarking}.

Another strategy is to render simple artificial images and increase the realism of the images in a second iteration, using Generative Adversarial Networks~\citep{shrivastava2017LearningSimulatedUnsupervised}. RenderGAN was used to directly generate realistic training images~\citep{sixt2018rendergan}. We refer the reader to the section on GANs (Section~\ref{sec:gan}) for more information on the use of GANs for style transfer.

We have seen that for the time being synthetic datasets can augment existing ones but not totally replace them for object detection, however, the domain shift between synthetic data and the target distribution is still too large to rely on synthetic data only.

\subsubsection{Object Localization}
Accurate localization remains one of the two biggest sources of error~\citep{hoiem2012diagnosing} in fully supervised object detection. It mainly originates from small objects and more stringent evaluation protocol applied in the latest datasets. The predicted boxes are required to have an IoU of up to 0.95 with the ground-truth boxes. Generally, localization is dealt by using smooth L1 or L2 losses along with classification loss. Some papers proposed a more detailed methodology to overcome this issue. Also, annotating bounding boxes for each and every object is expensive. We will also look into some methods that localize objects using only weakly annotated images.

\citeauthor{kong2016HyperNetAccurateRegion} \cite{kong2016HyperNetAccurateRegion} overcame the poor localization because of coarseness of the feature maps by aggregating hierarchical feature maps and then compressing them into a uniform space. It provided an efficient combination framework for deep but semantic, intermediate but complementary, and shallow but high-resolution CNN features. \citeauthor{chen2015ImprovingObjectProposals} \cite{chen2015ImprovingObjectProposals} proposed multi-thresholding straddling expansion (MTSE) to reduce localization bias and refine boxes during proposal time which is based on super-pixel tightness as opposed to objectness based models. \citeauthor{zhang2015ImprovingObjectDetection} \cite{zhang2015ImprovingObjectDetection} addressed the localization problem by using a search algorithm based on Bayesian optimization that sequentially proposed candidate regions for an object bounding box. \citeauthor{hosang2017learning} \cite{hosang2017learning} tried to integrate NMS in the convolutional network which in the end improved localization. 

Many papers~\citep{gidaris2016AttendRefineRepeat,jiang2018acquisition} also try to adapt the loss function to address the localization problem. \citeauthor{gidaris2016LocNetImprovingLocalization} \cite{gidaris2016LocNetImprovingLocalization} proposed to assign conditional probabilities to each row and column of a sample region, using a neural convolutional network adapted for this task. These probabilities allow more accurate inference of the object bounding box under a simple probabilistic framework. Since Intersection over Union (IoU) is used in the evaluation strategies of many detection challenges, \citeauthor{yu2016UnitBoxAdvancedObject} \cite{yu2016UnitBoxAdvancedObject} and \citeauthor{jiang2018acquisition} \cite{jiang2018acquisition} optimized over IoU directly. The loss-based papers have been discussed in Section~\ref{subsec:losses} in detail.

There is also an interesting case made by some papers that do we really need to optimize for localization? \citeauthor{oquab2015isobject} \cite{oquab2015isobject} used weakly annotated images to predict approximate locations of the object. Their approach performed comparably to the fully supervised counterparts. \citeauthor{zhou2015LearningDeepFeatures} \cite{zhou2015LearningDeepFeatures} were able to get localizable deep representations that exposed the implicit attention of CNNs on an image with the help of global average pooling layers. In comparison to the earlier approach, their localization is not limited to localizing a point lying inside an object but determining the full extent of the object. \citep{zhou2015ObjectDetectorsEmerge,DBLP:conf/eccv/ZeilerF14,bazzani2016SelftaughtObjectLocalization} have also tried to predict localizations by masking different patches of the image during test time. More weakly supervised methods have been discussed in Section~\ref{subsec:weakly_supervised}.

\subsubsection{Occlusions}
The occlusions lead to partial missing information from object instances. They may be occluded due to the background or other object instances. Less information naturally leads to harder examples and inaccurate localizations. The occlusions happen all the time in real-life images. However, since deep learning is based on convoluting filters and that occlusions by definition introduce parasite patterns most modern methods are not robust to it by construction. 

Training with occluded objects help for sure \citep{mitash2017PhysicsawareSelfsupervisedTraining} but it is often not doable because of a lack of data and furthermore, it cannot be bulletproof. \citeauthor{wu2016LearningAndOrModel} \cite{wu2016LearningAndOrModel} managed to learn an And-Or model for cars by dynamic programming, where the And stood for the decomposition of the objects into parts and the Or for all different configurations of parts (including occluded configurations). The learning was only possible thanks to the heavy use of synthetic data to model \emph{every possible type} of occlusion. Another way to generate examples of occlusions is to directly learn to mask the proposals of Fast R-CNN \citep{wang2017AFastRCNNHardPositive}.

For dense pedestrians crowds deformable models and parts can help improve detection accuracy (see \ref{sec:dpm}) \eg if some parts are masked some others will not be, therefore, the average score is diminished but not made zero like in \citep{ouyang2013JointDeepLearning,savalle2014DeformablePartModels,girshick2015DeformablePartModels}. Parts are also useful for occlusion handling in face detection where different CNNs can be trained on different facial parts \citep{yang2015FacialPartsResponses}. The survey already tackled Deformable RoI-Pooling (RoI-Pooling with parts) \citep{mordan2017DeformablePartbasedFully}. Another way of re-introducing parts in modern pipelines is the deformable kernels of \citep{dai2017deformable}. They presented a way to alleviate the occlusion problems by giving more flexibility to the usually fixed geometric structures.

Building special kinds of regression losses for bounding boxes acknowledging the proximity of each detection (which is reminiscent of the springs in the old part-based models) was done in \citep{wang2018RepulsionLossDetecting}. They, in addition, to the attraction term in the traditional regression loss that pushes predictions towards their assigned ground truth added a repulsion term that pushed predictions away from each other.

Traditional non-maximum suppression causes a lot of problems with occlusions because overlapping boxes are suppressed. Hence, if one object is in front of another only one is detected. To address this, \citeauthor{hosang2017learning} \cite{hosang2017learning} offered to learn non-maximum suppression making it continuous (and differentiable) and \citeauthor{bodla2017soft} \cite{bodla2017soft} used a soft version that only degraded the score of the overlapping objects (more details can be found about various other types of NMS in Section~\ref{sec:inference}. 

Other approaches used clues and context to help infer the presence of occluded objects. \citeauthor{zhang2018OccludedPedestrianDetection} \cite{zhang2018OccludedPedestrianDetection} used super-pixel labeling to help occluded objects detection. They hypothesized that if some pixels are visible then the object is there. This is also the approach of the recent \citep{he2017mask} but it needs pixel-level annotations. In videos, temporal coherence can be used \citep{yoshihashi2017LearningMultiframeVisual}, where heavily occluded objects are not occluded in every frame and can be tracked to help detection. 

But for now all the solutions seem to be far-off from the mentally inpainting ability of humans to infer missing parts. Using GANs for this purpose might be an interesting research direction.

\subsubsection{Detecting Small Objects}
\label{sec:small_objects}

Detecting small objects is harder than detecting medium sized and large sized objects because of less information associated with them, easier possibility of confusion with the background, higher precision requirement for localization, large image size, etc. In COCO metrics evaluation, objects occupying areas lesser than and equal to $32 \times 32$ pixels come under this category and this size threshold is generally accepted within the community for datasets related to common objects. Datasets related to aerial images \citep{xia2017dota}, traffic signs \citep{zhu2016traffic}, faces \citep{nada2018PushingLimitsUnconstraineda}, pedestrians \citep{enzweiler2008monocular} or logos \citep{su2017WebLogo2MScalableLogo} are generally abundant with small object instances. 

In case of objects like logos or traffic signs, objects have an expected shape, size and aspect ratio of the objects to be detected, and this information can be embedded to bias the deep learning model. This strategy is much harder and not feasible for common objects as they are a lot more diverse. As an illustration, the winner of the COCO challenge 2017 \citep{peng2017megdet}, which used many of the latest techniques and ensemble of four detectors reported a performance of 34.5\% mAP on small objects and 64.9\% mAP on large objects. The following entries reported even a greater dip for smaller objects than the larger ones. \citeauthor{pham2017EvaluationDeepModels} \cite{pham2017EvaluationDeepModels} have presented an evaluation, focusing on real-time small object detection, of three state-of-the-art models, YOLO, SSD and Faster R-CNN with related trade-off between accuracy, execution time and resource constraints.

There are different ways to tackle this problem, such as: i) up-scaling the images ii) shallow networks, iii) contextual information, iv) super-resolution. These four directions are discussed in the following.

The first -- and most trivial direction -- consists in up-scaling the image before detection. But a naive upscaling is not efficient as the large images become too large to fit into a GPU for training. \citeauthor{gao2017DynamicZoominNetwork} \cite{gao2017DynamicZoominNetwork}, first, down-sampled the image and then used reinforcement learning to train attention-based models to dynamically search for the interesting regions in the image. The selected regions are then studied at higher resolution and can be used to predict smaller objects. This avoided the need of analyzing each pixel of the image with equal attention and saved some computational costs. Some papers \citep{dai2017deformable,dai2016r,singh2018analysis} used image pyramids during training time in the context of object detection while \citep{DBLP:journals/pami/RenHGZS17} used it during inference time.

The second direction is to use shallow networks. Small objects are easier to predict by detectors which have smaller receptive field. The deeper networks with their large receptive field tend to lose some information about the small objects in their coarser layers. \citeauthor{sommer2017FastDeepVehicle} \cite{sommer2017FastDeepVehicle,ogier2017icip} proposed very shallow networks with less than 5 convolutional layers and three fully connected layers for the purpose of detecting objects in aerial imagery. Such type of detectors are useful when the expected instances are only of type small. But if expected instances are of diverse size it is more beneficial to use finer feature maps of very deep networks for small objects and coarser feature maps for larger objects. We have already discussed this approach in Section~\ref{subsec:scalevariance}. Please refer to Section~\ref{subsec:fastandlowpower} for more low power and shallow detectors.

The third direction is to make use of context surrounding the small object instances. \citet{gidaris2015object,zhu2015SegDeepMExploitingSegmentation} used context to improve the performance but \citeauthor{chen2016RCNNSmallObject} \cite{chen2016RCNNSmallObject} used context specifically for improving the performance for small objects. They augmented the R-CNN with the context patch in parallel to the proposal patch generated from region proposal network. \citeauthor{zagoruyko2016multipath} \cite{zagoruyko2016multipath} combined their approach of making the information flow through multiple paths with DeepMask object proposals \citep{pinheiro2015LearningSegmentObject,pinheiro2016learning} to gain a massive improvement in the performance for small objects. Context can also be used by fusing coarser layers of the network with finer layers \citep{lin2017feature,shrivastava2016beyond,lin2017focal}. Context related literature has been covered in Section~\ref{subsec:context} in detail.

Finally, the last direction is to use Generative Adversarial Networks to selectively increase the resolution of small objects, as proposed by \citeauthor{li2017PerceptualGenerativeAdversarial} \cite{li2017PerceptualGenerativeAdversarial}. Its generator learned to enhance the poor representations of the small objects to super-resolved ones that are similar enough to real large objects to fool a competing discriminator.
Table~\ref{table:sumcha} summarizes the past subsection by grouping the articles by main idea and target challenge. We find it very useful to see which ideas have been thoroughly investigated by the literature and which are under-explored.

 \begin{table*}
 \resizebox{\textwidth}{!}{
 \begin{tabular}{|p{3cm}|c|c|}\hline
 Article references&  Main idea& Challenge(s) addressed \\\hline\hline
\citep{he2015spatial,girshick2015fast,Felzenszwalb2010Object,sermanet2013overfeat} & Image Pyramids & Scale Variance, Small Objects\\\hline

\citep{dollar2014FastFeaturePyramids, liu2016ssd, cai2016unified, yang2016wider, shrivastava2016beyond, kong2016HyperNetAccurateRegion, zagoruyko2016multipath, woo2017stairnet, najibi2017SSHSingleStage, li2017scale, ess2007depth, DBLP:journals/pami/RenHGZS17,lin2017feature, shen2017LearningObjectDetectors, singh2018analysis,singh2018sniper} & Features Fusion  & Scale Variance, Small Objects\\\hline

\citep{singh2018analysis, singh2018sniper} &Selective Backpropagation (SN) & Scale Variance, Small Objects, Object Localization\\\hline

\citep{bodla2017soft, tychsen2018fitness}&Better NMS & Small Objects, Occlusions, Object Localization\\\hline

\citep{shrivastava2016training, lin2017focal} & Hard Examples Mining (Explicit and Implicit) & Small Objects, Occlusions\\\hline

\citep{yang2016exploit}&Scale Dependent Pooling & Scale Variance, Small Objects\\\hline

\citep{jiang2017R2CNNRotationalRegion, michalbusta2017DeepTextSpotterEndToEnd, ma2018ArbitraryOrientedSceneText,he2018SingleShotTextSpotter}& Oriented Bounding boxes & Rotational Variance \\\hline

\citep{michalbusta2017DeepTextSpotterEndToEnd, ma2018ArbitraryOrientedSceneText,he2018SingleShotTextSpotter}&Oriented Pooling & Rotational Variance\\\hline

\citep{shi2017detecting, liu2017DeepMatchingPrior} & Flexible anchors (segments, quadrangles) & Rotational Variance \\\hline

\citep{liao2018RotationSensitiveRegressionOriented} & Rotating Filters & Rotational Variance \\\hline

\citep{cheng2016RIFDCNNRotationInvariantFisher,laptev2016TIPOOLINGTransformationinvariantPooling} & Rotation Invariant Features & Rotational Variance \\\hline

\citep{zhang2016MultiorientedTextDetectionentedTextDetectionentedTextDetection,he2017mask} & Auxiliary Task (semantic segmentation) & Rotational Variance, Occlusions \\\hline

\citep{raj2015SubspaceAlignmentBased, chen2018DomainAdaptiveFaster} & Aligning Feature distribution & Domain Adaptation \\\hline

\citep{shrivastava2017LearningSimulatedUnsupervised, inoue2018cross} & Image Transformations (GANs) & Domain Adaptation \\\hline

\citep{pepik2015holding,peng2015LearningDeepObject,georgakis2017SynthesizingTrainingData,rajpura2017ObjectDetectionUsing,dwibedi2017CutPasteLearn} & Data Augmentation using Synthetic Datasets & Domain Adaptation \\\hline 

\citep{Tremblay_2018_CVPR_Workshops} & Domain Randomization & Domain Adaptation \\\hline 

\citep{chen2015ImprovingObjectProposals, zhang2018OccludedPedestrianDetection} & Super-Pixels & Object Localization, Occlusions \\\hline

\citep{zhang2015ImprovingObjectDetection} & Sequential Search & Object Localization \\\hline

\citep{gidaris2016LocNetImprovingLocalization, gidaris2016AttendRefineRepeat, lin2017focal, jiang2018acquisition,yu2016UnitBoxAdvancedObject}  & Loss Function Modifications  & Small Objects, Object Localization \\\hline

\citep{ouyang2013JointDeepLearning,savalle2014DeformablePartModels,yang2015FacialPartsResponses,girshick2015DeformablePartModels,dai2017deformable, mordan2017DeformablePartbasedFully} & Part Based Models & Occlusions \\\hline

\citep{dai2017deformable, mordan2017DeformablePartbasedFully} & Deformable CNN Modules & Occlusions \\\hline

\citep{yoshihashi2017LearningMultiframeVisual} & Tracking (in videos) & Occlusions \\\hline

\citep{gao2017DynamicZoominNetwork} & Dynamic Zooming & Small Objects \\\hline

\citep{sommer2017FastDeepVehicle,ogier2017icip} & Shallow Networks & Small Objects \\\hline

\citep{gidaris2015object,zhu2015SegDeepMExploitingSegmentation,chen2016RCNNSmallObject} & Use of Contextual Information & Small Objects \\\hline

\citep{li2017PerceptualGenerativeAdversarial} & Features Super Resolution  & Small Objects \\\hline

 \end{tabular}
}

\caption{Summary of the main ideas found in the literature to account for the limitations of the current deep learning architectures. For each idea we list the papers that implement it and the challenges they (sometimes only partially) address.\label{table:sumcha}}
\end{table*}

\subsection{Complementary New Ideas in Object Detection}
In this subsection we review ideas which haven't quite matured yet but we feel could bring major breakthroughs in the near future. If we want the field to advance, we should embrace new grand ideas like these, even if that means completely rethinking all the architectural ideas evoked in Section~\ref{sec:detector_design}.

\begin{figure}
    \centering
    \includegraphics[scale=0.18]{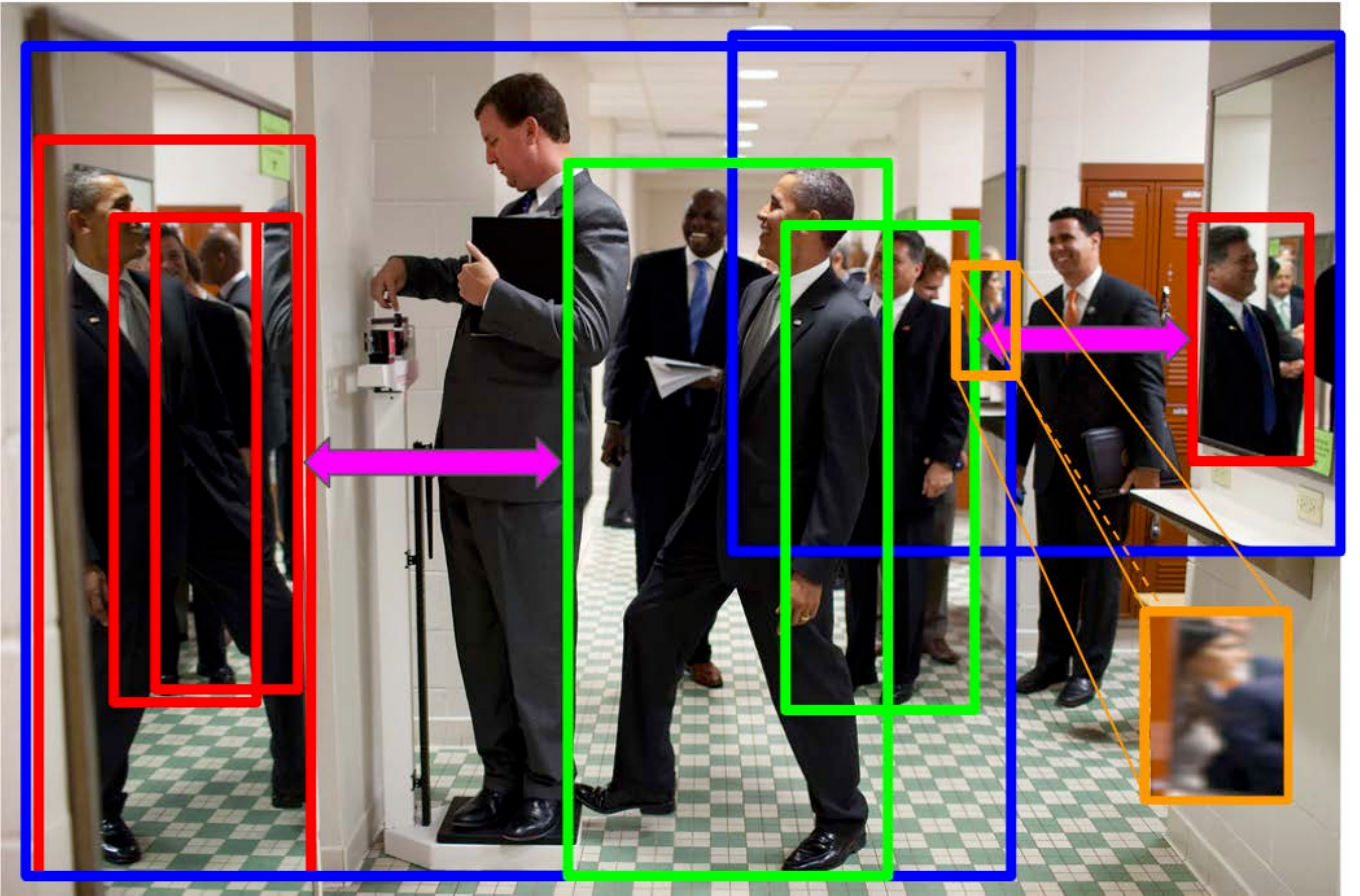}
    \caption{Good detections of persons are marked in green and bad detections in red. Helpful context in blue (the presence of mirror frames) can help lower the score of a box. The relationships (in Fuchsia) between bounding boxes can also help: a person cannot be present twice in a picture. Enhancing parts of the picture using SR ( Figure~\ref{fig:gansr}) is yet another way to better make a decision. All those "reasoning" modules are not included in the mainstream detectors.}
    \label{fig:newideas}
\end{figure}
\subsubsection{Graph Networks}
\label{sec:graph}
The dramatic failings of state-of-the-art detectors on perturbed versions of the COCO validation sets, spotted by 
\citeauthor{rosenfeld2018elephant} \cite{rosenfeld2018elephant}, are raising questions for better understanding of compositionality, context and relationships in detectors.

\citeauthor{Battaglia2018RelationalIB} \cite{Battaglia2018RelationalIB} recently wrote a position article arguing about the need to introduce more representational power into Deep Learning using graph networks. It means finding new ways to enforce the learning of graph structures of connected entities instead of outputting independent predictions. Convolutions are too local and translation equivariant to reflect the intricate structure of objects in their context.

One embodiment of this idea in the realm of detection can be found in the work of \citeauthor{wang2017NonlocalNeuralNetworks} \cite{wang2017NonlocalNeuralNetworks}, where long-distance dependencies were introduced in deep-learning architectures. These combined local and non-local interactions are reminiscent of the CRF~\citep{Lafferty:2001:CRF:645530.655813}, which sparked a renewed interest for graphical models in 2001. Dot products between features determine their influences on each other, the closest they are in the feature space, the stronger their interactions will be (using a Gaussian kernel for instance). This seems to go against the very principles of DCNNs, which are, by nature, local. However this kind of layer can be integrated seamlessly in any DCNN to its benefit, it is very similar to self-attention \citep{cheng2016long}. It is not clear yet if these new networks will replace their local counterparts in the long-term but they are definitely suitable candidates.

Graph structures also emerge when one needs to incorporate a priori (or inductive biases) on the spatial relationships of the objects to detect (relational reasoning) \citep{hu2018RelationNetworksObject}. The relation module uses attention to learn object dependencies, also using dot products of features. Similarly, \citeauthor{wang2017NonlocalNeuralNetworks} \cite{wang2017NonlocalNeuralNetworks} incorporated geometrical features to further disambiguate relationships between objects. One of the advantages of this pipeline is the last relation module, which is used to remove duplicates similarly to the usual NMS step but adaptively. We mention this article in particular because although relationships between detected objects have been used in the literature before, it was the first attempt to have it as a differentiable module inside a CNN architecture.

\subsubsection{Adversarial Trainings}
\label{sec:gan}
No one in the computer vision community was spared by the amazing successes of the Generative Adversarial Networks \citep{goodfellow2014generative}. By pitting a con-artist (a CNN) against a judge (another CNN) one can learn to generate images from a target distribution up to an impressive degree of realism. This new tool keeps the flexibility of the regular CNN architectures as it is implemented using the same bricks and therefore, it can be added in any detection pipeline.

\begin{figure}
    \centering
    \includegraphics[width=\columnwidth]{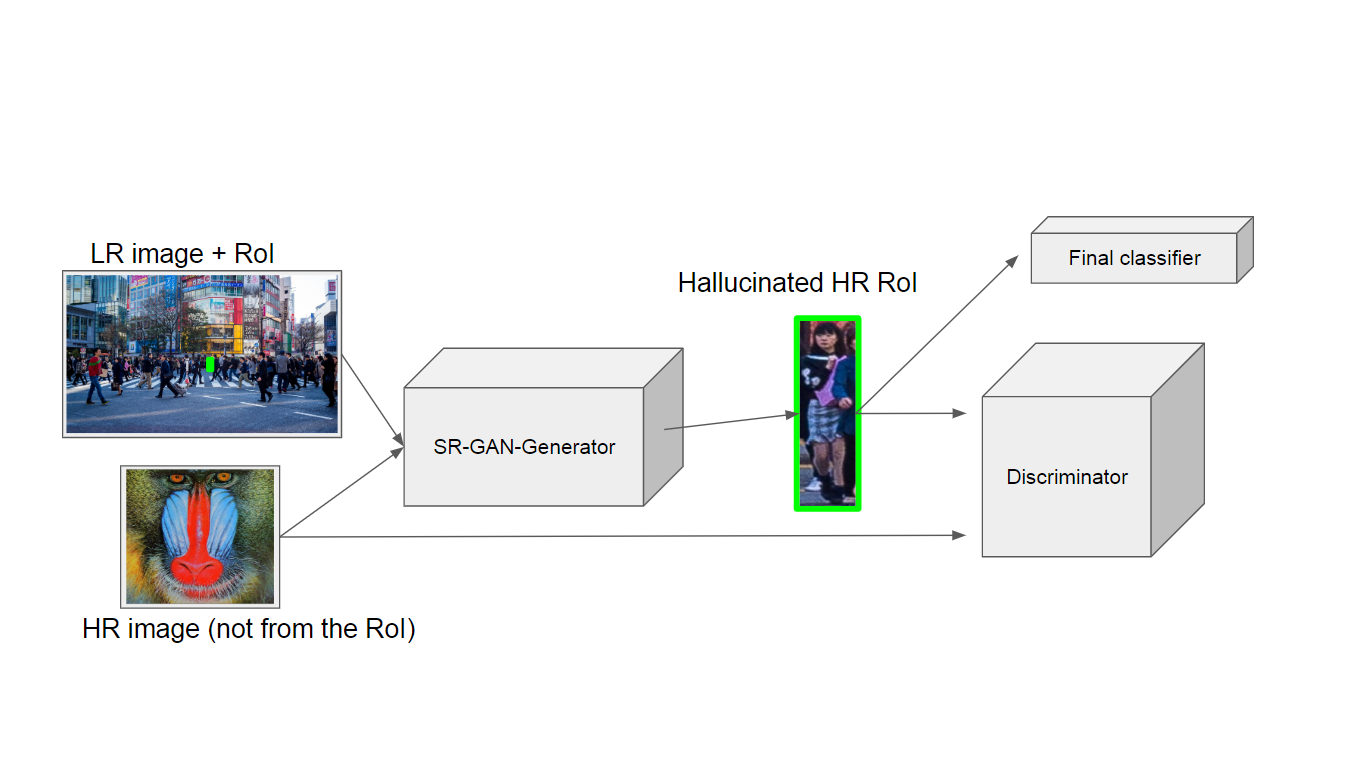}
    \caption{Small object patches from Regions of Interest are enhanced to better help the classifier make a decision in SOD-MTGAN~\cite{bai2018SODMTGANSmallObject}.
    \label{fig:gansr}}
\end{figure}

Even if \citep{wang2017AFastRCNNHardPositive} does not belong to the GAN family per say, the adversarial training it uses: dropping pixels in examples to make them harder to classify and hence, render the network robust to occlusions, obviously drew its inspiration from GANs. \citeauthor{ouyang2018PedestrianSynthesisGANGeneratingPedestrian} \cite{ouyang2018PedestrianSynthesisGANGeneratingPedestrian} went a step further and used the GAN formalism to learn to generate pedestrians from white noise in large images and showed how those created examples were beneficial for the training of object detectors.
There are numerous recent papers, \eg, \citep{peng2017SyntheticRealAdaptation,bousmalis2017UnsupervisedPixelLevelDomain}, proposing approaches for converting synthetic data towards more realistic images for classification. \citeauthor{inoue2018cross} \cite{inoue2018cross} used the latest CycleGAN~\citep{zhu2017UnpairedImagetoImageTranslation} to convert real images to cartoons and by doing so gained free annotations to train detectors on weakly labeled images and became the first work to use GANs to create full images for detectors.
As stated in the introduction, GANs can also be used, not in a standalone manner but, directly embedded inside a detector too:
\citeauthor{li2017PerceptualGenerativeAdversarial} \cite{li2017PerceptualGenerativeAdversarial} operated at the feature level by adapting the features of small objects to match features obtained with well resolved objects. \citeauthor{bai2018SODMTGANSmallObject} \cite{bai2018SODMTGANSmallObject} trained a generator directly for super-resolution of small objects patches using traditional GAN loss in addition to classification losses and MSE loss per pixel. Integrating the module in modern pipelines brought improvement to the original mAP on COCO, this very simple pipeline is summarized Figure~\ref{fig:gansr}. 
These two articles addressed the detection of small objects, which will be tackled in more details in Section~\ref{sec:small_objects}.

\citet{Shen_2018_CVPR} used GANs to completely replace the Multiple Instance Learning paradigm (see Section \ref{subsec:weakly_supervised}) using the GAN framework to generate candidate boxes following the real distribution of the training images boxes and built a state-of-the-art detector that is faster than all the others by two orders of magnitude.

Thus, this extraordinary breakthrough is starting to produce interesting results in object detection and its importance is growing. Considering the latest result in the generation of synthetic data using GANs for instance the high resolution examples of \citep{karras2018progressive} or the infinite image generators, BiCycleGAN from \citeauthor{zhu2017UnpairedImagetoImageTranslation} \cite{zhu2017UnpairedImagetoImageTranslation} and MUNIT from \citet{DBLP:journals/corr/abs-1804-04732}, it seems the tsunami that started in 2014 will only get bigger in the years to come. 

\subsubsection{Use of Contextual Information}
\label{subsec:context}
We will see in this section that the word context can mean a lot of different things but taking it into account gives rise to many new methods in object detection. Most of them (like spatial relationships or using {\em stuff} to find {\em things}) are often overlooked in competitions, arguably for bad reasons (too complex to implement in the time frame of the challenge). 

Methods have evolved a lot since \citeauthor{heitz2008LearningSpatialContext} \cite{heitz2008LearningSpatialContext} used clustering of stuff/backgrounds to help detect objects in aerial imagery. Now, thanks to the CNN architectures, it is possible to do detection of things and stuff segmentation in parallel, both tasks helping the other \citep{brahmbhatt2017StuffNetUsingStuff}. 

Of course, this finding is not surprising. Certain objects are more likely to appear in certain stuff or environments (or context): thanks to our knowledge of the world, we find it weird to have a flying train: \citeauthor{katti2016ObjectDetectionCan} \cite{katti2016ObjectDetectionCan} showed that adding this human knowledge helps existing pipelines. 
The environments of the visual objects also comprise of other objects that they are present with, which advocates for learning spatial relationships between objects.
\citeauthor{mrowca2015spatial} \cite{mrowca2015spatial} and \citeauthor{gupta2015ExploringPersonContext} \cite{gupta2015ExploringPersonContext} independently used spatial relationships between proposals and classes (using WordNet hierarchy) to post-process detections. This is also the case in \citep{zuo2016LearningContextualDependence} where RNNs were used to model those relationships \emph{at different scales} and in \citep{chen2017SpatialMemoryContext} where an external memory module was keeping track of the likelihood of objects being together. \citeauthor{hu2018RelationNetworksObject} \cite{hu2018RelationNetworksObject}, that we mentioned in Section \ref{sec:graph}, went even further with a trainable relation module inside the structure of the network.
In a different but not unrelated manner \citet{gonzalez-garcia2017ObjectsContextPart} improved the detection of parts of objects by associating parts with their root objects.

All multi-scale architectures use different sized context, as we saw in Section~\ref{sec:backbone}. \citeauthor{zeng2016GatedBidirectionalCNN} \cite{zeng2016GatedBidirectionalCNN} used features from different sized regions (different contexts) in different layers of the CNN with message-passing in between features related to different context. \citeauthor{kong2017RONReverseConnection} \cite{kong2017RONReverseConnection} used skip connections and concatenation directly in the CNN architecture to extract multi-level and multi-scale information. 

\begin{table*}
\resizebox{\textwidth}{!}{
\begin{tabular}{|c|c|c|}\hline
Article references& Ideas & Type of Context \\\hline\hline
\citep{brahmbhatt2017StuffNetUsingStuff, katti2016ObjectDetectionCan} & Segmenting Stuff/Using background cues   & Background context (Environment) \\\hline
\citep{mrowca2015spatial, gupta2015ExploringPersonContext,zuo2016LearningContextualDependence,yu2016RoleContextSelection, chen2017SpatialMemoryContext,hu2018RelationNetworksObject} & Likelihood of Objects being together/Infering Relationships/Memory Modules & Other Objects Context \\\hline
\citep{gonzalez-garcia2017ObjectsContextPart} & Finding Root to find parts & Parts-Objects Context \\\hline
\citep{zeng2016GatedBidirectionalCNN, kong2017RONReverseConnection} & Using dfferent feature scales& Multi-scale Context \\\hline
\citep{ouyang2017LearningChainedDeep, chen2016RCNNSmallObject, gidaris2016LocNetImprovingLocalization,gidaris2016AttendRefineRepeat, li2016ObjectDetectionEndtoEnd} & Adding variable sized context/Adding Borders of RoI & Surrounding Pixels \\\hline
\citep{redmon2016you,wang2017NonlocalNeuralNetworks,li2018RFCNAccurateRegionBased} & Adding connections to all pixels & Full Image Context \\\hline

\end{tabular}
}
\caption{Summary of the approaches taken to exploit different types of context. \label{table:sumcont}}
\end{table*}

Sometimes, even the simplest local context surrounding a region of interest can help (see, for instance, the methods presented in Section~\ref{sec:boost}, where the amount of context varies in between the classifiers).
Extracted proposals can include variable amounts of pixels (context means size of the proposal) to help the classifiers such as in \citep{ouyang2017LearningChainedDeep} or in \citep{chen2016RCNNSmallObject, gidaris2016LocNetImprovingLocalization,gidaris2016AttendRefineRepeat}. \citeauthor{li2016ObjectDetectionEndtoEnd} \cite{li2016ObjectDetectionEndtoEnd} included global image context in addition to regional context.
Some approaches went as far as integrating all the image context: it was done for the first time in YOLO \citep{redmon2016you} with the addition of a fully connected layer on the last feature map. \citeauthor{wang2017NonlocalNeuralNetworks} \cite{wang2017NonlocalNeuralNetworks} modified the convolutional operator to put weights on every part of the image, helping the network use context outside the object to infer their existence. This use of global context is also found with the Global Context Module of the recent detection pipeline from Megvii \citep{peng2017large}. \citeauthor{li2018RFCNAccurateRegionBased} \cite{li2018RFCNAccurateRegionBased} proposed a fully connected layer on all the feature maps (similar to \citeauthor{redmon2016you} \cite{redmon2016you}) with dilated kernels. 

Other kinds of context can also be put to work. \citeauthor{yu2016RoleContextSelection} \cite{yu2016RoleContextSelection} used latent variables to decide on which context cues to use to predict the bounding boxes. It is not clear yet which method is the best to take context into account, another question is: do we want to? Even if the presence of an object in a context is unlikely, do we actually want to blind our detectors to unlikely situations?
All the types of context that can be leveraged, have been summarized in Table~\ref{table:sumcont}.

\subsection{Concluding Remarks}
This section finished the tour of all the principal CNN based approaches past, present and future that treat general object detection in the traditional settings. It has allowed to peer through the armor of the CNN detectors and see them for what they are: impressive machines having amazing generalization capabilities but still powerless in a variety of cases, in which a trained human would have no problem (domain adaptation, occlusions, rotations, small objects) for an example of a difficult test case even for so-called robust detectors see Figure~\ref{fig:newideas}. 
Potential ideas to go past these obstacles have also been mentioned among them the use of adversarial training and context are the most prominent.
The following section will go into more specific set-ups, less traditional problems or environments that will frame the detector abilities even further.

%% file: article_extensions.tex
\section{Extending Object Detection}
\label{sec:extensions}
Object detection may still feel like a narrow problem: one has a big training set of 2D images, huge resources (GPUs, TPUs, etc.) and wants to output 2D bounding boxes on a similar set of 2D images. However, these basic assumptions are often not present in practical scenarios. Firstly, because there exists many other modalities where one can perform object detection. These require conceptual changes in architectures to perform equally well. Secondly, sometimes one might be constrained to learn from exceedingly few fully annotated images, therefore, training a regular detector is either irrelevant or not an optimal choice because of overfitting. Also detectors are not built to be run in research labs alone but to be integrated into industrial products, which often come with an upper bound on energy consumption and speed requirements to satisfy the customer. The aim of the following discussion will be to know more about the research work done to extend the deep learning based object detection into new modalities and with tough constraints. It ends with reflections on what other interesting functionalities a strong detector in the future might possess.

\subsection{Detecting Objects in Other Modalities}
\label{sec:detecting_in_other_modalities}
There are several modalities other than 2D images that can be interesting: videos, 3D point clouds, medical imaging, hyper-spectral imagery, etc. We will be discussing in this survey the former two. We did not treat for instance the volumetric images from the medical domain (MRI, etc.) or hyper-spectral imagery, which are outside of the scope of this article and would deserve their own survey.
\begin{figure}
    \centering
    \includegraphics[scale=0.18, trim={0.5cm, 0.1cm, 0.5cm, 0.5cm}, clip]{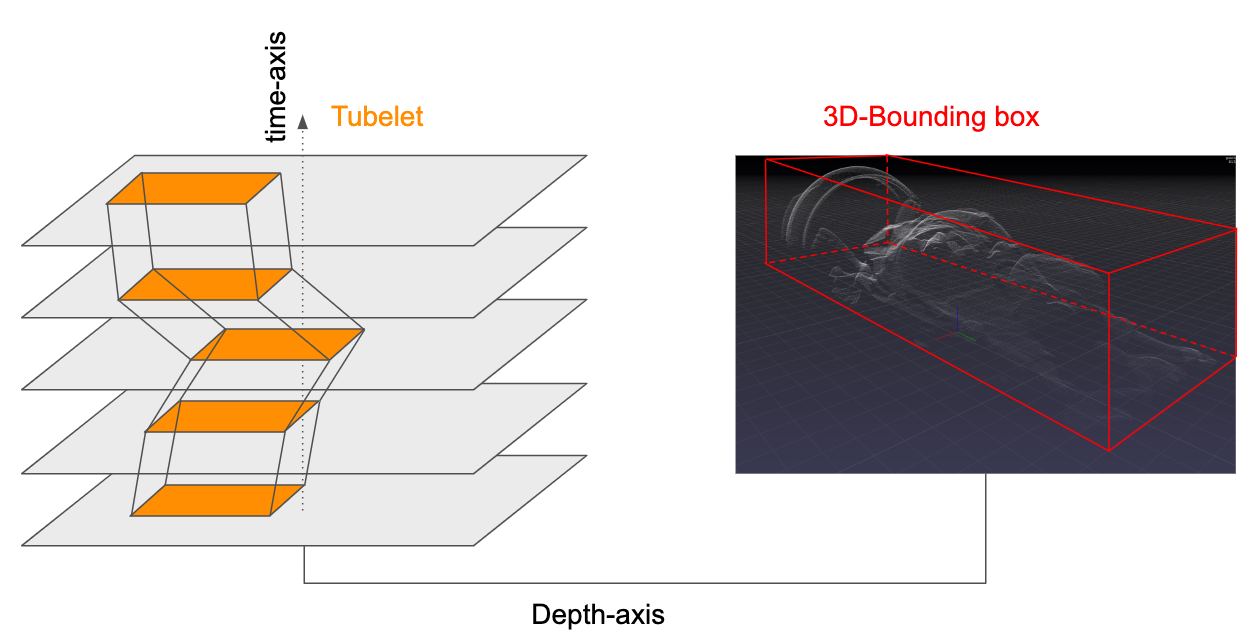}
    \caption{Detecting objects in other modalities: \textbf{Left} videos. \textbf{Right} 3D point-clouds.} 
    \label{fig:my_label}
\end{figure}

\subsubsection{Object Detection in Videos}
\label{sec:odinvideos}
The upside of detecting objects in videos is that it provides additional temporal information but it also has unique challenges associated with it: motion blur, appearance changes, video defocus, pose variations, computational efficiency \etc It is a recent research domain due to the lack of large scale public datasets. One of the first video datasets is the ImageNet VID \citep{russakovsky2015ImageNetLargeScale}, proposed in 2015. This dataset as well as the recent datasets for object detection in video are mentioned in Section~\ref{sec:videodatasets}.

One of the simplest ways to use temporal information for detecting object is the detection by tracking paradigm. As an example, \citeauthor{ray2017ObjectDetectionSpatioTemporal} \cite{ray2017ObjectDetectionSpatioTemporal} proposed a spatio-temporal detector of motion blobs, associated into tracks by a tracking algorithm. Each track is then interpreted as a moving object. Despite its simplicity, this type of algorithm is marginal in the literature as it is only interesting when the appearances of the objects are not available. 
 
The most widely used approaches in the literature are those relying on {\em tubelets}. Tubelets have been introduced in the T-CNN approach of \citet{kang2017TCNNTubeletsConvolutional,kang2016object}. T-CNN relied on 4 steps. First, still-image object detection (with Faster R-CNN like detectors) was performed. Second, multi-context suppression removed detection hypotheses having the lowest scores: highly ranked detection scores were treated as high-confidence classes and the rest were suppressed. Third, motion-guided propagation transferred detection results to adjacent frames to reduce false negatives. Fourth, temporal tubelet rescoring used a tracking algorithm to obtain sequences of bounding boxes, classified into positive and negative samples. Positive samples were mapped to a higher range, thus, increasing the score margins. T-CNN has several follow ups. The first was Seq-NMS \citep{han2016SeqNMSVideoObject} which constructed sequences along nearby high-confidence bounding boxes from consecutive frames, rescoring to the average confidence. Other boxes close to this sequence were suppressed. Another one was MCMOT \citep{lee2016MultiClassMultiObjectTracking} in which a post-processing stage, under the form of a multi-object tracker, was introduced, relying on hand-crafted rules (e.g., detector confidences, color/motion clues, changing point detection and forward-backward validation) to determine whether bounding boxes belonged to the tracked objects, and to further refine the tracking results. \citeauthor{tripathi2016context} \cite{tripathi2016context} exploited temporal information by training a recurrent neural network that took as input, sequences with predicted bounding boxes, and optimized an objective enforcing consistency across frames. 

The most advanced pipeline for object detection in videos is certainly the approach of \citeauthor{feichtenhofer2017detect} \cite{feichtenhofer2017detect}, borrowing ideas from tubelets as well as from feature aggregation. The approach relies on a multitask objective loss, for frame-based object detection and across-frame track regression, correlating features that represented object co-occurrences across time and linking the frame level detections based on across-frame tracklets to produce the detections. 

 \begin{table*}
 \resizebox{\textwidth}{!}{
\begin{tabular}{|c|c|c|}\hline
Article references& Highlight & Type of Detections \\\hline\hline
\citep{lee2016MultiClassMultiObjectTracking, ray2017ObjectDetectionSpatioTemporal} & Context cues/Motion blobs & Basic Tracking \\\hline
\citep{kang2017TCNNTubeletsConvolutional,kang2016object,han2016SeqNMSVideoObject,feichtenhofer2017detect} & Motion Propagation / Tracking / Seq-NMS / Feature aggregation  & Tubelets \\\hline
\citep{tripathi2016context} & Enforcing Consistency  & RNNs \\\hline
\citep{zhu2017deep,zhu2018HighPerformanceVideo} & Sparse Key frame aggregation / Fast Computation of flow & Flow Field \\\hline
\citep{shafiee2017FastYOLOFast,chen2018optimizing} & Motion-based Inference & Adaptive Computation \\\hline
\end{tabular}
}
\caption{Summary of the video object detection methods.\label{table:sumvid}}
\end{table*}

The literature on object detection in videos also addressed the question of computing time, since applying a detector on each frame can be time consuming. In general, it is non-trivial to transfer the state-of-the-art object detection networks to videos, as per-frame evaluation is slow. Deep feature flow \citep{zhu2017deep, zhu2018flow} ran the convolutional sub-network only on sparse key frames, propagated deep feature maps to other frames via a flow field. It led to significant speedup as flow computation is relatively fast. In the {\em impression network} \citep{hetang2017ImpressionNetworkVideo} proposed to iteratively absorb sparsely extracted frame features, impression features being propagated all the way down the video which helped enhance features of low-quality frames. In the same way, the {\em light flow} of \citep{zhu2018HighPerformanceVideo} is a very small network designed to aggregate features on key frames. For non-key frames, sparse feature propagation was performed, reaching a speed of 25.6 fps. Fast YOLO \citep{shafiee2017FastYOLOFast} came up with an optimized architecture that has 2.8X fewer parameters with just a 2\% IOU drop, by applying a motion-adaptive inference method. Finally, \citep{chen2018optimizing} proposed to reallocate computational resources over a scale-time space: while expensive detection is done sparsely and propagated across both scales and time. Cheaper networks did the temporal propagation over a scale-time lattice.

An interesting question is "What can we expect from using temporal information?" The improvement of the mAP due to the direct use of temporal information can vary from +2.9\% \citep{zhu2017flow} to +5.6\% \citep{feichtenhofer2017detect}. Table~\ref{table:sumvid} does a recap of this sub-subsection.

\subsubsection{Object Detection in 3D Point Clouds}

This section addresses the literature about object detection in 3D data, whether it is true 3D point clouds or 2D images augmented with depth data (RGBD images). These problems raise novel challenges, especially in the case of 3D point clouds for which the nature of the data is totally different (both in terms of structure and contained information).  We can distinguish 4 main types of approaches depending on i) the use of 2D images and geometry, ii) the detections made in raw 3D point clouds, iii) the detections made in a 3D voxel grid iv) the detections made in 2D after projecting the point cloud on a 2D plane. Most of the presented methods are evaluated on the KITTI benchmark \citep{geiger2012we}. Section \ref{sec:3Ddatasets} introduces the datasets used for 3D object detection and quantitatively compares best methods on these datasets.

The methods belonging to the first category, monocular, start by the processing of RGB images and then add shape and geometric prior or occlusion patterns to infer 3D bounding boxes, as proposed by \citeauthor{chen2016Monocular3dObject} \cite{chen2016Monocular3dObject}, \citeauthor{mousavian20163DBoundingBox} \cite{mousavian20163DBoundingBox} and \citeauthor{xiang2015Datadriven3DVoxel} \cite{xiang2015Datadriven3DVoxel}. \citeauthor{deng2017AmodalDetection3D} \cite{deng2017AmodalDetection3D} revisited the amodal 3D detection by directly relating 2.5D visual appearance to 3D objects and proposed a 3D object detection system that simultaneously predicted 3D locations and orientations of objects in indoor scenes. \citeauthor{li2016VehicleDetection3D} \cite{li2016VehicleDetection3D} represented the data in a 2D point map and used a single 2D end-to-end fully convolutional network to detect objects and predicted full 3D bounding boxes even while using a 2D convolutional network. Deep MANTA \citep{chabot2017DeepMANTACoarsetofine} is a robust convolutional network introduced for simultaneous vehicle detection, part localization, visibility characterization and 3D dimension estimation, from 2D images.

\begin{table*}
\resizebox{\textwidth}{!}{
\begin{tabular}{|c|c|c|}\hline
Article references& Implementation & Operates on \\\hline\hline
\citep{chen2016Monocular3dObject, mousavian20163DBoundingBox, xiang2015Datadriven3DVoxel,li2016VehicleDetection3D, deng2017AmodalDetection3D,chabot2017DeepMANTACoarsetofine} & 3D priors & 2D images \\\hline
\citep{qi2017PointNetDeepLearning, qi2017pointnetplusplus,  qi2017FrustumPointNets3Da, srivastava2017LargeScaleNovel} & PointNetworks / Graph Convolutions / SuperPixels & PointClouds \\\hline
\citep{zhou2017VoxelNetEndtoEndLearning, li20173DFullyConvolutional, engelcke2017Vote3DeepFastObject} & 3/4D convolutions & Voxels \\\hline
\citep{chen2016MultiView3DObject, minemura2018LMNetRealtimeMulticlass, beltran2018BirdNet3DObject, simon2018ComplexYOLORealtime3D} & Plane choices / Discretization / Counting & Projections (Bird's eye) \\\hline
\citep{ku2017Joint3DProposal} & Feature Fusion & Multi-modal \\\hline

\end{tabular}
}
\caption{Summary of the 3D object detection approaches. \label{table:sum3D} }
\end{table*}

Among the methods using 3D point clouds directly, we can mention the series of papers relying on PointNet \citep{qi2017PointNetDeepLearning} and PointNet++ \citep{qi2017pointnetplusplus} networks, which are capable of dealing with the irregular format of point clouds without having to transform them into 3D voxel grids. F-PointNet \citep{qi2017FrustumPointNets3Da} is a 3D detector operating on raw point clouds (RGB-D scans). It leveraged mature 2D object detector to propose 2D object regions in RGB images and then collected all points within the frustum to form a frustum point cloud. 

Voxel based methods such as VoxelNet \citep{zhou2017VoxelNetEndtoEndLearning} represented the irregular format of point clouds by fixed size 3D Voxel grids on which standard 3D convolution can be applied. \citeauthor{li20173DFullyConvolutional} \cite{li20173DFullyConvolutional} discretized the point cloud on square grids, and represented discretized data by a 4D array of fixed dimensions. Vote3Deep \citep{engelcke2017Vote3DeepFastObject} examined the trade-off between accuracy and speed for different architectures applied on a voxelized representation of input data. 
 
Regarding approaches based on bird’s eye view, MV3D \citep{chen2016MultiView3DObject} projected LiDAR point cloud to a bird’s eye view on which a 2D region proposal network is applied, allowing the generation of 3D bounding box proposals. In a similar way, LMNet~\citep{minemura2018LMNetRealtimeMulticlass} addressed the question of real-time object detection using 3D LiDAR by projecting the point cloud onto 5 different frontal planes. More recently, BirdNet \citep{beltran2018BirdNet3DObject} proposed an original cell encoding mechanisms for bird’s eye view, which is invariant to distance and differences on LiDAR devices resolution, as well as a detector taking this representation as input. One of the fastest methods (50 fps) is ComplexYOLO \citep{simon2018ComplexYOLORealtime3D}, which expanded YOLOv2 by a specific complex regression strategy to estimate multi-class 3D boxes in Cartesian space, after building a bird’s eye view of the data. 

Some recent methods, such as \citep{ku2017Joint3DProposal}, combined different sources of information (eg., bird’s eye view, RGB images, 3D voxels, etc.) and proposed an architecture performing multimodal feature fusion on high resolution feature maps. \citeauthor{ku2017Joint3DProposal} \cite{ku2017Joint3DProposal} is one of the top performing methods on KITTI benchmark \citep{geiger2012we}. Finally, it is worth mentioning the super-pixel based method by \citeauthor{srivastava2017LargeScaleNovel} \cite{srivastava2017LargeScaleNovel} allowed to discover novel objects in 3D point clouds. Using the same fashion as in the previous section we display a recap in Table~\ref{table:sum3D}.

\subsection{Detecting Objects Under Constraints}
In object detection, challenges arise not only because of the naturally expected problems (scale, rotation, localization, occlusions, etc.) but also due to the ones that are created artificially. The first motivation for the following discussion is to know and understand the research works that deal with the inadequacy of annotations in certain datasets. This inadequacy could be due to weak (image-level) labels, scarce bounding box annotations or no annotations at all for certain classes. The second motivation is to discuss the approaches dealing with hardware and application constraints, real-world detectors might encounter.

\subsubsection{Weakly Supervised Detection}
\label{subsec:weakly_supervised}
Research teams want to include as many images as possible in their proposed datasets. Due to budget constraints or to save costs or for some other reasons, sometimes, they chose not to annotate precise bounding boxes around objects and include only image level annotations or captions. The object detection community has proven that it is still possible with enough weakly annotated data to train good object detectors.

The most obvious way to address Weakly Supervised Object Detection (WSOD) is to use the Multiple Instance Learning (MIL) framework~\citep{NIPS1997_1346}. The image is considered as being a bag of regions extracted by conventional object proposals: at least one of these candidate regions is positive if the image has the appropriate weak label, if not, no region is positive. The classical formulation of the problem at hand (before CNNs) then becomes a latent-SVM on the region's features where the latent part is the assignment of each proposal (that is weakly constrained by the image label). This problem being highly non-convex is heavily dependent on the quality of the initialization.

\citet{song2014learning,song2014weakly} thus focused on the initialization of the boxes by starting from selective-search proposals. They used for each proposal, its K-nearest neighbors in other images to construct a bipartite graph. The boxes were then pruned by taking only the patches that occur in most positive images (covering) while not belonging to the set of neighbors of regions found in negative images. They also applied Nesterov smoothing on the SVM objective to make the optimization easier.
Of course, if proposals do not spin enough of the image some objects will not be detected and thus the performance will be bad as there is no re-localization. The work of \citeauthor{sun2016ProNetLearningPropose} \cite{sun2016ProNetLearningPropose} also belongs to this category. \citet{Bilen_2015_CVPR} added regularization to the smoothened optimization problem of \citeauthor{song2014learning} \cite{song2014learning} using prior knowledge, but followed the same general directions. 
In another related research direction \citeauthor{wang2014WeaklySupervisedObject} \cite{wang2014WeaklySupervisedObject} learned to cluster the regions extracted with selective search into K-categories using unsupervised learning (pLSA) and then learned category selection using bag of words to determine the most discriminative clusters per class.

However, it is not always a requirement to explicitly solve the latent-SVM problem. Thanks to the fully convolutional structure of most CNNs it is sometimes possible to get a rough idea where an object might be while training for classification. For example, the arg-max of the produced spatial heat maps before global max-pooling is often located inside a bounding box as shown in \citep{oquab2015isobject,oquab2014WeaklySupervisedObject}. It is also possible to learn to detect objects without using any ground truth bounding boxes for training by masking regions of the image and see how the global classification score is impacted, as proposed by \citeauthor{bazzani2016SelftaughtObjectLocalization} \cite{bazzani2016SelftaughtObjectLocalization}.  

\begin{table*}
\resizebox{\textwidth}{!}{
\begin{tabular}{|c|c|c|}\hline
Article references& Implementation & Paradigm \\\hline\hline
\citep{song2014learning,song2014weakly,wang2014WeaklySupervisedObject,sun2016ProNetLearningPropose, Bilen_2015_CVPR} & Optimization Tricks (smoothing, EM, etc.) & Full MIL \\\hline
\citep{bazzani2016SelftaughtObjectLocalization} & Monitor score change & Masking \\\hline
\citep{oquab2015isobject,oquab2014WeaklySupervisedObject,zhou2015LearningDeepFeatures,pinheiro2015ImagelevelPixellevelLabeling,bilen2016WeaklySupervisedDeep} & Global Pooling / GAPooling / LogSumExp Pooling & Refining Pooling \\\hline
\citep{Durand_MANTRA_ICCV_2015, Durand_WELDON_CVPR_2016, Durand_WILDCAT_CVPR_2017} & top-k max/min & Contradictory Evidence Pooling \\\hline
\citep{kumar2016track,chen2017DiscoverLearnNew,yuan2017temporal,papadopoulos2016WeDonNeed,tang2016LargeScaleSemisupervised,huval2013DeepLearningClassgeneric,hoffman2015detector,rochan2015WeaklySupervisedLocalization} & Subtitles / Motion Cues / User clicks / Strong Annotations & Auxiliary Supervision \\\hline
\end{tabular}
}   
\caption{Summary of the weakly supervised approaches.\label{table:sumweak}}
\end{table*}

This free localization information can be improved through the use of different pooling strategies. For instance: producing a spatial heat map and using a global average pooling instead of global max pooling to train in classification. This strategy was used in \citep{zhou2015LearningDeepFeatures} where the heat maps per class were thresholded to obtain bounding boxes. In this line of work, \citeauthor{pinheiro2015ImagelevelPixellevelLabeling} \cite{pinheiro2015ImagelevelPixellevelLabeling} went a step further by producing pixel-level label segmentation maps using Log-Sum-Exp pooling in conjunction with some image and smoothing prior. Other pooling strategies involved aggregating minimum and maximum evidences to get a more precise idea where the object is and isn't, \eg, as in the line developed in \citet{Durand_MANTRA_ICCV_2015, Durand_WELDON_CVPR_2016, Durand_WILDCAT_CVPR_2017}. 
\citeauthor{bilen2016WeaklySupervisedDeep} \cite{bilen2016WeaklySupervisedDeep} used the spatial pyramid pooling module to take MIL to the modern-age by incorporating it into a Fast R-CNN like architecture with a two-stream Fast R-CNN proposal classification part: one with classification score and the other with relative rankings of proposals that are merged together using hadamard products. Thus, producing region level labels predictions like in classic detection settings. They then aggregated all labels per image by taking the sum. They trained it end-to-end using image level labels thanks to their aggregation module while adding a spatial-regularization constraint on the features obtained by the SPP module.

Another idea, which can be combined with MIL is to draw the supervision from elsewhere. Tracked object proposals were used by \citeauthor{kumar2016track} \cite{kumar2016track} to extract pseudo-groundtruth to train detectors. This idea was further explored by \citeauthor{chen2017DiscoverLearnNew} \cite{chen2017DiscoverLearnNew} where the keywords extracted from the subtitles of documentaries allowed to further ground and cluster the generated annotations. In a similar way, \citeauthor{yuan2017temporal} \cite{yuan2017temporal} used action description supervision via LSTMs. Cheap supervision can also be gained by involving user feedback \citep{papadopoulos2016WeDonNeed}, where the users iteratively improved the pseudo-ground truth by saying if the objects were missed or partly included in the detections. Click supervision by users, far less demanding than full annotations, also improved the performance of detectors \citep{papadopoulos2017training}. \citep{roy2016ActiveLearningVersion} used active learning to select the right images to annotate and thus get the same performance by using far fewer images. One can also leverage strong annotations for other classes to improve the performance of weakly supervised classes. This was done in \citep{tang2016LargeScaleSemisupervised} by using the powerful LSDA framework~\citep{hoffman2014lsda}. This was also the case in \citep{huval2013DeepLearningClassgeneric,hoffman2015detector,rochan2015WeaklySupervisedLocalization}.

This year, a lot of interesting new works continued to develop the MIL+CNN framework using diverse approaches \citep{Ge_2018_CVPR,Zhang_2018_CVPR,Wan_2018_CVPR,Zhang_2018_CVPR2,Zhang_2018_CVPR1,tang2017multiple}. These articles will not be treated in detail because the focus of this survey is object detection in general and not WSOD.

As of this writing, the state-of-the-art mAP on VOC2007 in WSOD is 47.6\% \citep{Zhang_2018_CVPR1}. The gap is being reduced at an exhilarating pace but we are still far from the 83.1\% state-of-the-art with full annotations~\citep{mordan2017DeformablePartbasedFully} (without COCO pre-training). We present a recap in Table~\ref{table:sumweak}.

\subsubsection{Few-shot Detection}
The cost of annotating thousands of boxes over hundreds of classes is too high. Although some large scale datasets are created, but it is not practical to do it for every single target domain. Collecting and annotating training examples in the case of video is even costlier than still images, making few shot detection more interesting. For this purpose, researchers have come up with ways to train the detectors with as low as three to five bounding boxes per target class and get lower but competitive performance as compared to the fully supervised approach on a large scale dataset. Few shot learning usually relies on semi-supervised learning mechanisms.

\citeauthor{dong2017FewshotObjectDetection} \cite{dong2017FewshotObjectDetection} took up an iterative approach to simultaneously train the model and generate new samples which are used in the following iterations for training. They observed that as the model becomes more discriminative it is able to sample harder as well as more number of instances. Iterating between multiple kinds of detectors was found to outperform the single detector approach. One interesting aspect of the paper is that their approach with only three to four annotations per class gives results comparable to weakly annotated approaches with image level annotations on the whole PASCAL VOC dataset. A similar approach was used by \citeauthor{keren2018weakly} \cite{keren2018weakly}, who proposed a model which can be trained with as few as one single exemplar of an unseen class and a larger target example that may or may not contain an instance of the same class as the exemplar (weakly supervised learning). This model was able to simultaneously identify and localize instances of classes unseen at training time. 

Another way to deal with few-shot detection is to fine-tune a detector trained on a sourced domain to a target domain for which only few samples are available. This is what \citeauthor{chen2018lstd} \cite{chen2018lstd} did, by introducing a novel regularization method, involving, depressing the background and transferring the knowledge from the source domain to the target domain to enhance the fine-tuned detector. 

For videos, \citeauthor{misra2015WatchLearnSemisuperviseda} \cite{misra2015WatchLearnSemisuperviseda} proposed a semi-supervised framework in which some initial labeled boxes allowed to iteratively learn and label hundreds of thousands of object instances automatically. Criteria for reliable object detection and tracking constrained the semi-supervised learning process and minimized semantic drift. 

\subsubsection{Zero-shot Detection} 
Zero-shot detection is useful for a system where large number of classes are to be detected. Its hard to annotate a large number of classes as the cost of annotation gets higher with more classes. This is a unique type of problem in the object detection domain as the aim is to classify and localize new categories, without any training examples, during test time with the constraint that the new categories are semantically related to the objects in the training classes. Therefore, in practice the semantic attributes are available for the unseen classes. The challenges that come with this problem are: \textit{First,} zero-shot learning techniques are restricted to recognize a single dominant objects and not all the object instances present in the image. \textit{Second,} the \textit{background} class during fully supervised training may contain objects from unseen classes. The detector will be trained to discriminatively treat these classes as background.

While there is a comparably large literature present for zero shot classification, well covered in the survey \citep{fu2018recent}, zero shot detection has only a few papers to the best of our knowledge. \citeauthor{zhu2018ZeroShotDetection} \cite{zhu2018ZeroShotDetection} proposed a method where semantic features are utilized during training but it is agnostic to semantic information during test time. This means they incorporated semantic attribute information in addition to seen classes during training and generated proposals only, but no identification label. for seen and unseen objects at test time. \citeauthor{rahman2018zero} \cite{rahman2018zero} proposed a multitask loss that combines max-margin, useful for separating individual classes, and semantic clustering, useful for reducing noise in semantic vectors by positioning similar classes together and dissimilar classes far apart. They used ILSVRC \citep{deng2009imagenet} which contains an average of only three objects per image. They also proposed another method for a more general case when unseen classes are not predefined during training. \citeauthor{bansal2018zero} \cite{bansal2018zero} proposed two background-aware approaches, statically assigning the background image regions into a single background class embedding and latent assignment based alternating algorithms which associated background to different classes belonging to a large open vocabulary, for this task. They used MSCOCO \citep{lin2014microsoft} and VisualGenome \citep{krishna2017visual} which contain an average of 7.7 and 35 objects per image respectively. They also set number of unseen classes to be higher, making their task more complex than previous two papers. 

Since, it is quite a new problem there is no well-defined experimental protocol for this approach. They vary in number and nature of unseen classes, use of semantic attribute information of unseen classes during training, complexity of the visual scene, \etc

\subsubsection{Fast and Low Power Detection}
\label{subsec:fastandlowpower}
There is generally a trade-off between performance and speed (we refer to the comprehensive study of \citep{huang2017speed} for instance). When one needs real time detectors, like for video object detection, one loses some precision. However, researchers have been constantly working on improving the precision of fast methods and making precise methods faster. Furthermore, not every setup can have powerful GPUs, so for most industrial applications the detectors have to run on CPUs or on different low power embedded devices like Raspberry-Pie. 

Most real-time methods are single stage because they need to perform inference in a quasi fully constitutional manner. The most iconic methods have already been discussed in detail in the rest of the paper \citep{redmon2016you,liu2016ssd, redmon2016YOLO9000BetterFaster,redmon2018YOLOv3IncrementalImprovement,lin2017focal}. \citeauthor{zhou2018ScaleTransferrableObjectDetection} \cite{zhou2018ScaleTransferrableObjectDetection} designed a scale transfer module to replace the feature pyramid and thus got a detection network more accurate and faster than YOLOv2. \citeauthor{iandola2014DenseNetImplementingEfficient} \cite{iandola2014DenseNetImplementingEfficient} provided a framework to efficiently compute multi-scale features. \citeauthor{redmon2015RealtimeGraspDetection} \cite{redmon2015RealtimeGraspDetection} used a YOLO-like architecture to provide oriented bounding boxes symbolizing grasps in real time. \citeauthor{shafiee2017FastYOLOFast} \cite{shafiee2017FastYOLOFast} built a faster version of YOLOv2 that runs on embedded devices other than GPUs. \citeauthor{li2017FSSDFeatureFusion} \cite{li2017FSSDFeatureFusion} managed to speed-up the SSD detector, bringing it to almost 70 fps, using a more lightweight architecture.  

In single stage methods most of the computations are found in the backbone networks so researchers started to design new backbones for detection in order to have fewer operations like PVANet \citep{kim2016PVANETDeepLightweight} that built a deep and thin networks with fewer channels than its classification counterparts, or SqueezeDet \citep{wu2016SqueezeDetUnifiedSmall} that is similar to YOLO but with more anchors and fewer parameters.

\citeauthor{iandola2016squeezenet} \cite{iandola2016squeezenet} built an AlexNet backbone with 50 times fewer parameters. \citeauthor{howard2017mobilenets} \cite{howard2017mobilenets} used depth-wise-separable convolutions and point-wise convolutions to build an efficient backbone called MobileNets for image classification and detection. \citeauthor{sandler2018mobilenetv2} \cite{sandler2018mobilenetv2} improved upon it by adding residual connections and removing non-linearities. Very recently, \citeauthor{tan2018MnasNetPlatformAwareNeural} \cite{tan2018MnasNetPlatformAwareNeural} used architecture search to come up with an even more efficient network (1.5 times faster than \citeauthor{sandler2018mobilenetv2} \cite{sandler2018mobilenetv2} and with lower latency). ShuffleNet \citep{DBLP:journals/corr/ZhangZLS17} attained impressive performance on ARM devices.  They can sustain only that many computations (40MFlops). Their backbone is 13 times faster than AlexNet.

Finally, \citeauthor{wang2018PeleeRealTimeObject} \cite{wang2018PeleeRealTimeObject} proposed PeleeNet, a light network that is 66\% of the model size of MobileNet, achieving 76.4\% mAP on PASCAL VOC2007 and 22.4\% mAP on MS COCO at a speed of 17.1 fps on iPhone 6s and 23.6 fps on iPhone 8. \citep{li2018TinyDSODLightweightObject} is also very efficient, achieving 72.1\% mAP on PASCAL VOC2007 with 0.95M parameters and 1.06B FLOPs.

Fast double-staged methods exist, although the NMS part becomes generally the bottleneck. Among them one can also mention for the second time \citeauthor{singh2017RFCN300030fpsDecoupling} \cite{singh2017RFCN300030fpsDecoupling}, which is one of the double-staged methods that researchers have brought to 30 fps by using superclass (sets of similar classes) specific detection. Using a mask obtained by a fast and coarse face detection method the authors of \citep{chen2016SupervisedTransformerNetwork} reduced the computational complexity of their double stage detector by a great amount at test time by only computing convolutions on non-masked regions. \citeauthor{singh2017RFCN300030fpsDecoupling} \cite{singh2017RFCN300030fpsDecoupling} sped up R-FCN by using detection heads super classes (sets of similar classes) specific and thus decouple detection from classification. SNIPER~\citep{singh2018sniper} can train on 512x512 images using an adaptive sampling of the region of interests. Therefore, it's training can use larger batch size and therefore, be way faster but it needed 30\% more pixels than original images at inference time making it slower.

There have also been lots of work done on pruning and/or quantifying the weights of CNNs for image classification \citep{han2015deep, lin2015neural, rastegari2016xnor,hubara2016binarized,zhou2016dorefa,hubara2017quantized,huang2017condensenet, huang2018learning,2018arXiv180710029Z,2018arXiv180711254P}, but much fewer in detection yet. Although, one can find some detection articles that used pruning. \citeauthor{girshick2015fast} \cite{girshick2015fast} used SVD on the weights of the fully connected layers in Fast R-CNN. \citeauthor{masana2016OntheflyNetworkPruning} \cite{masana2016OntheflyNetworkPruning}, who pruned near-zero weights in detection networks and extended the compression to be domain-adaptive in \citeauthor{masana2017domain} \cite{masana2017domain}.

\begin{figure*}
\centering
\resizebox{.99\linewidth}{!}{\includegraphics{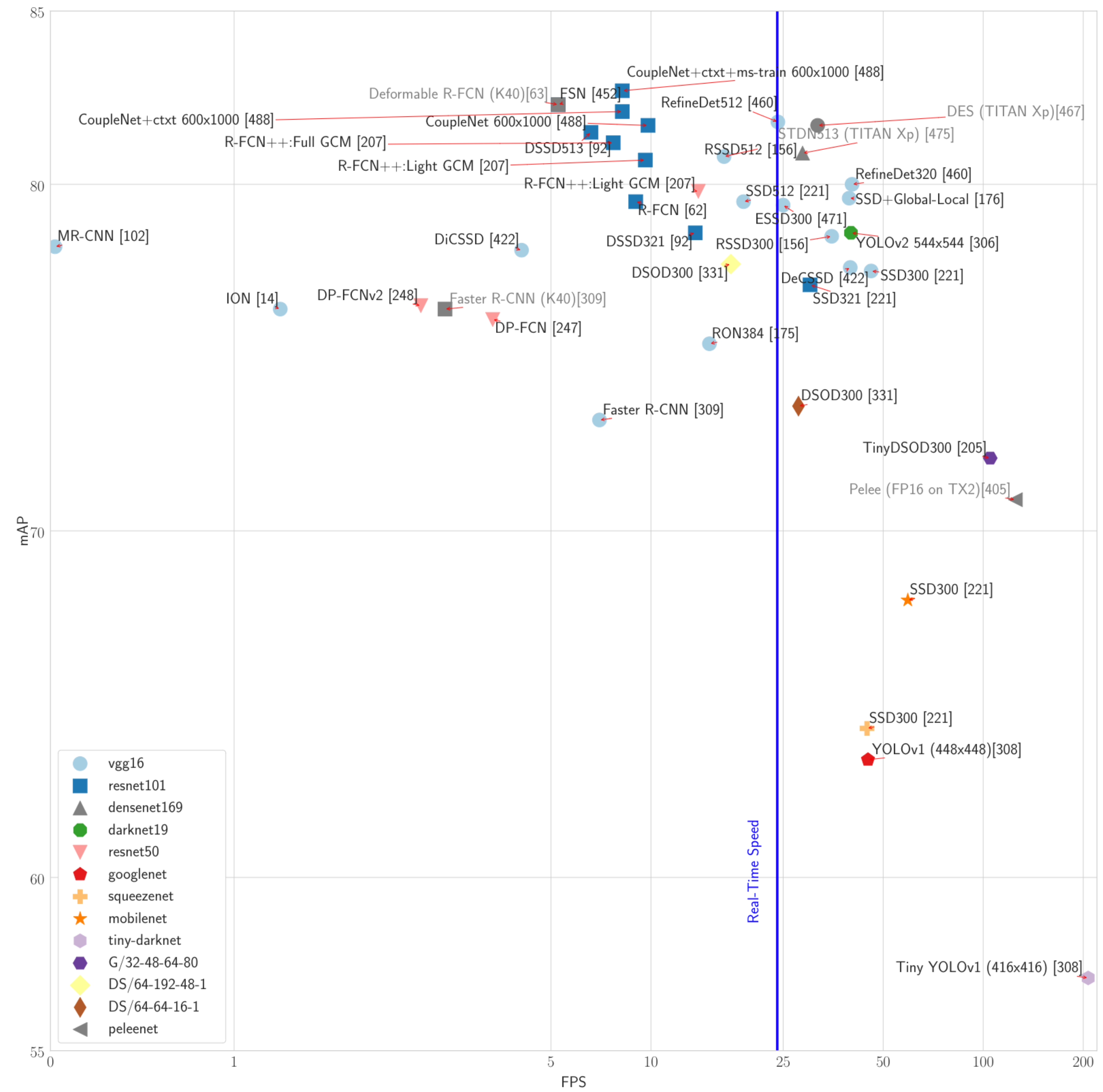}}
\caption{Performance on VOC07 with respect to Inference speed \textbf{on a TitanX} GPU. The vertical line represents the limit of Real-Time Speed (indistinguishable from continuous motion for the human eye). We also added in light gray some relevant work measured on similar devices (K40, TITAN Xp, Jetson TX2). Only RefineDet \cite{zhang2018single}, DES \cite{Zhang2018CVPR0} and STDN \cite{zhou2018ScaleTransferrableObjectDetection} are simultaneously real-time and above 80\% in mAP although for some of them (DES, STDN) better hardware (TITAN Xp) must have helped. \label{fig:accspeed}}
\end{figure*}

To help the reader better encompass the different accuracy vs speed trade-offs present in the modern methods, we display some of the leading methods on PASCAL-VOC 2007~\cite{everingham2010pascal} with their inference speed on one image (batch size of 1) in Figure~\ref{fig:accspeed}. 

It is not only necessary to respect available material constraints (data and machines) but detectors have to be reliable too. They must be robust to perturbations and they can make mistakes but the mistakes also need to be interpretable, which is a challenge in itself with the millions of weights and the architectural complexity of modern pipelines. It is a good sign to outperform all other methods on a benchmark, it is something else to perform accurately in the wild. That is why we dedicate the following sections to the exploration of such challenges.

\subsection{Towards Versatile Object Detectors}
\begin{figure*}
    \centering
    \includegraphics[width=\textwidth]{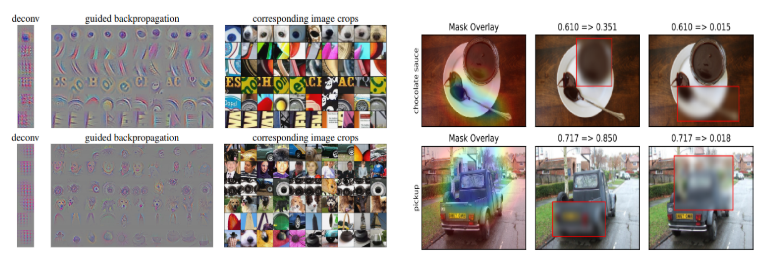}
    \caption{On the left side we display an example of guided backpropagation to visualize the pattern that make the neurons fire from~\cite{springenberg2014striving} and on the right side we show the approach of gradients mask to find important zones for a classifier on an image from~\cite{fong2017interpretable}, which can lead to bad surprises (the network uses the spoon as a proxy for the presence of coffee).
    \label{fig:interpretability}}
\end{figure*}
So far in all this survey, detectors were tested on limited, well-defined benchmarks. It is mandatory to assess their performances. However, at the end we are really interested in their behaviors {\em in the wild} where no annotations are present. Detectors have to be robust to unusual situations and one would wish for detectors to be able to evolve themselves. This section will review the state of deep learning methods \wrt these expectations.
\subsubsection{Interpretability and Robustness}
With the recent craze about self-driving cars, it has become a top priority to build detectors, that can be trusted with our lives. Hence, detectors should be robust to physical adversarial attacks \citep{lu2017stopsigns,shang2018adversarial} and weather conditions, which was the reason for building KITTI \citep{geiger2012we} and DETRAC \citep{detrac2015} back then and has now led to the creation of two amazingly huge datasets: ApolloScape from Baidu \citep{DBLP:journals/corr/abs-1805-04687} and BDD100K from Berkeley \citep{DBLP:journals/corr/abs-1805-04687} car detection datasets. The driving conditions of the real world are so complex: changing environments, reflections, different traffic signs and rules for different countries. So far, this open problem is largely unsolved even if some industry players seem to be confident enough to leave self-driving cars without safety nets in specific cities. 
It will surely involve at some point the heavy use of synthetic data otherwise it would take a lifetime to gather the data necessary to be confident enough.
To finish on a positive note detectors in self-driving cars can benefit from multi-sensory inputs such as LiDAR point clouds \citep{himmelsbach2008lidar}, other lasers and multiple cameras so it can help disambiguate certain difficult situations (reflections on the cars in front of it for instance).

But most of all the detectors should incorporate a certain level of interpretability so that if a dramatic failure happens it can be understood and fixed. It is also a need for legal matters. Very few works have done so because it requires delving into the feature maps of the backbone network. A few works proposed different approaches for classification only but no consensus has been reached yet. Among the popular methods one can cite the gradient map in the image space of \citeauthor{simonyan2013visualizing} \cite{simonyan2013visualizing}, the occlusion analysis of \citeauthor{zeiler2014visualizing} \cite{zeiler2014visualizing}, the guided back propagation of \citeauthor{springenberg2014striving} \cite{springenberg2014striving} and, recently, the perturbation approach of \citeauthor{fong2017interpretable} \cite{fong2017interpretable}. Figure~\ref{fig:interpretability} shows the insights gained by using two of the mentioned methods on a classifier.

No method exists yet for object detectors to the best of our knowledge. It would be a very interesting research direction for future works.

\subsubsection{Universal Detector, Lifelong Learning}
Having object detectors able to iteratively, and without any supervision, learn to detect novel object classes and improve their performance would be one of the Holy Grails of computer vision. This can have the form of lifelong learning, where goal is to sequentially retrain learned knowledge and to selectively transfer the knowledge when learning a new task, as defined in \citep{silver2013LifelongMachineLearning}. Or never ending learning \citep{mitchell2018NeverEndingLearning}, where the system has sufficient self-reflection to avoid plateaus in performances and can decide how to progress by itself. However, one of the biggest issues with current detectors is they suffer from {\em catastrophic forgetting}, as say \citeauthor{castro2018EndtoEndIncrementalLearning} \cite{castro2018EndtoEndIncrementalLearning}. It means their performance decreases when new classes are added incrementally. Some authors tried to face this challenge. For example, the {\em knowledge distillation loss} introduced by \citeauthor{li2018LearningForgetting} \cite{li2018LearningForgetting} allows to forget old data while using previous models to constraint updated ones during learning. In the domain of object detection, the only recent contribution we are aware of is the incremental learning approach of \citeauthor{shmelkov2017IncrementalLearningObject} \cite{shmelkov2017IncrementalLearningObject}, relying on a distillation mechanism. Lifelong learning and never ending learning are domains where a lot still have to be discovered or developed.

\subsection{Concluding Remarks}
It seems that deep learning in its current form is not yet fully ready to be applied to other modalities than 2D images: in videos, temporal consistency is hard to take into account with DCNNs because 3D convolutions are expensive, tubelets and tracklets are interesting ideas but lack the elegance of DCNNs on still images. For point clouds the picture is even worse. The voxelisation of point clouds does not deal with their inherent sparsity and create memory issues and even the simplicity and originality of the PointNet articles \citet{qi2017PointNetDeepLearning,qi2017pointnetplusplus} that leaves the point clouds untouched has not matured enough yet to be widely adopted by the community.
Hopefully, dealing with other constraints like weak supervision or few training images is starting to produce worthy results without too much change to the original DCNN architectures \citep{Durand_WILDCAT_CVPR_2017,Ge_2018_CVPR,Zhang_2018_CVPR,Wan_2018_CVPR,Zhang_2018_CVPR2,Zhang_2018_CVPR1,tang2017multiple}. It seems to be only a matter of refining cost functions and coming-up with more building blocks than reinventing DCNNs entirely.
However, the Achilles heel of deep-learning methods is their interpretability and trustworthiness. The object detection community seems focused on improving the performances on static benchmarks instead of finding ways to better understand the behavior of DCNNs. It is understandable but it shows that Deep Learning has not yet reached full maturity. Eventually, one can hope that the performances of new detectors will plateau and when it does, researchers will be forced to come back to the basics and focus instead on interpretability and robustness before the next paradigm washes off deep-learning entirely.

%% file: article_conclusions.tex
\section{Conclusions \label{sec:conclusion}}

Object detection in images, a key topic attracting a substantial part of  the computer vision community, has been revolutionized by the recent arrival of convolutional neural networks, which swept all the methods previously dominating the field. This article provides a comprehensive survey of what happened in the domain since 2012. It shows that, even if top-performing methods concentrate around two main alternatives -- single stage methods such as SSD or YOLO, or two stages methods in the footsteps of Faster RCNN -- the domain is still very active. Graph networks, GANs, context, small objects, domain adaptation, occlusions, \etc are the directions that are actively studied in the context of object detection. Extension of object detection to other modalities, such as videos or 3D point clouds, as well as constraints, such as weak supervision is also very active and has been addressed. The appendix of this survey also provides a very complete list of the public datasets available to the community and highlights top performing methods on these datasets. We believe this article will be useful to better understand the recent progress and the bigger picture of this constantly moving field.

%% file: article_datasets.tex
\section{Datasets and Results}
\label{sec:datasets}

Most of the object detection's influential ideas, concepts and literature having been now reviewed, the rest of the article  dives into the datasets used to train and evaluate these detectors. 

Public datasets play an essential role as they not only allow to measure and compare the performance of object detectors but also provides resources allowing to learn object models from examples. In the area of deep learning, these resources play an essential role, as it has been clearly shown that deep convolutional neural networks are designed to benefit and learn from massive amount of data \citep{zhou2014LearningDeepFeatures}. This section discusses the main datasets used in the recent literature on object detection and present state-of-the-art methods for each dataset.

\subsection{Classical Datasets with Common Objects}
\label{subsec:datasets_commonobjects}
We first start by presenting the datasets containing everyday life object taken from consumer cameras. This category contains the most important datasets for the domain, attracting the largest part of the community. We will discuss in a second section the datasets devoted to specific detection tasks (\eg, face detection, pedestrian detection, \etc).

\subsubsection{Pascal-VOC}
Pascal-VOC~\citep{everingham2010pascal} is the most iconic object detection dataset. It has changed over the years but the format everyone is familiar with is the one that emerged in 2007 with 20 classes (Person: person; Animal: bird, cat, cow, dog, horse, sheep; Vehicle: aeroplane, bicycle, boat, bus, car, motorbike, train; Indoor: bottle, chair, dining table, potted plant, sofa, tv/monitor). It is now used as a test bed for most new algorithms. As it is quite small there have been claims that we are starting to overfit on the test set and therefore, MS-COCO (see next section) is preferred nowadays to demonstrate the quality of a new algorithm. The 0.5 IoU based metrics this dataset introduced has now become the de facto standard for every single detection problem. Overall, this dataset's impact on the development of innovative methods in object detection cannot be overstated. 
It is quite hard to find all relevant literature but we have tried to be as thorough as possible in terms of best performing methods. The URL of the dataset is \url{http://host.robots.ox.ac.uk/pascal/VOC/}.

Two versions of Pascal-VOC are commonly used in the literature, namely VOC2007 and VOC2012:

\begin{itemize}
	\item VOC07, with 9,963 images containing 24,640 annotated objects, is small. For this reason, papers using VOC07 often train on the union of VOC07 and VOC12 trainvals (VOC07+12). The Average Precision (AP) averaged across the 20 classes is saturating at around 80 points @0.5 IoU. Some methods got extra points but it seems one cannot go over around 85 points (without pre-training on MS COCO). Using MS COCO data in addition, one can get up to 86.3 AP (see \citep{li2018RFCNAccurateRegionBased}). We chose to display methods with mAP over 80 points only on Table \ref{table:voc_2007}. We do not distinguish between the methods that do multiple inference tricks or the methods that reports results as is. However for each method we reported for the highest published results we could get.
	\item VOC12 is a little bit harder than its 2007 counterpart, and we have just gone over the 80 point mark. As it is harder, this time, most literature uses the union of the whole VOC2007 data (trainval+test) and VOC2012 trainval; It is referred to as 07++12. Again better results are obtained with pre-training on COCO data (83.8 points in \citep{he2016deep}). Results above 75 points are presented in Table~\ref{table:voc_2012}.
\end{itemize}
On both splits all backbones used by the leaders of the board are heavy backbones with more than a 100 layers except for \citep{Zhang2018CVPR0} that gets close to state of the art using only VGG-16.

\begin{table}[tb]
\centering
\begin{tabular}{|c|c|c|}
\hline
Method & Backbone & mAP \\
\hline
\citep{mordan2017DeformablePartbasedFully} & ResNeXt-101 & \textbf{83.1}\\
\citep{DBLP:journals/corr/abs-1712-02408} & ResNet-101 & \textbf{83.1}\\
\citep{Zhai2018CVPR} & ResNet-101 & 82.9 \\
\citep{dai2017deformable} & ResNet-101& 82.6\\
\citep{2018arXiv180807993K} & ResNet-101 & 82.4 \\
\citep{li2018RFCNAccurateRegionBased} & ResNet-101 & 82.1\\
\citep{Zhang2018CVPR0} & VGG-16 & 81.7 \\
\citep{fu2017dssd} & ResNet-101& 81.5\\
\citep{zhou2018ScaleTransferrableObjectDetection} & DenseNet-169 & 80.9\\
\citep{Zhao_2018_CVPR} & ResNet-101 & 80.7 \\
\citep{dai2016r} & ResNet-101 & 80.5 \\

\hline
\end{tabular}
\caption{State-of-the-art methods on VOC07 test set (Using VOC07+12).\label{table:voc_2007}}
\end{table}

\begin{table}[tb]
\centering
\begin{tabular}{|c|c|c|}
\hline
Method & Backbone & mAP \\
\hline
\citep{DBLP:journals/corr/abs-1712-02408} & ResNet-101 & \textbf{81.2}\\
\citep{2018arXiv180807993K} & ResNet-101 & 81.1 \\
\citep{mordan2017DeformablePartbasedFully} & ResNeXt-101 & 80.9\\
\citep{li2018RFCNAccurateRegionBased} & ResNet-101 & 80.6\\
\citep{Zhai2018CVPR} & ResNet-101 & 80.5 \\
\citep{Zhang2018CVPR0} & VGG-16 & 80.3 \\
\citep{fu2017dssd} & ResNet-101& 80.0\\
\citep{liu2016ssd} & ResNet-101 & 78.5 \\
\citep{dai2016r} & ResNet-101 & 77.6 \\
\hline
\end{tabular}
\caption{State-of-the-art methods on VOC12 test set (Using VOC07++12).\label{table:voc_2012}}
\end{table}

\subsubsection{MS COCO}
MS COCO \citep{lin2014microsoft} is the most challenging object detection dataset available today. It consists of 118,000 training images, 5,000 validation images and 41,000 testing images. They have also released 120K unlabeled images that follow the same class distribution as the labeled images. They may be useful for semi-supervised learning on COCO. The MS COCO challenge has been ongoing since 2015. There are 80 object categories, over 4 times more than Pascal-VOC. MS COCO is a fine replacement for Pascal-VOC, that has arguably started to age a little. Like ImageNet in its time, MS-COCO has become the de facto standard for the object detection community and any method winning the state-of-the-art on it is assured to gain much traction and visibility. The AP is calculated similar to Pascal-VOC but averaged on multiple IoUs from 0.5 to 0.95. 

\begin{figure*}
    \centering
    \includegraphics[clip, trim={2px 0px 0px 0px}, width=\textwidth]{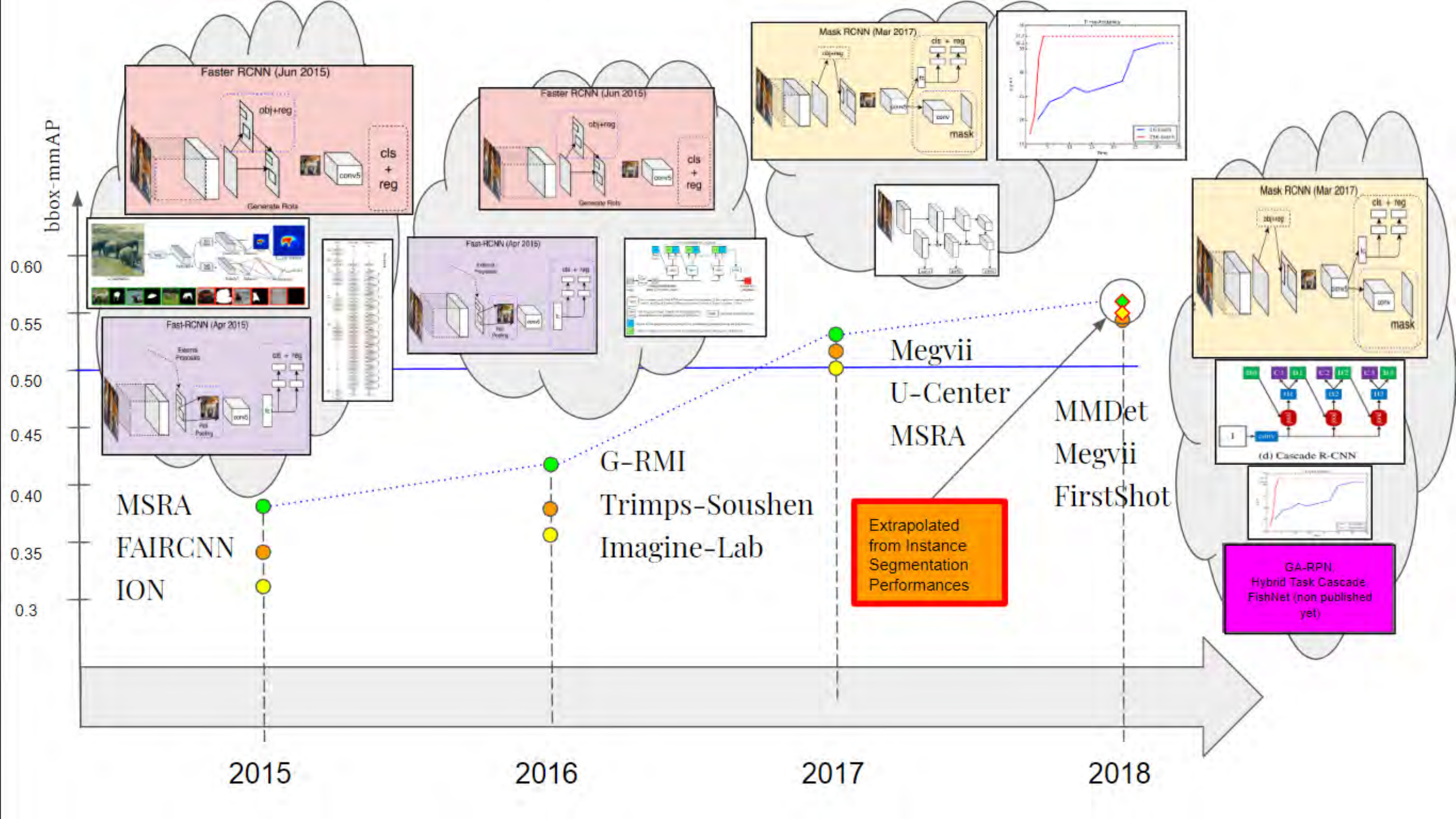}
    \caption{This plot displays the performance advances in the bounding boxes detection COCO challenge over the years. For each year we present the main ideas behind the three best performing entries in terms of \textbf{mmAP}. In 2015 the main frameworks were Fast R-CNN~\cite{girshick2015fast}, DeepMask~\cite{pinheiro2015LearningSegmentObject} and Faster R-CNN~\cite{ren2015faster} supported by the new Deep ResNets~\cite{he2016deep}. In 2016, the same pipelines won the competition with the addition of AttractioNet~\cite{gidaris2016AttendRefineRepeat} and LocNet~\cite{gidaris2016LocNetImprovingLocalization} for better proposals and localization accuracy. In 2017 Mask R-CNN~\cite{he2017mask}, FPN~\cite{lin2017feature} and MegDet~\cite{peng2017megdet} proved that more complex ideas could allow to go over the 50\% mark. In 2018 the same pipelines, as in 2017 (namely Mask R-CNN), were enriched with the multi-stages of Cascade R-CNN~\cite{cai2018cascade}, a new RPN and backbones that were for the first time specifically designed for the detection task. The last entry of 2018 reached 53\% mmAP and we can extrapolate the two first entries to be around 55\% bbox mmAP based on their ranking for instance segmentation.}
    \label{fig:coco_res}
\end{figure*}

Most available alternatives stemmed from Faster R-CNN \citep{ren2015faster}, which in its first iteration won the first challenge with 37.3 mAP with a ResNet101 backbone. In the second iteration of the challenge the mAP went up to 41.5 with an ensemble of Faster R-CNN \citep{ren2015faster} that used a different implementation of RoI-Pooling. This maybe inspired the RoI-Align of Mask R-CNN \citep{he2017mask}. Tao Kong claimed that a single Faster R-CNN with HyperNet features \citep{kong2016HyperNetAccurateRegion} can reach 42.0 mAP. The best published single model method \citep{peng2017megdet} nowadays is around 50.5 (52.5 with an ensemble) and relied on different techniques already mentioned in this survey. Among them one can mention FPN \citep{lin2017feature}, large batch training \citep{peng2017megdet} and GCN \citep{peng2017large}. Ensembling Mask R-CNNs \citep{he2017mask} gave around the same performance as \citep{peng2017megdet} at around 50.3 mAP. Deformable R-FCN \citep{dai2017deformable} is not lagging too far behind with 48.5 mAP single model performance (50.4 mAP with an ensemble) using Soft NMS \citep{bodla2017soft} and the "mandatory" FPN~\citep{lin2017feature}. Other entries were based mostly on Mask R-CNN \citep{he2017mask}. We display the current leaderboard (\url{http://cocodataset.org/#detection-leaderboard}) also visible at for all the past challenges with the main-ideas present in the winning entries Figure~\ref{fig:coco_res}. The URL of the dataset is \url{http://cocodataset.org}.

\subsubsection{ImageNet Detection Task}
ImageNet is a dataset organized according to the nouns of the WordNet hierarchy. Each node of the hierarchy is depicted by hundreds and thousands of images, with an average of over 5,000 images per node. Since 2010, the {\em Large Scale Visual Recognition Challenge} is organized each year and contains a detection challenge using ImageNet images. The detection task, in which each object instance has to be detected, has 200 categories. There is also a classification and localization task, with 1,000 categories in which algorithms have to produce 5 labels (and 5 bounding boxes) only, allowing not to penalize the detection of objects that are present, but not included in the ground truth. In the 2017 contest, the top detector was proposed by a team from Nanjing University of Information Science and Imperial College London. It ranked first on 85 categories with an overall AP of 73.13. As far as we know, there is no paper describing the approach precisely (but some slides are available at the workshop page). The 2nd ranked method was from \citeauthor{bae2017RankExpertsDetection} \cite{bae2017RankExpertsDetection}, who observed that modern convolutional detectors behave differently for each object class. The authors consequently built an ensemble detector by finding the best detector for each object class. They obtained a AP of 59.30 points and won 10 categories. ImageNet is available at \url{http://image-net.org}.

\subsubsection{VisualGenome}
VisualGenome~\citep{krishna2017visual} is a very peculiar dataset focusing on object relationships. It contains over 100,000 images. Each image has bounding boxes but also complete scene graphs. Over 17,000 categories of objects are present. The first ones in terms of representativeness by far are man and woman followed by trees and sky. On average there are 21 objects per image. It is unclear if it qualifies as an object detection dataset as the paper does not include clear object detection metrics or evaluation as its focus is on scene graphs and visual relationships. However, it is undoubtedly an enormous source of strongly supervised images to train object detectors. The Visual Genome Dataset has huge number of classes, most of them being small and hard to detect. The mAP reported in the literature is therefore, much smaller compared to previous datasets. One of the best performing approaches is of \citeauthor{li201717scenegraph} \cite{li201717scenegraph} which reached 7.43 mAP by linking object detection, scene graph generation and region captioning. Faster R-CNN \citep{girshick2015fast} has a mAP of 6.72 points on this dataset. The URL of the dataset is \url{https://visualgenome.org}.

\subsubsection{OpenImages}
The challenge OpenImagesV4~\citep{openimages} that will be organized for the first time at ECCV2018 offers the largest to date common objects detection dataset with up to 500 classes (including the familiar ones from Pascal-VOC) on 1,743,000 images and more than 12,000,000 bounding boxes with an average of 7 objects per image for training, and 125,436 images for tests (41,620 for validation). The object detection metric is the AP@0.5IoU averaged across classes taking into account the hierarchical structure of the classes with some technical subtleties on how to deal with groups of objects closely packed together. This is the first detection dataset to have so many classes and images and it will surely require some new breakthrough to get it right. At the time of writing there is no published or non-published results on it, although the results of an Inception ResNet Faster R-CNN baseline can be found on their site to have 37 mAP. The URL of the project is \url{https://storage.googleapis.com/openimages/web/index.html}.

For industrial applications, more often than not, the objects to detect does not come from the categories present in VOC or MS-COCO. Furthermore, they do not share the same variances; Rotation variance for instance, is a property of several applications domains but is not present in any classical common object dataset. That is why, pushed by the industry needs, several other object detection domains have appeared all with their respective literature. The most famous of them are listed in the following sections.

\subsection{Specialized datasets}
\label{subsec:datasets_specialized}
To find interesting domains one has to find interesting products or applications that drive them. The industry has given birth to many sub-fields in object detection: they wanted to have self-driving cars so we built pedestrian detection and traffic signs detection datasets; they wanted to monitor traffic so we had to have aerial imagery datasets; they wanted to be able to read text for blind persons or automatic translations of foreign languages so we constructed text detection datasets; some people wanted to do personalized advertising (arguably not a good idea) so we engineered logo datasets. They all have their place in this specialized dataset section. 

\subsubsection{Aerial Imagery}
The detection of small vehicles in aerial imagery is an old problem that has gained much attraction in recent times. However, it was only in the last years that large dataset have been made publicly available, making the topic even more popular. The following paragraphs take inventory of these datasets and of the best performing methods. 

{\em Google Earth \citep{heitz2008LearningSpatialContext}} comprises 30 images of the city of Bruxelles with 1,319 small cars and vertical bounding boxes, its variability is not enormous but it is still widely used in the literature. There are 5 folds. The CNN best result is \citep{cheng2016RIFDCNNRotationInvariantFisher} with 94.6 AP. It was later augmented with angle annotations by \citeauthor{henriques2017WarpedConvolutionsEfficient} \cite{henriques2017WarpedConvolutionsEfficient}. The data can be found on Geremy Heitz webpage (\url{http://ai.stanford.edu/~gaheitz/Research/TAS/}). 

{\em OIRDS \citep{tanner2009overhead}}, with only 180 vehicles this dataset, is not very much used by the community.

{\em DLR 3k Munich Dataset \citep{liu2015fast}}
is one of the most used datasets in the small vehicle detection literature with 20 extra large images. 10 training images with up to 3,500 cars and 70 trucks and 10 test images with 5,800 cars 90 trucks. Other classes are also available like car or truck's trails and dashed lines. The state-of-the-art seems to belong to \citep{tang2017VehicleDetectionAerial} at 83\% of F1 on both cars and trucks and \citep{tianyutang2017ArbitraryOrientedVehicleDetection} at 82\%, which provide oriented boxes. Some relevant articles that compare on this dataset are \citep{sommer2017DeepLearningBased, sommer2017FastDeepVehicle,deng2017FastAccurateVehicle}. The data can be downloaded by asking the provided contact on \url{https://www.dlr.de/eoc/en/desktopdefault.aspx/tabid-5431/9230_read-42467/}.

{\em VeDAI \citep{razakarivony2016vehicle}} is for vehicle detection is aerial images.
The vehicles contained in the database, in addition to being small, exhibit different variability such as multiple orientations, lighting/shadowing changes, occlusions. \etc Furthermore, each image is available in several spectral bands and resolutions. They provide the same images in 2 resolutions 512x512 and 1024x1024. There are a total of 9 classes and 1,200 images with an average of 5.5 instances per image. It is one of the few datasets to have 10 folds and the metric is based on an ellipse based distance between the center of the ground truth and the centers of the detections. The state-of-the-art is currently held by \citep{ogier2017icip}. Although many recent articles used their own metrics, which makes them difficult to compare \citep{tianyutang2017ArbitraryOrientedVehicleDetection,sakla2017DeepMultimodalVehicle, sommer2017FastDeepVehicle,carlet2017FastVehicleDetection,tang2017VehicleDetectionAerial}. VeDAI is available at \url{https://downloads.greyc.fr/vedai/}.

{\em COWC \citep{mundhenk2016large}}, 
introduced in ECCV2016, is a very large dataset with regions from all over the world and more than 32,000 cars. It also contains almost 60,000 hard negative patches hand-picked, which is a blessing when training detectors that do not include hard-example mining strategies. Unfortunately, no test data annotations are available so detection methods cannot yet be properly tested on it. COWC is available at \url{https://gdo152.llnl.gov/cowc/}.

{\em DOTA \citep{xia2017dota}}, released this year at CVPR, is the first mainstream dataset to change its metric to incorporate rotated bounding boxes similar to the text detections datasets. The images are of very different resolutions and zoom factors. There are 2,800 images with almost 200,000 instances and 15 categories. This dataset will surely become one of the important ones in the near future. The leader board \url{https://captain-whu.github.io/DOTA/results.html} shows that Mask R-CNN structures are the best at this task for the moment with the winner culminating at 76.2 oriented mAP but no other published method apart from \citep{xia2017dota} yet. UCAS-AOD \citep{zhu2015orientation}, NWPU VHR10 \citep{cheng2016learning} and HRSC2016 \citep{liu2017high} all provided oriented annotations also but they are hard to find and very few articles actually use them. DOTA is available at \url{https://captain-whu.github.io/DOTA/dataset.html}

{\em xView \citep{lam2018XViewObjectsContext}}
 is a very large scale dataset gathered by the pentagon, containing 60 classes and 1 million instances. It is split in three parts train, val and test. xView is available at \url{http://xviewdataset.org}. First challenge will end in August 2018, no results are available yet.

{\em VisDrone \citep{zhu2018VisionMeetsDrones}}
 is the most recent dataset including aerial images. Images, captured by different drones flying over 14 different cities separated by thousands of kilometers in China, in different scenarios under various weather and lighting conditions. The dataset consists of 263 video sequences formed by 179,264 frames and 10,209 static images and contains different objects such pedestrian, vehicles, bicycles, \etc and density (sparse and crowded scenes). Frames are manually annotated with more than 2.5 million bounding boxes and some attributes, \eg scene visibility, object class and occlusion, are provided. VisDrone is very recent and no results are available yet. VisDrone is available at \url{http://www.aiskyeye.com}. 

\subsubsection{Text Detection in Images}
Text detection in images or videos is a common way to extract content from images and opens the door to image retrieval or automatic text translation applications. We inventory, in the following, the main datasets as well as the best practices to address this problem. 

{\em ICDAR 2003 \citep{1227749}} was one of the first public datasets for text detection. The dataset contains 509 scene images and the scene text is mostly centered and iconic. \citeauthor{delakis2008text} \cite{delakis2008text} was one of the first to use CNN on this dataset. 

{\em Street View Text (SVT) \citep{wang2010word}}.
Taken from Google StreetView, it is a dataset filled with business names mostly, from outdoor streets. There are 350 images and 725 instances. One of the best performing methods on SVT is \citep{zhao2018SignTextDetection} with a F-measure of 83\%. SVT can be downloaded from \url{http://tc11.cvc.uab.es/datasets/SVT_1}.

{\em MSRA-TD500 \citep{tu2012detecting}}
 contains 500 natural images, which are taken from indoor (office and mall) and outdoor (street) scenes. The resolutions of the images vary from $1296 \times 864$ to $1920 \times 1280$. There are Chinese and English texts and mixed too. The training set contains 300 images randomly selected from the original dataset and the remaining 200 images constitute the test set. Best performing method on MSRA-TD500 is \citep{liao2018RotationSensitiveRegressionOriented} with a F-measure of 79\%. \citeauthor{shi2017detecting} \cite{shi2017detecting}, \citeauthor{yao2016SceneTextDetection} \cite{yao2016SceneTextDetection}, \citeauthor{ma2018ArbitraryOrientedSceneText} \cite{ma2018ArbitraryOrientedSceneText} and \citeauthor{zhang2016MultiorientedTextDetectionentedTextDetectionentedTextDetection} \cite{zhang2016MultiorientedTextDetectionentedTextDetectionentedTextDetection} also performed very well (F-measures of 77\%, 76\%, 75\% and 75\% respectively). The dataset is available at \url{http://www.iapr-tc11.org/mediawiki/index.php/MSRA_Text_Detection_500_Database_(MSRA-TD500)}.

{\em IIIT 5k-word \citep{mishra2012scene}}
has 1,120 images and 5,000 words from both street scene texts and born-digital images. 380 images are used to train and the remaining to test. Each text has also a category label easy or hard. \citep{liao2018RotationSensitiveRegressionOriented} is state-of-the-art, as for MSRA-TD500. IIIT 5k-word is available at \url{http://cvit.iiit.ac.in/projects/SceneTextUnderstanding/IIIT5K.html}.

{\em Synth90K~\citep{jaderberg2014synthetic}}
is a completely generated grayscale text dataset with multiple fonts and vocabulary well blended into scenes with 9 million images from a 90,000 vocabulary. It can be found on the VGG page at \url{http://www.robots.ox.ac.uk/~vgg/data/text/}

{\em ICDAR 2015 \citep{7333942}}
is another popular iteration of the ICDAR challenge, following ICDAR 2013. \citeauthor{michalbusta2017DeepTextSpotterEndToEnd} \cite{michalbusta2017DeepTextSpotterEndToEnd} got state-of-the-art 87\% of F measure in comparison to the 83.8\% of \citeauthor{liao2018RotationSensitiveRegressionOriented} \cite{liao2018RotationSensitiveRegressionOriented} and the 82.54\% of \citeauthor{jiang2017R2CNNRotationalRegion} \cite{jiang2017R2CNNRotationalRegion}. TextBoxes++ \citep{DBLP:journals/corr/abs-1801-02765} reached 81.7\% and \citeauthor{shi2017detecting} \cite{shi2017detecting} is at 75\%. 

{\em COCO Text \citep{veit2016cocotext}}, based on MS COCO, is the biggest dataset for text detection. It has 63,000 images with 173,000 annotations. \citep{liao2018RotationSensitiveRegressionOriented} is the only published result with \citep{Zhou_2017_CVPR} yet that differs from the baselines implemented in the dataset paper \citep{veit2016cocotext}. So there must still be room for improvement. The very recent \citep{DBLP:journals/corr/abs-1801-02765} outperformed \citep{Zhou_2017_CVPR}. COCO Text is available at \url{https://bgshih.github.io/cocotext/}.

{\em RCTW-17 (ICDAR 2017)~\citep{DBLP:journals/corr/abs-1708-09585}} is the latest ICDAR database. It is a large line-based dataset with mostly Chinese text. \citeauthor{liao2018RotationSensitiveRegressionOriented} \cite{liao2018RotationSensitiveRegressionOriented} achieved SOTA on this one too with 67.0\% of F measure. The dataset is available at \url{http://www.icdar2017chinese.site/dataset/}.

\subsubsection{Face Detection}
\label{sec:dataset-faces}
Face detection is one of the most widely addressed detection tasks. Even if the detection of frontal in high resolution images is an almost solved problem, there is room for improvement when the conditions are harder (non-frontal images, small faces, \etc). These harder conditions are reflected by the following recent datasets. The main characteristics of the different face datasets are proposed in Table \ref{table:face-detection}.

{\em Face Detection Data Set and Benchmark (FDDB) \citep{jain2010FDDBBenchmarkFace}}
is built using Yahoo!, with 2845 images and a total of 5171 faces; it has a wide range of difficulties such as occlusions, strong pose changes, low resolution and out-of-focus faces, with both grayscale and color images. \citeauthor{zhang2017FDSingleShot} \cite{zhang2017FDSingleShot} obtained an AUR of 98.3\% on this dataset and is currently state-of-the-art for this dataset. \citeauthor{najibi2017SSHSingleStage} \cite{najibi2017SSHSingleStage} obtained 98.1\%. The dataset can be downloaded at \url{http://vis-www.cs.umass.edu/fddb/index.html}.

{\em Annotated Facial Landmarks in the Wild (AFLW) \citep{kostinger2011AnnotatedFacialLandmarks}}
is made from a collection of images collected on Flickr, with a large variety in face appearance (pose, expression, ethnicity, age, gender) and environmental conditions. It has the particularity to not to be aimed at face detection only, but more oriented towards landmark detection and face alignment. In total 25,993 faces in 21,997 real-world images are annotated. Annotations come with rich facial landmark information (21 landmarks per faces). The dataset can be downloaded from \url{https://www.tugraz.at/institute/icg/research/team-bischof/lrs/downloads/aflw/}.

{\em Annotated Face in-the-Wild (AFW)
\citep{xiangxinzhu2012FaceDetectionPose}} is a dataset containing faces in real conditions, with their associated annotations (bounding box, facial landmarks and pose angle labels). Each image contains multiple, non-frontal faces. The dataset contains 205 images with 468 faces. \citeauthor{zhang2017FDSingleShot} \cite{zhang2017FDSingleShot} obtained an AP of 99.85\% on this dataset and is currently state-of-the-art for this dataset.

{\em PASCAL Faces \citep{yan2014FaceDetectionStructural}} 
contains images selected from PASCAL VOC \citep{everingham2010pascal} in which the faces have been annotated. \citep{zhang2017FDSingleShot} obtained an AP of 98.49\% on this dataset, and is currently state-of-the-art for this dataset.

{\em Multi-Attribute Labeled Faces (MALF ) \citep{binyang2015FinegrainedEvaluationFace}}
 incorporates richer semantic annotations such as pose, gender and occlusion information as well as expression information. It contains 5,250 images collected from the Internet and approximately 12,000 labeled faces. The dataset and up-to-date results of the evaluation can be found at \url{http: //www.cbsr.ia.ac.cn/faceevaluation/}.

{\em Wider Face \citep{yang2016wider}} is one of the largest datasets for face detection. Each annotation includes information such as scale, occlusion, pose, overall difficulty and events, which makes possible in-depth analyses. This dataset is very challenging especially for the 'hard set'. \citeauthor{najibi2017SSHSingleStage} \cite{najibi2017SSHSingleStage} obtained an AP of 93.1\% (easy), 92.1\% (medium) and 84.5\% (hard) on this dataset and is currently state-of-the-art for this dataset. \citeauthor{zhang2017FDSingleShot} \cite{zhang2017FDSingleShot} are also very good with AP of 92.8\% (easy), 91.3\% (medium) and 84.0\% (hard). Datasets and results can be downloaded at \url{http://mmlab.ie.cuhk.edu.hk/projects/WIDERFace/}.

{\em IARPA Janus Benchmark A (IJ-A) \citep{klare2015pushing}} contains images and videos from 500 subjects captured from 'in the wild' environment, and contains annotations for both recognition and detection tasks. All labeled faces are localized with bounding boxes as well as with landmarks (center of the two eyes, base of the nose). IJB-B \citep{whitelam2017iarpa} extended this dataset with 1,845 subjects, for 21,798 still images and 55,026 frames from 7,011 videos. IJB-C \citep{maze2018IARPAJanusBenchmark}, which is the new extended version of the IARPA Janus Benchmark A and B, adds 1,661 new subjects to the 1,870 subjects released in IJB-B. The NIST Face Challenges are at \url{ https://www.nist.gov/programs-projects/face-challenges}.

{\em Un-constrained Face Detection Dataset (UFDD) \citep{nada2018PushingLimitsUnconstraineda}} was built after noting that in many challenges large variations in scale, pose, appearance are successfully addressed but there is a gap in the performance of state-of-the-art detectors and real-world requirements, not captured by existing methods or datasets. UFDD aimed at identifying the next set of challenges and collect a new dataset of face images that involve variations such as weather-based degradations, motion blur and focus blur. The authors also provide an in-depth analysis of the results and failure cases of these methods. This dataset is very recent and has not been used specifically yet. However, \citeauthor{nada2018PushingLimitsUnconstraineda} \cite{nada2018PushingLimitsUnconstraineda} reported the performances (in terms of AP) of Faster-RCNN \citep{ren2015faster} (52.1\%), SSH \citep{najibi2017SSHSingleStage} (69.5\%), S3FD \citep{zhang2017FDSingleShot} (72.5\%) and HR-ER \citep{hu2016FindingTinyFaces} (74.2\%). Dataset and results can be downloaded at \url{http://www.ufdd.info/}.

{\em IIIT-Cartoon Faces in the Wild) \citep{mishra2016IIITCFWBenchmarkDatabase}}
 contains 8,927 annotated images of cartoon faces belonging to 100 famous personalities, harvested from Google image search, with annotations including attributes such as age group, view, expression, pose, etc. The benchmark includes 7 challenges: Cartoon face recognition, Cartoon face verification, Cartoon gender identification, photo2cartoon and cartoon2photo, face detection, pose estimation and landmark detection, relative attributes in Cartoon and attribute-based cartoon search. \citeauthor{jha2018BringingCartoonsLife} \cite{jha2018BringingCartoonsLife} have published SOTA detection results using a Haar features-based detector, with a F measure of 84\%. The dataset can be downloaded from \url{http://cvit.iiit.ac.in/research/projects/cvit-projects/cartoonfaces}

{\em Wildest Faces \citep{yucel2018WildestFacesFace}} is a dataset where the emphasis is put on violent scenes in unconstrained scenarios. It contains images of diverse quality, resolution and motion blur. It includes 68K images (aka video frames) and 2186 shots of 64 fighting celebrities. All of the video frames are manually annotated to foster research for detection and recognition, both. The dataset is not released at the time this survey is written. 

\begin{table}
        \centering
        \small
\resizebox{\columnwidth}{!}{
\begin{tabular}{|c|c|c|c|c|}\hline
Dataset&\#Images&\#Faces&Source&Type\\\hline\hline
FDDB \citep{jain2010FDDBBenchmarkFace} &2,845&5,171&Yahoo! News&Images\\\hline
AFLW \citep{kostinger2011AnnotatedFacialLandmarks} &21,997&25,993&Flickr&Images\\\hline
AFW \citep{xiangxinzhu2012FaceDetectionPose} &205&473&Flickr&Images\\\hline
PASCAL Faces \citep{yan2014FaceDetectionStructural} &851&1,335&Pascal-VOC&Images\\\hline
MALF \citep{binyang2015FinegrainedEvaluationFace} &5,250&11,931&Flickr, Baidu Inc.&Images\\\hline
IJB-A \citep{klare2015pushing} &24,327&67,183&Google, Bing, etc.&Images/Videos\\\hline
IIIT-CFW \citep{mishra2016IIITCFWBenchmarkDatabase}&8,927&8,928&Google&Images \\\hline
Wider Face \citep{yang2016wider} &32,203&393,703&Google, Bing&Images\\\hline
IJB-B \citep{whitelam2017iarpa} &76,824&125,474&Freebase&Images/Videos\\\hline
IJB-C \citep{maze2018IARPAJanusBenchmark} &148,876&540,630&Freebase&Images/Videos\\\hline
Wildest Faces \citep{yucel2018WildestFacesFace} &67,889&109,771&YouTube&Videos\\\hline
UFDD \citep{nada2018PushingLimitsUnconstraineda}&6,424&10,895&Google, Bing, etc. &Images\\\hline
\end{tabular}
}
\caption{Datasets for face detection. \label{table:face-detection}}
\end{table}

\begin{figure}
    \centering
    \includegraphics[height=5cm, width=8cm]{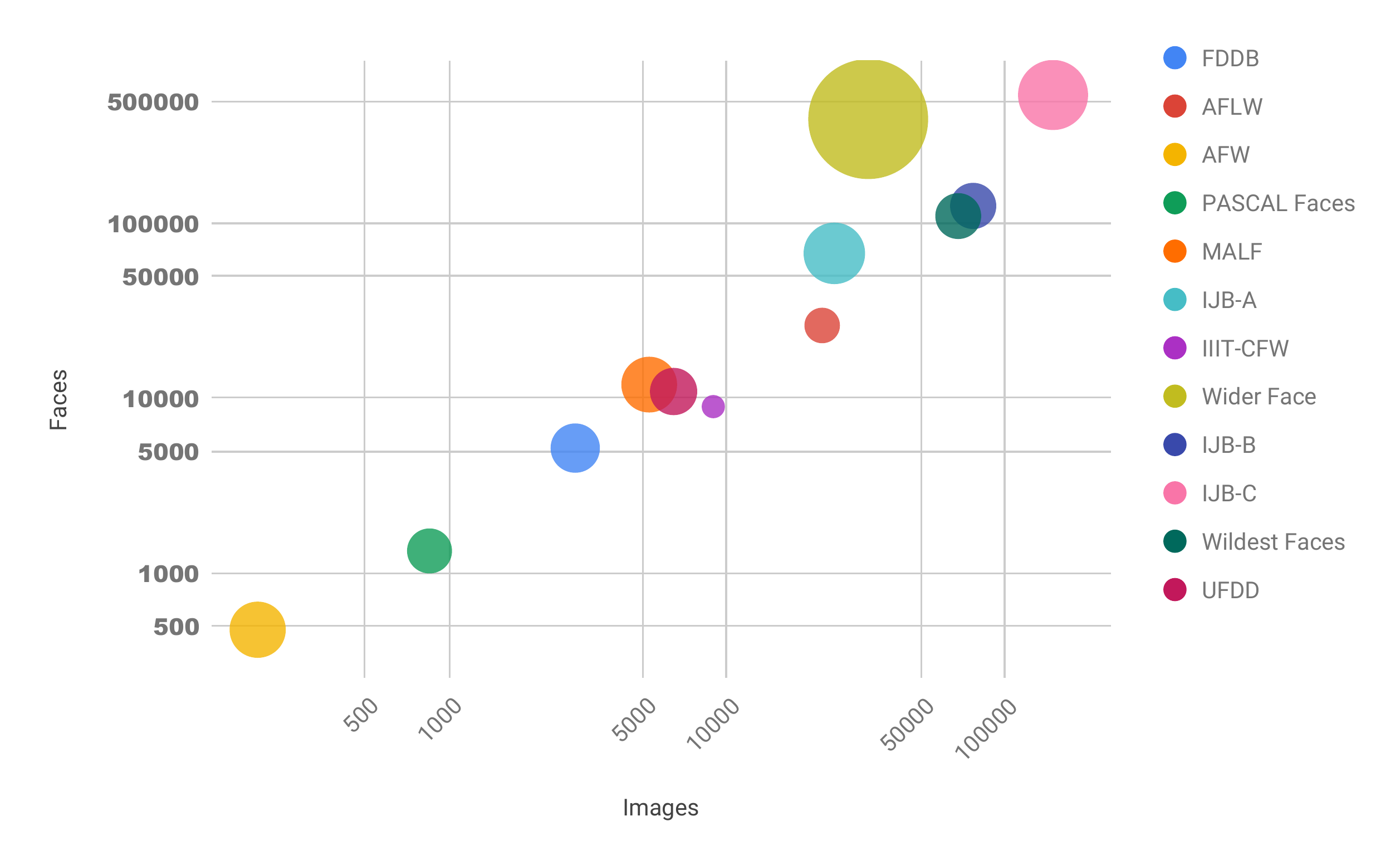}
    \caption{Number of images vs number of faces in each dataset (Table \ref{table:face-detection}) on a log scale. The size of the bubble indicates average number of faces per image which can be used as an estimate of complexity of the dataset.}
    \label{fig:face_chart}
\end{figure}

\subsubsection{Pedestrian Detection}
\label{sec:dataset-pedestrian}
Pedestrian detection is one of the specific tasks abundantly studied in the literature, especially since research on autonomous vehicles has intensified.

{\em MIT~\citep{papageorgiou2000trainable}}
is one of the first pedestrian datasets. It's puny in size (509 training and 200 testing images). The images were extracted from the LabelMe database. You can find it at \url{http://cbcl.mit.edu/software-datasets/PedestrianData.html}

{\em INRIA~\citep{dalal2005histograms}}
is currently one of the most popular static pedestrian detection datasets introduced in the seminal HOG paper \citep{dalal2005histograms}. It uses obviously the Caltech metric. \citeauthor{2018arXiv180708407Z} \cite{2018arXiv180708407Z} gained state-of-the-art with 6.4\% log average miss rate. Method at the second position is \citep{zhang2016FasterRCNNDoing} with 6.9\% using the RPN from Faster R-CNN and boosted forests on extracted features. The others are not CNN methods (the third one using pooling with HOG, LBP and covariance matrices). It can be found at \url{http://pascal.inrialpes.fr/data/human/}. Similarly, {\em PASCAL Persons dataset} is a subset of the aforementioned Pascal-VOC dataset.

{\em CVC-ADAS \citep{Gernimo2007AdaptiveIS}} 
is a collection of datasets including videos acquired on board, virtual-world pedestrians and real pedestrians. It can be found at following \url{http://adas.cvc.uab.es/site/}.

{\em USC \citep{wu2007cluster}} 
is an old small pedestrian dataset taken largely from surveillance videos. It is still downloadable at \url{http://iris.usc.edu/Vision-Users/OldUsers/bowu/DatasetWebpage/dataset.html}

{\em ETH \citep{ess2007depth}}
was captured from a stroller. There are 490 training frames with 1578 annotations. There are three test sets. The first test set has 999 frames with 5193 annotations, the second one 450 and 2359 and the third one 354 and 1828 respectively. The stereo cues are available. It is a difficult dataset where the state-of-the-art from \citeauthor{2018arXiv180708407Z} \cite{2018arXiv180708407Z} trained on CityPersons still remains at 24.5\% log average miss rate. The boosted forest of \citeauthor{zhang2016FasterRCNNDoing} \cite{zhang2016FasterRCNNDoing} gets 30.2\% only. It is available at \url{https://data.vision.ee.ethz.ch/cvl/aess/iccv2007/}

{\em Daimler DB~\citep{enzweiler2008monocular}}
is an old dataset captured in an urban setting, builds on DaimlerChrysler datasets with only grayscale images. It has been recently extended with Cyclist annotations into the Tsinghua Daimler Cyclist (TDC) dataset \citep{li2016new} with color images. The dataset is available at \url{http://www.gavrila.net/Datasets/datasets.html}.

{\em TUD-Brussels \citep{wojek2009multi}} is
from the TU Darmstadt University and contains image pairs recorded in a crowded urban setting with an on-board camera from a car. There are 1092 image pairs with 1776 annotations in the training set. The test set contains 508 image pairs with 1326 pedestrians. The evaluation is measured from the recall at 90\% precision, somehow reminiscent of KITTI dataset. TUD-Brussels is available at \url{https://www.mpi-inf.mpg.de/departments/computer-vision-and-multimodal-computing/research/people-detection-pose-estimation-and-tracking/multi-cue-onboard-pedestrian-detection/}.

{\em Caltech USA \citep{dollar2012pedestrian}}
contains images are captured in the Greater Los Angeles area by an independent driver to simulate real-life conditions without any bias. 192,000 pedestrian instances are available for training. 155,000 for testing. The evaluation use Pascal-VOC criteria at 0.5 IoU. The performance measure is the log average miss rate as application wise one cannot have too many False Positive per Image (FPPI). It is computed by averaging miss rates at 9 FPPIs from $10^{-2}$ to 1 uniformly in log scale. State-of-the-art algorithms are at around 4\% log average miss rate. \citeauthor{wang2018RepulsionLossDetecting} \cite{wang2018RepulsionLossDetecting} got 4.0\% by using a novel bounding box regression loss. Following it, we have \citeauthor{2018arXiv180708407Z} \cite{2018arXiv180708407Z} at 4.1\% using a novel RoI-Pooling of parts helping with occlusions and pre-training on CityPersons. \citeauthor{mao2017can} \cite{mao2017can} is lagging behind with 5.5\%, using a Faster R-CNN with additional aggregated features. There also exists a CalTech Japan dataset. The benchmark is hosted at \url{http://www.vision.caltech.edu/Image_Datasets/CaltechPedestrians/}.

{\em KITTI~\citep{geiger2012we}} is one of the most famous datasets in Computer Vision taken over the city of Karlsruhe in Germany. There are 100,000 instances of pedestrians. With around 6000 identities and one person in average per image. The preferred metric is the AP (Average Precision) on the moderate (persons who are less than 25 pixels tall are left behind for ranking) set. \citeauthor{li2017scale} \cite{li2017scale} got 65.01 AP on moderate by using an adapted version of Fast R-CNN with different heads to deal with different scales. The state-of-the-art of \citet{NIPS2015_5644} had to rely on stereo information to get good object proposals and 67.47 AP. All KITTI related datasets are found at \url{http://www.cvlibs.net/datasets/kitti/index.php}.

{\em GM-ATCI~\citep{silberstein2014vision}}
is a dataset captured from a fisheye-lens camera that uses CalTech evaluation system. We could not find any CNN detection results on it possibly because the state-of-the-art using multiple cues is already pretty good with 3.5\% log average miss rate. The sequences can be downloaded here \url{https://sites.google.com/site/rearviewpeds1/}

{\em CityPersons \citep{zhang2017CityPersonsDiverseDataset}}
is a relatively new dataset that builds upon CityScapes \citep{Cordts2016Cityscapes}. It is a semantic segmentation dataset recorded in 27 different cities in Germany. There are 19,744 persons in the training set and around 11,000 in the test set. There are way more identities present than in CalTech even though there are fewer instances (1300 in CalTech \wrt 19000 in CityPersons). Therefore, it is more diverse and thus, more challenging. The metric is the same as CalTech with some subsets like the Reasonable: the pedestrians that are more than 50 pixels tall and less than 35\% occluded. Again \citeauthor{2018arXiv180708407Z} \cite{2018arXiv180708407Z} and \citeauthor{wang2018RepulsionLossDetecting} \cite{wang2018RepulsionLossDetecting} take the lead with 11.32\% and 11.48\% respectively on the reasonable set \wrt the baseline on adapted Faster R-CNN that stands at 12.97\% log average miss rate. The dataset is available at \url{https://bitbucket.org/shanshanzhang/citypersons}.

{\em EuroCity~\citep{braun2018eurocity}}
is the largest pedestrian detection dataset ever released with 238,300 instances in 47,300 images. Images are taken over 31 cities in 12 different European countries. The metric is the same as CalTech. Three baselines were tested (Faster R-CNN, R-FCN and YOLOv3). Faster R-CNN dominated on the reasonable set with 8.1\%, followed by YOLOv3 with 8.5\% and R-FCN lagging behind with 12.1\%. On other subsets with heavily occluded or small pedestrians the ranking is not the same. We refer the reader to the dataset paper of \citep{braun2018eurocity}.

\subsubsection{Logo Detection}
\label{sec:dataset-logos}
Logo detection was attracting a lot of attention in the past, due to the specificity of the task. At the moment we write this survey, there are fewer papers on this topic and most of the logo detection pipelines are direct applications of Faster RCNN \citep{ren2015faster}. 

{\em BelgaLogos \citep{joly2009LogoRetrievalContrario}}
 images come from the BELGA press agency. The dataset is composed of 10,000 images covering all aspects of life and current affairs: politics and economics, finance and social affairs, sports, culture and personalities. All images are in JPEG format and have been re-sized with a maximum value of height and width equal to 800 pixels, preserving aspect ratio. There are 26 different logos. Only a few images are annotated with bounding boxes. The dataset can be downloaded at \url{https://www-sop.inria.fr/members/Alexis.Joly/BelgaLogos/BelgaLogos.html}. 

{\em FlickrLogos \citep{romberg2011scalable,eggert2017improving}}
 consists of real-world images collected from Flickr, depicting company logos in various situations. The dataset comes in two versions: The original FlickrLogos-32 dataset and the FlickrLogos-47 \citep{eggert2017improving} dataset. In FlickrLogos-32 the annotations for object detection were often incomplete, since only the most prominent logo instances were labeled. FlickrLogos-47 uses the same image corpus as FlickrLogos-32 but new classes were introduced (logo and text as separate classes) and missing object instances have been annotated. FlickrLogos-47 contains 833 training and 1402 testing images. The dataset can be downloaded at \url{http://www.multimedia-computing.de/flickrlogos/}.

{\em Logo32plus \citep{bianco2017DeepLearningLogo}}
 is an extension of the train set of FlickrLogos-32 \citep{eggert2017improving}. It has the same classes of objects but much more training instances (12,312 instances). The dataset can be downloaded at \url{http://www.ivl.disco.unimib.it/activities/logorecognition}.

{\em WebLogo-2M \citep{su2017WebLogo2MScalableLogo}} is very large, but annotated at image level only and does not contain bounding boxes. It contains 194 logo classes and over 2 million logo images. Labels are noisy as the annotations are automatically generated. Therefore, this dataset is designed for large-scale logo detection model learning from noisy training data. For performance evaluation, the dataset includes 6,569 test images with manually labeled logo bounding boxes for all the 194 logo classes. The dataset can be downloaded at \url{http://www.eecs.qmul.ac.uk/%7Ehs308/WebLogo-2M.html/}.

{\em SportsLogo \citep{liao2017MutualEnhancementDetection}},
in the absence of public video logo dataset, was collected on a set of tennis videos containing 20 different tennis video clips with camera motions (blurring) and occlusion. The logos can appear on the background as well as on players’ and staff’s clothes. 20 logos are annotated, with about 100 images for each logo. 

{\em Logos in the Wild \citep{tuzko2017open}}
contains images collected from the web with logo annotations provided in Pascal-VOC style. It contains large varieties of brands in-the-wild. The latest version (v2.0) of the dataset consists of 11,054 images with 32,850 annotated logo bounding boxes of 871 brands. It contains from 4 to 608 images per searched brand, and 238 brands occur at least 10 times. It has up to 118 logos in one image. Only the links to the images are released, which is problematic as numerous images have already disappeared, making exact comparisons impossible. The dataset can be downloaded from \url{https://www.iosb.fraunhofer.de/servlet/is/78045/}.

{\em Open Logo Detection Challenge \citep{su2018OpenLogoDetection}}.
This dataset assumes that only on a small proportion of logo classes are annotated whilst the remaining classes have no labeled training data. It contrasts with previous logo datasets which assumed all the logo classes are annotated. The OpenLogo challenge contains 27,189 images from 309 logo classes, built by aggregating/refining 7 existing datasets and establishing an open logo detection evaluation protocol. The dataset can be downloaded at \url{https://qmul-openlogo.github.io}.

\begin{table}
\resizebox{\columnwidth}{!}{
\begin{tabular}{|c|c|c|}\hline
Dataset&\#Classes&\#Images\\\hline\hline
BelgaLogos \citep{joly2009LogoRetrievalContrario}&26&10,000\\\hline
FlickrLogos-32 \citep{romberg2011scalable}&32&8,240\\\hline
FlickrLogos-47 \citep{eggert2017improving}&47&8,240\\\hline
Logo32plus \citep{bianco2017DeepLearningLogo}&32&7,830\\\hline
WebLogo-2M \citep{su2017WebLogo2MScalableLogo} &194&2,190,757\\\hline
SportsLogo \citep{liao2017MutualEnhancementDetection}&20&1,978\\\hline
Logos in the Wild \citep{tuzko2017open}&871&11,054\\\hline
OpenLogos \citep{su2018OpenLogoDetection}&309&27,189\\\hline
\end{tabular}
}
\caption{Datasets for logo detection.}
\end{table}
 
\subsubsection{Traffic Signs Detection}
This section reviews the 4 main datasets and benchmarks for evaluating traffic sign detectors \citep{mogelmose2012VisionBasedTrafficSign,Houben-IJCNN-2013,timofte2014multi,zhu2016traffic}, as well as the Bosch Small Traffic Lights \citep{BehrendtNovak2017ICRA}. The most challenging one is the Tsinghua Tencent 100k (TTK100) \citep{zhu2016traffic}, on which Faster RCNN like detectors detectors such as \citep{pon2018HierarchicalDeepArchitecture} have an overall precision/recall of 44\%/68\%, which shows the difficulty of the dataset.

{\em LISA Traffic Sign Dataset \citep{mogelmose2012VisionBasedTrafficSign}}
 was among the first datasets for traffic sign detection. It contains 47 US signs and 7,855 annotations on 6,610 video frames. Sign sizes vary from 6x6 to 167x168 pixels. Each sign is annotated with sign type, position, size, occluded (yes/no), on side road (yes/no). The URL for this dataset is \url{http://cvrr.ucsd.edu/LISA/lisa-traffic-sign-dataset.html}

{\em The German Traffic Sign Detection Benchmark (GTSDB) \citep{Houben-IJCNN-2013}} is one of the most popular traffic signs detection benchmarks. It introduced a dataset with evaluation metrics, baseline results, and a web interface for comparing approaches. The dataset provides a total of 900 images with 1,206 traffic signs. The traffic sign sizes vary between 16 and 128 pixels \wrt the longest edge. The image resolution is $1360 \times 800$; images capture different scenarios (urban, rural, highway) during the daytime and dusk featuring various weather conditions. It can be found at \url{http://benchmark.ini.rub.de/?section=gtsdb&subsection=news}.

{\em Belgian TSD \citep{timofte2014multi}} consists of 7,356 still images for training, with a total of 11,219 annotations, corresponding to 2,459 traffic signs visible at less than 50 meters in at least one view. The test set contains 4 sequences, captured by 8 roof-mounted cameras on the van, with a total of 121,632 frames and 269 different traffic signs for evaluating the detectors. For each sign, the type and 3D location is given. The dataset can be downloaded at \url{https://btsd.ethz.ch/shareddata/}.

{\em Tsinghua Tencent 100k (TTK100) \citep{zhu2016traffic}}
provides $2048 \times 2048$ images for traffic signs detection and classification, with various illumination and weather conditions. It's the largest dataset for traffic signs detection, with 100,000 images out of which 16,787 contain traffic signs instances, for a total of 30,000 traffic instances. There are a total of 128 classes. Each instance is annotated with class label, bounding box and pixel mask. It has small objects in abundance and huge scale variations. Some signs which are naturally rare, $e.g.$ signs to warn the driver to be cautious on mountain roads appear, have quite low number of instances. There are 45 classes with at least 100 instances present. The dataset can be obtained at \url{http://cg.cs.tsinghua.edu.cn/traffic%2Dsign/}.

{\em Bosch Small Traffic Lights \citep{BehrendtNovak2017ICRA}}
 is made for benchmarking traffic light detectors. It contains 13,427 images of size $1280 \times 720$ pixels with around 24,000 annotated traffic lights, annotated with bounding boxes and states (active light). Best performing algorithm is \citep{pon2018HierarchicalDeepArchitecture} which obtained a mAP of 53 on this dataset. Bosch Small Traffic Lights can be downloaded at \url{https://hci.iwr.uni-heidelberg.de/node/6132}.

\subsubsection{Other Datasets}
Some datasets do not fit in any of the previously mentioned category but deserve to be mentioned because of the interest the community has for them.

{\em iNaturalist Species Classification and Detection Dataset \citep{vanhorn2018INaturalistSpeciesClassification}} contains 859,000 images from over 5,000 different species of plants and animals. The goal of this dataset is to encourage the development of algorithms for 'in the wild' data featuring large numbers of imbalanced, one-grained, categories. The dataset can be downloaded at \url{https://github.com/visipedia/inat_comp/tree/ master/2017}. 

Below we give all known datasets that can be used to tackle object detection with the different modalities that we presented in the Sec. \ref{sec:detecting_in_other_modalities}.

\subsection{3D Datasets}
\label{sec:3Ddatasets}
{\em KITTI object detection benchmark \citep{geiger2012we}}
 is the most widely used dataset for evaluating detection in 3D point clouds. It contains 3 main categories (namely 2D, 3D and birds-eye-view objects), 3 object categories (cars, pedestrians and cyclists), and 3 difficulty levels (easy, moderate and hard considering the object size, distance, occlusion and truncation). The dataset is public and contains 7,481 images for training and 7,518 for testing, comprising a total of 80,256 labeled objects. The 3D point clouds are acquired with a Velodyne laser scanner. 3D object detection performance is evaluated using the PASCAL criteria also used for 2D object detection. For cars a 3D bounding box overlap of 70\% is required, while for pedestrians and cyclists a 3D bounding box overlap of 50\% is required. For evaluation, precision-recall curves are computed and the methods are ranked according to average precision. The algorithms can use the following sources of information: i) Stereo: Method uses left and right (stereo) images ii) Flow: Method uses optical flow (2 temporally adjacent images) iii) Multiview: Method uses more than 2 temporally adjacent images iv) Laser Points: Method uses point clouds from Velodyne laser scanner v) Additional training data: Use of additional data sources for training. The datasets and performance of SOTA detectors can be download at \url{http://www.cvlibs.net/datasets/kitti/}, and the leader board is at \url{http://www.cvlibs.net/datasets/kitti/eval_object.php?obj_benchmark=3d}. One of the leading methods is \citep{simon2018ComplexYOLORealtime3D} which is at an mAP of 67.72/64.00/63.01 (Easy/Mod./Hard) for the car category, at 50 fps. Slower (10 fps) but more accurate, \citep{ku2017Joint3DProposal} has a performance of 81.94/71.88/66.38 on cars. \citeauthor{chen2016MultiView3DObject} \cite{chen2016MultiView3DObject}, \citeauthor{zhou2017VoxelNetEndtoEndLearning} \cite{zhou2017VoxelNetEndtoEndLearning} and \citeauthor{qi2017FrustumPointNets3Da} \cite{qi2017FrustumPointNets3Da} also gave very good results.

{\em Active Vision Dataset (AVD) \citep{ammirato2017DatasetDevelopingBenchmarking}} contains 30,000+ RGBD images, 30+ frequently occurring instances, 15 scenes, and 70,000+ 2D bounding boxes. This dataset focused on simulating robotic vision tasks in everyday indoor environments using real imagery. The dataset can be downloaded at \url{http://cs.unc.edu/~ammirato/active_vision_dataset_website/}.

{\em SceneNet RGB-D \citep{mccormac2017SceneNetRGBDCan}}
is a synthetic dataset designed for scene understanding problems such as semantic segmentation, instance segmentation, and object detection. It provides camera poses and depth data and permits to create any scene configuration. 5M rendered RGB-D images from 16K randomly generated 3D trajectories in synthetic layouts are also provided. The dataset can be downloaded at \url{http://robotvault. bitbucket.io/scenenet-rgbd.html}.

{\em Falling Things \citep{tremblay2018FallingThingsSynthetic}}
introduced a novel synthetic dataset for 3D object detection and pose estimation, the Falling Things dataset. The dataset contains 60k annotated photos of 21 household objects taken from the YCB dataset. For each image, the 3D poses, per-pixel class segmentation, and 2D/3D bounding box coordinates for all objects are given. To facilitate testing different input modalities, mono and stereo RGB images are provided, along with registered dense depth images. The dataset can be downloaded at \url{http://research.nvidia.com/publication/2018-06_Falling-Things}.

\subsection{Video Datasets}
\label{sec:videodatasets}
The two most popular datasets for video object detection are the YouTube-BoundingBoxes \citep{real2017YouTubeBoundingBoxesLargeHighPrecision} and the ImageNet VID challenge \citep{russakovsky2015ImageNetLargeScale}. Both are reviewed in this section.

{\em YouTube-BoundingBoxes \citep{real2017YouTubeBoundingBoxesLargeHighPrecision}}
is a data set of video URLs with the single object bounding box annotations. All video sequences are annotated with classifications and bounding boxes, at 1 frame per second. There is a total of about 380,000 video segments of 15-20 seconds, from 240,000 publicly available YouTube videos, featuring objects in natural settings, without editing or post-processing. \citeauthor{real2017YouTubeBoundingBoxesLargeHighPrecision} \cite{real2017YouTubeBoundingBoxesLargeHighPrecision} reported a mAP of 59 on this dataset. This dataset can be downloaded at \url{https://research.google.com/youtube-bb/}.

{\em ImageNet VID challenge \citep{russakovsky2015ImageNetLargeScale}}
was a part of the ILSVRC 2015 challenge. It has a training set of 3,862 fully annotated video sequences having a length from 6 frames to 5,492 frames per video. The validation set contains 555 fully annotated videos, ranging from 11 frames to 2898 frames per video. Finally, the test set contains 937 video sequences and the ground-truth annotation are not publicly available. One of the best performing methods on ImageNet VID is \citep{feichtenhofer2017detect} with a mAP of 79.8, by combining detection and tracking. \citeauthor{zhu2017flow} \cite{zhu2017flow} reached 76.3 points with a flow best approach. This dataset can be downloaded at \url{http://image-net.org/challenges/LSVRC}.

{\em VisDrone \citep{zhu2018VisionMeetsDrones}}
 contains video clips acquired by drones. This dataset is presented in Section 5.2.1

\subsection{Concluding Remarks}
This appendix gave a large overview of the datasets introduced by the community for developing and evaluating object detectors in images, videos or 3D point clouds. Each object detection dataset presents a very biased view of the world, as shown in \citep{torralba2011unbiased,khosla2012undoing,tommasi2017deeper}, representative of the user's needs when they built it. The bias is not only in the images they chose (specific views of objects, objects imbalance \citep{ouyang2016FactorsFinetuningDeep}, objects categories) but also in the metric they created and the evaluation protocol they devised. The community is trying its best to build more and more datasets with less and less bias and as a result it has become quite hard to find its way in this jungle of datasets, especially when one needs: older datasets that have fallen out of fashion or even exhaustive lists of state-of-the-art algorithms performances on modern ones. Through this survey we have partially addressed this need of a common source for information on datasets.